%% file: main.tex
\documentclass{article}

\usepackage[main, final]{neurips_2026}

\usepackage[utf8]{inputenc} 
\usepackage[T1]{fontenc}    
\usepackage{amsmath}
\usepackage{amssymb}
\usepackage{mathtools}
\usepackage{amsthm}

\usepackage{graphicx} 
\PassOptionsToPackage{square,sort,comma,numbers}{natbib}

\usepackage{hyperref}
\hypersetup{colorlinks=true,citecolor=blue,linkcolor=blue}
\usepackage{url}
\usepackage{graphicx}
\usepackage{caption}
\usepackage{subcaption}
\usepackage{booktabs}
\usepackage{multirow}
\usepackage{amsmath,bm}
\usepackage{amsfonts,amssymb}

\usepackage[algo2e]{algorithm2e}

\usepackage{xcolor}
\usepackage{colortbl}
\usepackage{enumitem}
\usepackage{pifont}
\usepackage{wrapfig}

\usepackage{amssymb}
\usepackage{amsmath,mathrsfs,dsfont}
\usepackage{nicefrac}
\usepackage{booktabs}

\usepackage{color}
\usepackage{enumitem}

\usepackage{array}
\usepackage{graphicx,tikz}
\usepackage[mathscr]{euscript}
\PassOptionsToPackage{square,sort,comma,numbers}{natbib}
\usepackage{multirow}

\usepackage{bm}
\usepackage{bbm}
\usepackage{color}
\usepackage{epstopdf}
\usepackage{mathtools}
\usepackage{thmtools}
\usepackage{thm-restate}
\usepackage{bbding}
\usepackage{algorithmic}
\usepackage{algorithm}

\graphicspath{{figures/}}

\allowdisplaybreaks
\usepackage{makecell}
\input{symbol_tit.tex}

\usepackage{wrapfig}

\usepackage[capitalize,noabbrev]{cleveref}
\title{RGDM: Robust Graph Diffusion Model}

\author{%
  Yancheng Wang\,$^{1}$, Ping Li\,$^{2}$, Dongfang Sun\,$^{1}$,
   Yingzhen Yang\,$^{1}$ \\[0.5em]
  \begin{tabular}[t]{@{}c@{\hspace{1.5em}}c@{}}
    $^{1}$ School of Computing and Augmented Intelligence
      & $^{2}$ VecML Inc. \\
    Arizona State University
      & Bellevue, WA 98004, USA \\
    Tempe, AZ 85281, USA
      & \texttt{pingli98@gmail.com} \\
    \texttt{\{yancheng.wang,dsun30,yingzhen.yang\}@asu.edu}
      &
  \end{tabular}%
}

\begin{document}

\maketitle

\begin{abstract}
Diffusion models represent a powerful class of generative models known for their solid theoretical foundations and remarkable performance across diverse tasks and domains. Existing node level diffusion methods synthesize nodes and edges within a graph, but are not designed to learn a robust generator from adversarially corrupted graphs. In this study, we introduce the Robust Graph Diffusion Model (RGDM), designed to generate labeled synthetic graph structures consisting of nodes and edges that integrate seamlessly into a given graph.
The RGDM consists of a Robust Graph Autoencoder (RGAE) and a Latent Diffusion Model (LDM). Leveraging an edge selection mechanism and an innovative low-rank regularization on the latent feature, the RGDM produces clean and high-quality synthetic graph structures, even when trained on graphs subject to adversarial attacks. Comprehensive experimental evaluations reveal that Graph Neural Networks (GNNs) trained on the augmented graph, which is formed by merging the original attacked graph with the synthetic graph structures, exhibit significantly improved robustness against various graph adversarial attacks in the context of semi-supervised node classification.
The code of the RGDM is available at \url{https://github.com/Statistical-Deep-Learning/RGDM}.
\end{abstract}

\section{Introduction}\label{sec:introduction}
\begin{wrapfigure}{r}{0.3\textwidth}
    \centering
    \vspace{-3mm}
    \includegraphics[width=0.25\textwidth]{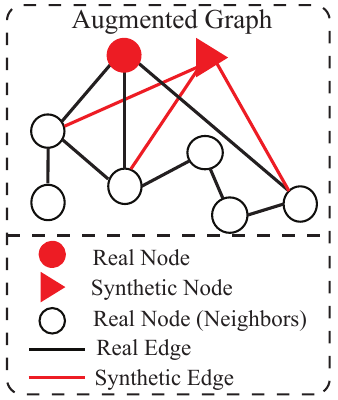}
    \vspace{-2mm}
  \caption{Illustration of the 2-hop neighborhood of a real node and a synthetic node in the augmented graph. Figure~\ref{fig:tsne} in Section~\ref{sec:tsne} illustrates the t-SNE visualization of the augmented graph.}
  \vspace{-3mm}
  \label{fig:visualization}
\end{wrapfigure}

\vspace{-2mm}
Building on the success of diffusion and latent diffusion models, recent works use diffusion-generated data to augment training sets~\citep{trabucco2023effective, azizi2023synthetic}. However, extensions to graphs~\citep{ScorebasedGraph, ScoreBasedSDE, DiffusionGraphBenefitDiscrete, DiGressDDGraph, SaGessSamplingGraph} mainly target graph-level augmentation. Most node-level methods expand labels through mix-up~\citep{G-mixup, wang2021mixup, verma2021graphmix,zhao2024geomix, jeong2024igraphmix, lu2024nodemixup} or GAN-based synthesis for imbalance~\citep{GAN-jia2023deep, GAN-wu2023graph, GAN-wang2018graphgan, GAN-liang2020feedergan, GAN-yang2019conditional}. Node level diffusion methods DoG~\citep{WangLY25} and LR-FGDM~\citep{Wang2026LRFGDM} share the autoencoder and diffusion template, but do not train the generator specifically for adversarial robustness. Moreover, GANs can be unstable and produce distributions misaligned with real data~\citep{dhariwal2021diffusion}.
As a result, the challenge of generating synthetic graph structures (SGS) within a given graph, which are the synthetic nodes and their corresponding edges, in a robust manner to enhance node classification under attacks, remains underexplored.
In this work, we tackle the problem of node-level graph generation and propose the Robust Graph Diffusion Model (RGDM), designed to synthesize high-quality graph structures within a single graph, even under adversarial attacks. These synthetic structures include both synthetic nodes and edges connecting them within the same graph, forming an augmented graph as illustrated in Figure~\ref{fig:visualization}. Robust GNNs will be trained on the augmented graph, and superior performance is expected due to the enlarged set of labeled nodes. 

Generating high-quality SGS in adversarial scenarios for improved performance is challenging, important, and unaddressed in the robust graph learning literature.
To address this challenge, RGDM integrates a Robust Graph Autoencoder (RGAE) with a Latent Diffusion Model (LDM). The RGAE incorporates an edge selection mechanism and a low-rank regularization on the latent features to counteract adversarial perturbations in the attacked graph.
The low-rank regularization is motivated by the Low Frequency Property (LFP)~\citep{rahaman19a-spectral-bias,AroraDHLW19-fine-grained-two-layer,CaoFWZG21-spectral-bias,ChorariaD0MC22-spectral-bias-pnns,WangGNSWY24,WangLY25}: in the evaluated latent spaces, label information is more concentrated in leading eigendirections than adversarial perturbations (Section~\ref{sec:proj_sup}). Theorem~\ref{theorem:optimization-linear-kernel} provides a complementary, conditional justification through the spectral complexity of a bounded linear probe on this latent space. It is not a guarantee for the downstream GNN trained on the augmented graph.

 Figure~\ref{fig:visualization} illustrates the 2-hop neighborhood of a real node and a synthetic node that have similar node features, evidenced by the t-SNE visualization in Figure~\ref{fig:tsne} in Section~\ref{sec:tsne} of the appendix.
The fidelity evaluations in Tables~\ref{tab:quality} and~\ref{tab:quality_edge} compare synthetic structures with clean reference graphs through Frechet Node Distance (FND) and Frechet Edge Distance (FED). RGDM has lower distances than vanilla DDPM~\citep{DDPM} and the compared node augmentation methods. On Pubmed under $25\%$ Metattack, its FND/FED are $3.95/5.34$, compared with $3.79/5.15$ for the clean reference. These are empirical fidelity measurements. Clean graphs are used for evaluation, not for training RGDM.
The augmentation pipeline retains the original attacked adjacency. Edge selection acts inside the generator, reducing the reproduction of unreliable connections. The generated labeled structures supply additional supervision and relatively clean message passing paths. On Citeseer, adding $785$ synthetic edges changes the normalized perturbation density from $25\%$ to approximately $20.6\%$. This is a denominator effect, not removal of the original perturbations or a causal robustness guarantee (Section~\ref{sec:perturbation_density}).
\noindent\textbf{Difference from diffusion based purification.}
Diffusion models have also been used for data and graph purification~\citep{GDDM,LuoSYF025,ZhaoYLZG025,XieLZ025,ChenGZY023}. Purifying an observed graph and generating additional labeled nodes with incident edges are different tasks. Graph level purification methods do not directly provide the node level augmentation interface of RGDM. Conversely, vanilla node generation trained on a corrupted graph can reproduce its noise. RGDM trains its robust generator without paired clean graphs. Section~\ref{sec:sup_purification_compare} compares GDDM and DiffSP preprocessing with RGDM augmentation using the same SG-GSR backbone, while Section~\ref{sec:sup_rgdm_purification} separately evaluates a purification adaptation of RGDM.

\noindent\textbf{Contributions.}
First, RGDM learns a robust node level generator through edge selection, sparse hierarchical reconstruction, and low rank regularization of the latent features used by diffusion. Its Sparse Hierarchical Edge Decoder (SHED) evaluates selected clusters instead of globally scoring every candidate neighbor. Controlled comparisons with DoG and LR-FGDM, componentwise ablations, and wall clock measurements distinguish these contributions from the shared autoencoder and diffusion architecture.
Second, RGDM augments existing classifiers without changing their architecture or objective. We evaluate its incremental benefit with each backbone fixed, including SG-GSR~\citep{in2024self}, and report paired variability. The gains over this strong backbone are modest in absolute magnitude, but consistent in the tested settings. Fidelity, heterophily, and adaptive attack evaluations support this empirical conclusion. The linear probe theorem motivates the latent regularizer and does not establish an end to end robustness guarantee.
\vspace{-2mm}
\section{Related Works}
\label{sec:related_works}
\vspace{-1mm}
\noindent\textbf{Data Augmentation for Graph Learning.}
To improve the performance of GNNs, node-level graph data augmentation methods have been studied to modify the structure~\citep{gasteiger2019diffusion, zhao2021data, RongHXH20, GRAND+, lai2024efficient}, features~\citep{YouCSCWS20,KongLDWZGTG22,azad2024learned}, or node labels~\citep{YouCSCWS20,KongLDWZGTG22,azad2024learned}.
Studies have demonstrated that increasing the number of labeled nodes based on Mix-up strategies~\citep{G-mixup, wang2021mixup, verma2021graphmix,zhao2024geomix, jeong2024igraphmix, lu2024nodemixup} can greatly enhance the performance of the GNNs.
Recent works address node and edge imbalance by generating synthetic nodes and edges~\citep{SGS-GAN-qu2021imgagn, SGS-GAN-zhao2021graphsmote, hsu2024fincgan, gao2023semantic, SGS-GAN-hsu2023gan}, but predominantly rely on Generative Adversarial Networks (GANs). However, the computer vision literature~\citep{dhariwal2021diffusion} has demonstrated that GANs often exhibit instability during training and that the distribution of synthetic data generated by GANs poorly matches the distribution of real data.
Graph level diffusion models~\citep{ScoreBasedSDE,ScorebasedGraph,DiffusionGraphBenefitDiscrete,DiGressDDGraph,SaGessSamplingGraph} generate entire graphs. DoG~\citep{WangLY25} and LR-FGDM~\citep{Wang2026LRFGDM} instead generate nodes and associated edges within an existing graph, as does RGDM. DoG reconstructs observed neighborhoods with a Bi-Level Neighborhood Decoder (BLND) and applies low rank regularization to its downstream GNN. LR-FGDM addresses few shot classification using a Hierarchical Graph Autoencoder (HGAE) with prototypical regularization, followed by a low rank transductive classifier. RGDM places edge selection, a sparsification objective, and low rank regularization inside the generator. Its TNN penalty acts on the latent features used by diffusion. SHED reconstructs selected sparse neighborhoods rather than all observed connections. The resulting augmented graph can be used with GCN, Pro-GNN, or SG-GSR. Tables~\ref{tab:compare_augmentation} and~\ref{tab:component_isolation} compare these methods with a fixed backbone. Table~\ref{tab:BLND_compare} additionally compares hierarchical reconstruction variants.

\noindent\textbf{Graph Adversarial Attacks and Defense.}
Despite the success of GNNs in graph-structured tasks~\citep{kipf2017semi, zhang2018link}, recent studies reveal their vulnerability to adversarial attacks~\citep{dai2022comprehensive}.  Depending on the attack objective, threat models are categorized as: targeted attacks~\citep{zugner2018adversarial}, which mislead predictions on specific nodes, and untargeted attacks~\citep{zugner_adversarial_2019, sun2020adversarial}, which reduce overall model accuracy. To defend against these threats, robust learning strategies are grouped into three categories: adversarial training, graph processing, and model robustification. Adversarial training~\citep{li2022spectral, feng2019graph} augments training data with crafted adversarial samples. Graph processing~\citep{wu2019adversarial, entezari2020all, jin2020graph, lei2022evennet} seeks to denoise perturbed graphs. Model robustification~\citep{xie2023adversarially, rusch2022graph, song2022robustness, zhao2023adversarial, jia2023enhancing, GADC, GCORNs, in2024self} enhances model resilience.

\vspace{-2mm}
\section{Methods}
\label{sec:methods}
\vspace{-1mm}
\subsection{Preliminaries of the Attributed Graph}
\vspace{-1mm}
\label{sec:notations}
An attributed graph of $N$ nodes is represented by $\cG = (\cV, \bX, \bA)$, where $\cV = \set{v_1, v_2, \dots, v_N}$ and $\cE\subseteq\cV\times\cV$ denote the nodes and edges respectively. $\bX \in \RR^{N \times D}$ contains the node attributes. $\bX_i \in \RR^D$, the $i$-th row of $\bX$, denotes the attributes of the $i$-th node. $\bA \in \set{0, 1}^{N \times N}$ is the adjacency matrix of the graph $\cG$. $\Tilde{\bA} = \bA + \bI$ is the adjacency matrix of a graph with self-loops added. $\Tilde{\mathbf{D}}$ denotes the diagonal degree matrix of $\Tilde{\bA}$. We use $\cN(i)=\set{j\mid \tilde{\bA}_{i,j}=1}$ to denote the set of nodes connected to $v_i$. $\bZ \in \RR^{N \times D'}$ contains the latent features of all the nodes in the graph, where $D'$ is the latent dimension.
Let $\cV_{\cL}$ and $\cV_{\cU}$ denote the set of labeled nodes and unlabeled test nodes, respectively, and $\abth{\cV_{\cL}} = m$, $\abth{\cV_{\cU}} = u$.
Let $Y_L=\set{y_i\mid v_i\in\cV_L,\ y_i\in[C]}$ denote the labels of nodes in $\cV_{\cL}$, where $C$ is the number of classes. Let $\bY = \bth{\by_1;\by_2;\ldots \by_N}\in \RR^{N \times C}$ be the ground truth label matrix of all the nodes in $\cG$, where $\by_i$ is the one-hot label vector of node $v_i$.
 Let $\bu \in \RR^N$ be a vector. We use $\bth{\bu}_{\cA}$ to denote a vector formed by elements of $\bu$ with indices in $\cA$ for $\cA \subseteq [N]$. If $\bu$ is a matrix, then $\bth{\bu}_{\cA}$ denotes a submatrix formed by rows of $\bu$ with row indices in $\cA$. $\fnorm{\cdot}$ denotes the Frobenius norm of a matrix, and $\norm{\cdot}{p}$ denotes the $p$-norm of a vector.
\vspace{-2mm}
\subsection{Robust Graph Autoencoder (RGAE)}
\vspace{-1mm}
\label{sec:robust_GAE}
In this section, we propose the Robust Graph Autoencoder (RGAE) capable of encoding node attributes and associated edges into continuous latent features and decoding these features to reconstruct the node attributes and edges in the graph. Beyond reconstruction, the RGAE is designed to be robust to adversarial perturbations in the node attributes and the edges in the graph, enhancing the resilience against potential graph adversarial attacks.
\begin{figure*}[!htbp]
  \centering
  \vspace{-2mm}
    \includegraphics[width=1\textwidth]{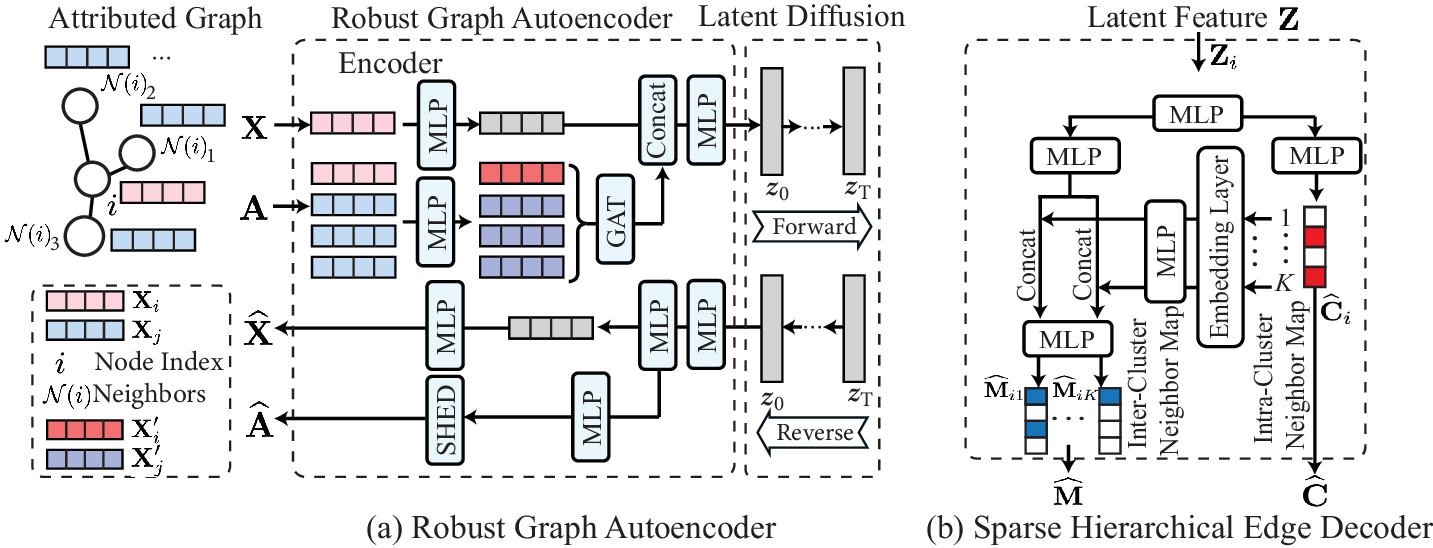}
    \vspace{-3mm}
  \caption{Figure (a) illustrates the synthetic node generation process in Robust Graph Diffusion Model. Figure (b) illustrates the structure of the Sparse Hierarchical Edge Decoder (SHED).}
  \label{fig:overall_framework}
  \vspace{-3mm}
\end{figure*}
We encode each node's attributes together with its neighborhood. Positional embeddings give $\bX'_j=\bX_j+\textup{pos}(j)$, and GAT layers~\citep{GAT} aggregate these features. For $j\in\cN(i)$, the attention weight is $\bB_{ij}=\exp(a_{ij})/\sum_{\ell\in\cN(i)}\exp(a_{i\ell})$, where $a_{ij}=\sigma(\bX'_i\bQ\Vert\bX'_j\bQ)\bV$, $\Vert$ denotes concatenation, $\sigma$ is the sigmoid function, and $\bQ,\bV$ are learnable matrices. Attention is restricted to observed neighbors.
To suppress unreliable edges during RGAE training, a learned binary mask $\bS$ selects connections. We use the sigmoid relaxation $\hat{\bS}_{ij}=\sigma(\btheta_{ij}/\tau)$ associated with binary Gumbel Softmax training~\citep{verelst2020dynamic,Bengio2013EstimatingComputation}. A linear layer applied to $\bX'_i\Vert\bX'_j$ predicts $\btheta_{ij}$. The forward pass uses $\bS_{ij}=\indict{\hat{\bS}_{ij}>1/2}$, while a straight through estimator differentiates through $\hat{\bS}_{ij}$. We use $\tau=0.5$ unless stated otherwise.
The selected neighbor aggregation is $\bZ'_i=\textup{ReLU}(\sum_{j\in\cN(i)}\bS_{ij}\bB_{ij}\bX'_j\bW)$, with learnable $\bW$. Two GAT layers yield the neighborhood representation. Concatenating it with an MLP of the node attributes gives $\bZ_i=\bZ'_i\Vert f(\bX_i)$. The mask changes the generator's training representation, not the original adjacency retained for downstream augmentation.
\noindent\textbf{Sparse Hierarchical Edge Decoder (SHED).} Decoding edges from latent features can be computationally intensive, as each node may connect to any other node. To mitigate this, we propose a Sparse Hierarchical Edge Decoder (SHED), which decodes each latent feature into an inter-cluster and an intra-cluster neighbor map, as illustrated in Figure~\ref{fig:overall_framework} (b).
The SHED begins by partitioning the nodes into $K$ balanced clusters using balanced $K$-means clustering~\citep{malinen2014balanced} based on node attributes. Each cluster has a maximum capacity of $M = \lceil \frac{N}{K} \rceil$. Within each cluster, nodes are indexed by their original graph order. For a node $v_i$, SHED first reconstructs an inter-cluster neighbor map $\bC \in \set{0,1}^{N \times K}$, where $\bC_{ik} = 1$ if $v_i$ is connected to at least one node in cluster $k$. It then reconstructs the intra-cluster neighbor map as a tensor $\bM \in \set{0,1}^{N \times K \times M}$, where $\bM_{ikm} = 1$ indicates that $v_i$ is connected to the $m$-th node in cluster $k$.
The SHED begins by computing the inter-cluster neighbor assignment $\hat{\bC}_i$ for each node $v_i$ using a single-layer MLP.
At generation time, the active cluster indices, $\cC(i) = \set{k \in [K] \mid \hat{\bC}_{ik} > 1/2}$, are extracted and passed through a cluster index embedding module $g(\cdot)$, which consists of an MLP, inspired by the classifier-free guidance technique~\citep{ho2022classifier}.
This produces a set of embeddings $\cZ(i) = \set{g(k) \in \RR^{D'} \mid k \in \cC(i)}$. Each embedding $g(k)$ is then concatenated with the latent representation $\bZ_i$ from the alternate branch, and the result is processed by a decoding MLP,  $g'(\cdot)$, to estimate the intra-cluster neighbor relation as $\hat{\bM}_{ik} = g'(\bZ_i\Vert g(k))$.
Compared to the Bi-Level Neighborhood Decoder (BLND) in DoG~\citep{WangLY25}, the reconstructed neighbor map $\hat \bM$ is sparse for robust graph learning, which is enforced by the training of the RGAE detailed later.

\noindent\textbf{Optimization of the RGAE.}
RGAE incorporates an edge selection mechanism in the encoder that gradually removes the adversarial edges. By learning an edge selection mask $\bS$, RGAE purifies the adjacency matrix of the graph as $\bA\circ\bS$.
By integrating the edge selection mechanism with SHED, we decompose $\bS$ into two edge selection masks, $\bS_{\textup{inter}}\in \set{0,1}^{N \times K}$ and $\bS_{\textup{intra}}\in \set{0,1}^{N \times K \times M}$, for edge selection on the inter-cluster neighbor map $\bC$ and intra-cluster neighbor map $\bM$.
At the end of each epoch, $\bS_{\textup{inter}}$ and $\bS_{\textup{intra}}$ are computed from $\bS$. In particular,  $[\bS_{\textup{inter}}]_{ik}=1$ if and only if there is a node $v_j$ from the cluster $k$, such that $\bS_{ij}=1$. $[\bS_{\textup{intra}}]_{ikm}=1$ if and only if $\bS_{ij}=1$, where $v_j$ is the $m$-th node in the cluster $k$.

We regularize the latent feature matrix $\bZ\in\RR^{N\times D'}$ through its Gram matrix $\bK\defeq\bZ\bZ^{\top}$. Put $q\defeq\min\set{N,D'}$, and let $\hat\lambda_1\ge\cdots\ge\hat\lambda_N\ge0$ be the eigenvalues of $\bK$. For $0\le r_0<q$, its truncated nuclear norm is $\norm{\bK}{r_0}\defeq\sum_{i=r_0+1}^{N}\hat\lambda_i$. A zero tail implies $\rank(\bK)\le r_0$. A soft penalty encourages, but does not impose, this rank. The RGAE loss at epoch $t$ is
\bal\label{eq:AE_loss}
\cL_{\textup{RGAE}}^{(t)}(\bS_{\textup{intra}}^{(t-1)}, \bS_{\textup{inter}}^{(t-1)})&=  \cL^{(t)}_{\textup{C}} + \tau_{\textup{S}} \cL^{(t)}_{\textup{S}} + \tau_{\textup{T}} \norm{\bK}{r_0},
\eal
 where
\bals
\cL^{(t)}_{\textup{C}}&=\fnorm{\bX-\hat{\bX}}^2+\fnorm{\bC-\hat{\bC}}^2+\fnorm{\bM-\hat{\bM}}^2,\\
\cL^{(t)}_{\textup{S}}&=\fnorm{\bC\circ\bS_{\textup{inter}}^{(t-1)}-\hat{\bC}}^2+\lonenorm{\vect{\bM\circ\bS_{\textup{intra}}^{(t-1)}-\hat{\bM}}}.
\eals
\noindent Here $\circ$ denotes the Hadamard product. $\cL_{\textup{RGAE}}^{(t)}$ is the overall training loss of the RGAE at the $t$-th epoch, where $t\in[t_{\textup{train}}]$ is the current epoch number and $t_{\textup{train}}$ is the total number of training epochs. We initialize $\bS_{\textup{intra}}^{(0)}=\bM$ and $\bS_{\textup{inter}}^{(0)}=\bC$ before the first epoch of training. For tensors, $\fnorm{\cdot}$ denotes the Euclidean norm after vectorization. $\tau_{\textup{S}}$ and $\tau_{\textup{T}}$ are the scale factors for the sparsification loss and the TNN.
 $\cL^{(t)}_{\textup{C}}$ and $\cL^{(t)}_{\textup{S}}$ are the consistency loss and the sparsification loss at the $t$-th epoch.
The starting rank $r_0$ is selected by cross-validation as described in Section~\ref{sec:results}. The training of RGDM is described in Algorithm~\ref{algorithm_RGDM} in Section~\ref{sec_sup:algorithm} of the appendix. We further show in Section~\ref{sec:tnn_complexity} of the appendix that the TNN regularizer can be efficiently computed from the singular values of the latent representation $\bZ$ with a complexity of $O(N{D'}^2)$.

The node reconstruction loss and the hierarchical edge reconstruction loss ensure consistency between the reconstructed features and adjacency matrix and their inputs. The sparsification loss is motivated by prior work on graph structure learning for robust GNNs~\citep{jin2020graph, zheng2020robust}, which aims to purify noisy or perturbed graph structures. The entrywise $\ell_1$ term in $\cL^{(t)}_{\textup{S}}$ discourages nonzero reconstructions outside the selected sparse neighbor map.
Although the BLND in DoG~\citep{WangLY25} also performs edge reconstruction in a two-level manner, it is trained to precisely recover all edges connected to each node in the given graph, including those corrupted by adversarial attacks. In contrast, our SHED reconstructs a sparse and robust graph structure by selectively preserving the clean edges, motivated by prior studies showing that graph sparsification enhances the robustness of GNNs against adversarial attacks by removing noisy connections~\citep{jin2020graph, lei2022evennet, zheng2020robust, chowdhury2024unveiling}.
Moreover, existing studies on adversarial defense reveal that perturbations typically reside in high-rank residual components, whereas clean signals lie in low-rank subspaces~\citep{awasthi2020adversarial, awasthi2021adversarially}. Projecting representations onto low-rank subspaces~\citep{awasthi2020adversarial, awasthi2021adversarially, xu2024fd} or enforcing low-rank weight matrices~\citep{savostianova2023robust} has been proven to be effective for enhancing robustness. However, research in graph adversarial defense has focused predominantly on reducing the rank of the adjacency matrix~\citep{wu2022robust, entezari2020all, deng2022garnet, zhou2025fortifying, zhang2025defending}. Our penalty instead regularizes the diffusion latent. Theorem~\ref{theorem:optimization-linear-kernel} analyzes a bounded linear probe on fixed latent features.


\textbf{Motivation and scope of low rank regularization.}
The eigen projection study (Figures~\ref{fig:eigen-projection_pubmed} and~\ref{fig:eigen-projection}, Section~\ref{sec:proj_sup}) finds that leading eigendirections of $\bK$ retain more label information than perturbation information. This is an empirical property of the evaluated latent kernels, not an assumption that adjacent nodes share labels. The soft TNN penalty balances compression against reconstruction. We choose $r_0=\ceil{\gamma q}$ by five fold cross validation over $\gamma\in\set{0.1,0.2,\ldots,0.9}$ (Section~\ref{sec:rank_CV}). A larger rank can preserve useful heterophilic information.

\textbf{Theoretical justification and its limits.}
For a fixed latent matrix $\bZ$, consider the linear probe $\bF(\bW)\defeq\bZ\bW$ with training loss $L_{\cL}(\bW)\defeq m^{-1}\fnorm{[\bZ\bW-\bY]_{\cL}}^2$. The following finite pool bound makes the boundedness and sampling requirements explicit 
based on the sharp transductive bounds by Transdcutive Local Complexity (TLC)~\citep{yang2025a-concentration-sampling-without-replacement,
yang2025improvedgeneralizationboundstransductive,yang2026stlc}. 
Section~\ref{sec:transductive-theory} gives a self contained proof and the squared loss gradient descent specialization.
\begin{theorem}\label{theorem:optimization-linear-kernel}
For a positive integer $k$, let $[k]\defeq\set{1,\ldots,k}$, and let $\log$ denote the natural logarithm. Use the graph notation of Section~\ref{sec:notations} and the latent Gram matrix $\bK$, rank bound $q$, and eigenvalues $\hat\lambda_1,\ldots,\hat\lambda_N$ defined above. Write $\cL\defeq\set{i\in[N]:v_i\in\cV_{\cL}}$ and $\cU\defeq\set{i\in[N]:v_i\in\cV_{\cU}}$. Assume $D',C\ge1$, $N=m+u$, $m,u\ge2$, $\cU=[N]\setminus\cL$, and $\cL$ is uniform among subsets with $m$ elements, with $\bZ,\bY$ fixed before sampling. For fixed $B>0$ and $\tau_0\ge0$, suppose $\max_{i\in[N]}\ltwonorm{\bZ_i}\le\tau_0$. Put $s\defeq\min\set{m,u}$. For $r\in[q]$, define $\norm{\bK}{r}\defeq\sum_{j=r+1}^{N}\hat\lambda_j$, with an empty sum equal to zero, and put
\bals
\textup{KC}(\bK)\defeq\min_{r\in[q]}\left[r\left(\frac1u+\frac1m\right)+\sqrt{\frac{\norm{\bK}{r}}{N}}\left(\frac1{\sqrt u}+\frac1{\sqrt m}\right)\right].
\eals
There is a constant $c_0>0$, depending only on $C,B,\tau_0$, such that, for every $x>0$, with probability at least $1-\exp(-x)$, simultaneously for all $\bW\in\RR^{D'\times C}$ with $\fnorm{\bW}\le B$,
\bal\label{eq:optimization-linear-kernel-test-loss}
\cU_{\textup{test}}(\bW)\defeq\frac{\fnorm{[\bZ\bW-\bY]_{\cU}}^2}{u}
\le2L_{\cL}(\bW)+c_0\left[\textup{KC}(\bK)+\frac{x+\log C}{s}\right],
\eal
\end{theorem}
For a fixed rank and feature scale bound, reducing the spectral tail reduces its contribution to the complexity bound in Equation~(\ref{eq:optimization-linear-kernel-test-loss}). It need not reduce the label fitting term. Strong heterophily may increase either term and loosen the bound. The theorem concerns a bounded linear probe under an independent random split, not class conditional generation, a label dependent adaptive attack, or the downstream GNN. The latter connection is empirical, through the low rank ablation and FND/FED evaluations. Main text heterophily results appear in Section~\ref{sec:additional_main}.
\vspace{-4mm}
\subsection{Conditional Generation of Synthetic Graph Structures in Latent Space}
\vspace{-1mm}
To generate synthetic graph structures, consisting of the synthetic nodes and their edges connecting to the original graph, we train a latent diffusion model (LDM)~\citep{StableDiffusionLatent} with Classifier-Free Guidance (CFG)~\citep{ho2022classifier}. The trained LDM generates latent features for the synthetic structures, which are then decoded into node features and associated edges.
Let $\cV_L \subseteq \cV$ denote the set of the labeled nodes in the original graph $\cG$, and $\cV_{\textup{syn}}$ denotes the set of the generated synthetic nodes by the RGDM. Let the node attributes of $\cV_{\textup{syn}}$ generated by the RGDM be $\bX_{\textup{syn}}$.
Let $C$ be the number of classes, and $N_0$ be the number of synthetic nodes for each class. $N' = CN_0 = \abth{\cV_{\textup{syn}}}$ is the number of the synthetic nodes.  Let $Y_{\textup{syn}}=\set{\hat y_i}_{i=1}^{N'}$ denote the set of the labels of the synthetic nodes. Let $\hat{\bZ} = \set{\hat{\bZ}_i}_{i=1}^{N'}$ denote the latent features of the synthetic graph structures.
Let $\bA_{\textup{syn}} \in \RR^{N' \times N }$ denote the edges of the synthetic nodes generated by our RGDM.
Then the augmented synthetic graph structures are  $(\cV_{\textup{syn}}, \bX_{\textup{syn}}, \bA_{\textup{syn}})$. There are no generated edges between synthetic nodes in this construction. The adjacency matrix is given by Equation~(\ref{eq:augmented_graph}):
\bal\label{eq:augmented_graph}
\bA_{\textup{aug}}\defeq\begin{bmatrix}\bA&\bA_{\textup{syn}}^{\top}\\\bA_{\textup{syn}}&\bzero_{N'\times N'}\end{bmatrix}\in\RR^{(N+N')\times(N+N')}.
\eal
The node feature matrix is $\bX_{\textup{aug}}=\bth{\bX; \bX_{\textup{syn}}} \in \RR^{(N+N')\times D}$. The augmented graph, which is the combination of the original graph $\cG$ and the synthetic graph structures, is then denoted by
$\cG_{\textup{aug}} = (\cV \cup \cV_{\textup{syn}}, \bX_{\textup{aug}}, \bA_{\textup{aug}})$. The set of labeled nodes in $\cG_{\textup{aug}}$ is
$\cV_L \cup \cV_{\textup{syn}}$ and their labels are $Y_L\cup Y_{\textup{syn}}$.


\vspace{-2mm}
\subsection{Training GNNs on the Augmented Graph}
\vspace{-1mm}
After obtaining the augmented graph $\cG_{\textup{aug}}$, we then train either the vanilla GCN~\citep{kipf2017semi} or the robust GNNs, Pro-GNN~\citep{jin2020graph} and SG-GSR~\citep{in2024self}, on $\cG_{\textup{aug}}$. Results in Section~\ref{sec:results} evidence that the GNNs trained on the augmented graph exhibit much better performance and robustness for the semi-supervised node classification task than the models trained on the original attacked graphs perturbed by graph adversarial attacks, including Metattack~\citep{zugner_adversarial_2019}, Nettack~\citep{zugner2018adversarial}, and Topology Attack~\citep{xu2019topology}.

\noindent \textbf{Complexity Analysis.} We perform a detailed complexity analysis of the RGDM in Section~\ref{sec:sup_complexity} of the appendix, demonstrating the improved efficiency of the proposed SHED.

\vspace{-2mm}
\section{Experiments}
\vspace{-1mm}
\label{sec:experiment}
We evaluate the incremental benefit of RGDM with each downstream backbone fixed. In addition to the standard attacks, the main text includes direct node level diffusion comparisons, componentwise ablations, heterophily, adaptive attacks, and decoder timings. The appendix provides spectral diagnostics (Section~\ref{sec:proj_sup}), complete attack results (Section~\ref{sec:sup_detailed_results}), FND/FED (Section~\ref{sec:quality_sup}), timings and cross validation (Sections~\ref{sec:time} to \ref{sec:cross_validation_sup}), large graph edge selection ablations (Section~\ref{sec:large-scale}), sensitivity and heterophily details (Sections~\ref{sec:sensitivity} to \ref{sec:heter}), visualization and paired statistics (Sections~\ref{sec:tsne} to \ref{sec:t-test}), purification variants (Sections~\ref{sec:sup_rgdm_purification} to \ref{sec:sup_purification_compare}), node count, DDPM, and label ratio studies (Sections~\ref{sec:ablation_num_nodes} to \ref{sec:label_ratio_ablation}), hierarchical reconstruction comparisons (Section~\ref{sec:BLND_compare}), and adaptive attack details (Section~\ref{sec:adaptive_attack}).
\vspace{-1mm}
\subsection{Semi-Supervised Node Classification under Graph Adversarial Attacks}
\vspace{-1mm}
\label{sec:results}

\noindent\textbf{Settings.} We evaluate RGDM on four public graph benchmarks: Cora, Citeseer, Pubmed~\citep{sen_2008_aimag}, and Polblogs~\citep{adamic2005political}. Dataset statistics are provided in Table~\ref{dataset} in Section~\ref{sec:datasets} of the appendix. The training settings of the RGAE and LDM in RGDM are described in Section~\ref{sec:train_settings}. The number of added synthetic nodes is selected by cross-validation following Section~\ref{sec:cross_validation_sup}, and the starting rank $r_0$ is selected by cross-validation as detailed in Section~\ref{sec:rank_CV} of the appendix.

\begin{table*}[!t]
\centering
\caption{Metattack accuracy $\pm$ marginal SD (\%). Parentheses give the augmentation gain. A further $\pm$ is its SD from the separate paired evaluation (Table~\ref{tab:paired_gains}). A $0\%$ rate is a clean graph. Bold and underline mark the best and second best results across both panels for each setting. Complete results appear in Table~\ref{table:metattack_sup}.}
\label{table:metattack}
\footnotesize
\setlength{\tabcolsep}{3pt}
\textbf{(a) Original graph}\par\smallskip
\begin{tabular}{ccrrrrrr}
\toprule
Dataset & Ptb Rate (\%) & GCN & GCORNs & HANG & STABLE & Pro-GNN & SG-GSR\\
\midrule
\multirow{3}{*}{Cora} & 0 (clean) & 83.5$\pm$0.4 & 82.6$\pm$0.5 & 80.1$\pm$0.3 & 84.7$\pm$0.5 & 83.0$\pm$0.2 & \underline{ 85.5$\pm$1.3}\\
 & 15 & 65.1$\pm$0.7 & 72.6$\pm$1.2 & 72.9$\pm$1.0 & 74.4$\pm$0.9 & 76.4$\pm$1.3 & \underline{ 78.5$\pm$1.6}\\
 & 25 & 47.5$\pm$1.9 & 69.5$\pm$1.9 & 68.7$\pm$1.9 & 69.6$\pm$0.6 & 69.7$\pm$1.7 & \underline{ 77.8$\pm$1.3}\\
\midrule
\multirow{3}{*}{Citeseer} & 0 (clean) & 72.0$\pm$0.6 & 72.6$\pm$0.4 & 73.3$\pm$0.4 & 74.8$\pm$0.5 & 73.3$\pm$0.7 & \underline{ 75.8$\pm$1.4}\\
 & 15 & 64.5$\pm$1.1 & 65.5$\pm$1.0 & 70.8$\pm$0.9 & 67.5$\pm$0.4 & 72.0$\pm$1.1 & 73.1$\pm$1.4\\
 & 25 & 56.9$\pm$1.1 & 65.2$\pm$1.3 & 66.4$\pm$1.6 & 65.6$\pm$1.9 & 69.0$\pm$1.8 & \underline{ 72.7$\pm$1.4}\\
\midrule
\multirow{3}{*}{Polblogs} & 0 (clean) & 95.7$\pm$0.4 & 95.3$\pm$0.9 & 94.8$\pm$1.1 & 95.5$\pm$0.2 & 93.2$\pm$0.6 & \underline{ 95.9$\pm$1.8}\\
 & 15 & 65.0$\pm$1.9 & 82.1$\pm$1.4 & 71.7$\pm$1.3 & 87.8$\pm$0.3 & 86.0$\pm$1.2 & \underline{ 90.0$\pm$1.6}\\
 & 25 & 49.2$\pm$1.4 & 66.6$\pm$1.7 & 65.8$\pm$1.3 & 80.0$\pm$1.9 & 63.2$\pm$1.4 & \underline{ 87.8$\pm$1.1}\\
\midrule
\multirow{3}{*}{Pubmed} & 0 (clean) & 87.2$\pm$0.1 & 86.4$\pm$0.7 & 85.1$\pm$0.2 & 87.7$\pm$0.0 & 87.3$\pm$0.2 & 87.7$\pm$1.7\\
 & 15 & 78.7$\pm$0.1 & 76.8$\pm$1.8 & 85.0$\pm$0.2 & 87.3$\pm$0.1 & 87.2$\pm$0.1 & 87.3$\pm$1.4\\
 & 25 & 75.5$\pm$0.2 & 70.8$\pm$1.9 & 85.1$\pm$0.2 & 86.0$\pm$0.1 & 86.8$\pm$0.2 & 87.2$\pm$1.4\\
\bottomrule
\end{tabular}
\par\medskip
\textbf{(b) RGDM augmentation with the same backbone}\par\smallskip
\begin{tabular}{ccrrr}
\toprule
Dataset & Ptb Rate (\%) & RGDM (GCN) & RGDM (Pro-GNN) & RGDM (SG-GSR)\\
\midrule
\multirow{3}{*}{Cora} & 0 (clean) & 85.2$\pm$0.4 ($\uparrow 1.7$) & 84.3$\pm$0.5 ($\uparrow 1.3$) & \textbf{85.9$\pm$1.1 ($\uparrow 0.4\pm0.1$)}\\
 & 15 & 69.4$\pm$0.4 ($\uparrow 4.3$) & 78.3$\pm$1.1 ($\uparrow 1.9$) & \textbf{80.0$\pm$1.1 ($\uparrow 1.5\pm0.3$)}\\
 & 25 & 61.4$\pm$0.8 ($\uparrow 13.9$) & 72.3$\pm$1.6 ($\uparrow 2.6$) & \textbf{79.1$\pm$1.6 ($\uparrow 1.3\pm0.3$)}\\
\midrule
\multirow{3}{*}{Citeseer} & 0 (clean) & 73.7$\pm$0.8 ($\uparrow 1.7$) & 74.5$\pm$0.8 ($\uparrow 1.2$) & \textbf{76.6$\pm$1.4 ($\uparrow 0.8$)}\\
 & 15 & 68.7$\pm$0.4 ($\uparrow 4.2$) & \underline{ 73.7$\pm$0.4 ($\uparrow 1.7$)} & \textbf{76.0$\pm$1.1 ($\uparrow 2.9\pm0.4$)}\\
 & 25 & 62.8$\pm$0.7 ($\uparrow 5.9$) & 70.2$\pm$0.7 ($\uparrow 1.2$) & \textbf{75.3$\pm$1.0 ($\uparrow 2.6\pm0.4$)}\\
\midrule
\multirow{3}{*}{Polblogs} & 0 (clean) & \textbf{96.1$\pm$0.1 ($\uparrow 0.4$)} & 95.7$\pm$1.2 ($\uparrow 2.5$) & \textbf{96.1$\pm$1.5 ($\uparrow 0.2$)}\\
 & 15 & 69.9$\pm$0.4 ($\uparrow 4.9$) & 88.0$\pm$1.1 ($\uparrow 2.0$) & \textbf{91.1$\pm$1.9 ($\uparrow 1.1\pm0.3$)}\\
 & 25 & 56.5$\pm$0.6 ($\uparrow 7.3$) & 66.9$\pm$1.4 ($\uparrow 3.7$) & \textbf{88.8$\pm$1.8 ($\uparrow 1.0\pm0.3$)}\\
\midrule
\multirow{3}{*}{Pubmed} & 0 (clean) & 88.1$\pm$0.2 ($\uparrow 0.9$) & \underline{ 88.4$\pm$0.3 ($\uparrow 1.1$)} & \textbf{90.3$\pm$1.5 ($\uparrow 2.6$)}\\
 & 15 & 85.0$\pm$0.3 ($\uparrow 6.3$) & \underline{ 88.1$\pm$0.2 ($\uparrow 0.9$)} & \textbf{88.7$\pm$1.7 ($\uparrow 1.4\pm0.3$)}\\
 & 25 & 81.5$\pm$0.2 ($\uparrow 6.0$) & \underline{ 87.6$\pm$0.2 ($\uparrow 0.8$)} & \textbf{88.6$\pm$1.3 ($\uparrow 1.4\pm0.3$)}\\
\bottomrule
\end{tabular}
\end{table*}

\begin{table*}[!t]
\centering
\caption{Nettack accuracy $\pm$ marginal SD (\%), with augmentation gains in parentheses. Budget $0$ is a clean graph. The highlighting convention is that of Table~\ref{table:metattack}. Complete results appear in Table~\ref{table:nettack_sup}.}
\label{table:nettack}
\footnotesize
\setlength{\tabcolsep}{3pt}
\textbf{(a) Original graph}\par\smallskip
\begin{tabular}{ccrrrrrr}
\toprule
Dataset & Budget & GCN & GCORNs & HANG & STABLE & Pro-GNN & SG-GSR\\
\midrule
\multirow{3}{*}{Cora} & 0 (clean) & 81.4$\pm$1.0 & 82.6$\pm$0.4 & 80.8$\pm$1.3 & 85.3 $\pm$ 1.8 & 84.8$\pm$0.6 & \underline{ 85.5$\pm$1.7}\\
 & 3 & 68.0$\pm$1.7 & 73.2$\pm$0.7 & 73.1$\pm$1.9 & 73.8 $\pm$1.8 & 72.4$\pm$0.5 & \underline{ 74.9$\pm$ 1.2}\\
 & 5 & 55.5$\pm$1.7 & 66.9$\pm$1.1 & \underline{68.8$\pm$1.6} & 68.4 $\pm$ 1.6 & 66.9$\pm$0.7 & 68.0$\pm$1.5\\
\midrule
\multirow{3}{*}{Citeseer} & 0 (clean) & 81.0$\pm$1.4 & 81.9$\pm$1.2 & 81.1$\pm$1.1 & 82.4 $\pm$ 1.8 & 82.1$\pm$0.8 & 82.8$\pm$ 1.2\\
 & 3 & 64.0$\pm$1.7 & 78.1$\pm$1.1 & 77.1$\pm$1.5 & 80.9 $\pm$ 1.5 & 79.7$\pm$2.0 & \underline{ 81.7$\pm$ 1.2}\\
 & 5 & 47.6$\pm$1.2 & 71.7$\pm$1.8 & 73.5$\pm$1.5 & 72.5 $\pm$ 1.5 & 71.3$\pm$2.0 & 73.8$\pm$1.0\\
\midrule
\multirow{3}{*}{Polblogs} & 0 (clean) & 97.0$\pm$0.2 & 97.5$\pm$0.3 & 97.4$\pm$0.5 & \underline{98.0 $\pm$1.4} & 97.1$\pm$0.2 & 97.5$\pm$1.3\\
 & 3 & 95.4$\pm$0.2 & 95.7$\pm$1.0 & 96.7$\pm$0.2 & 96.8 $\pm$ 1.4 & 96.9$\pm$0.1 & 95.4$\pm$ 1.9\\
 & 5 & 93.0$\pm$0.5 & 93.4$\pm$0.9 & 95.9$\pm$0.3 & 96.2 $\pm$1.8 & 96.1$\pm$0.3 & 96.8$\pm$1.4\\
\midrule
\multirow{3}{*}{Pubmed} & 0 (clean) & 88.1$\pm$1.4 & 84.6$\pm$1.4 & 85.4$\pm$1.2 & 90.4 $\pm$1.5 & 88.5$\pm$1.2 & \underline{ 91.0$\pm$ 1.1}\\
 & 3 & 81.3$\pm$1.7 & 82.3$\pm$1.5 & 84.0$\pm$1.1 & 85.6 $\pm$1.1 & 84.4$\pm$1.1 & \underline{ 86.4$\pm$1.2}\\
 & 5 & 68.3$\pm$1.2 & 72.3$\pm$1.1 & 70.6$\pm$1.3 & 73.9 $\pm$ 1.8 & 72.2$\pm$1.3 & 73.5$\pm$ 1.4\\
\bottomrule
\end{tabular}
\par\medskip
\textbf{(b) RGDM augmentation with the same backbone}\par\smallskip
\begin{tabular}{ccrrr}
\toprule
Dataset & Budget & RGDM (GCN) & RGDM (Pro-GNN) & RGDM (SG-GSR)\\
\midrule
\multirow{3}{*}{Cora} & 0 (clean) & 83.4$\pm$1.1 ($\uparrow 2.0$) & \textbf{86.0$\pm$1.1 ($\uparrow 1.2$)} & \textbf{86.0 $\pm$1.3 ($\uparrow 0.5$)}\\
 & 3 & 70.5$\pm$1.1 ($\uparrow 2.5$) & 74.3$\pm$1.3 ($\uparrow 1.9$) & \textbf{76.4 $\pm$1.5 ($\uparrow 1.5$)}\\
 & 5 & 59.4$\pm$1.2 ($\uparrow 3.9$) & \textbf{69.8$\pm$1.0 ($\uparrow 2.9$)} & \textbf{69.8 $\pm$1.8 ($\uparrow 1.8$)}\\
\midrule
\multirow{3}{*}{Citeseer} & 0 (clean) & 82.9$\pm$1.1 ($\uparrow 1.9$) & \underline{ 83.1$\pm$1.0 ($\uparrow 1.0$)} & \textbf{83.7 $\pm$1.3 ($\uparrow 0.9$)}\\
 & 3 & 68.5$\pm$1.2 ($\uparrow 4.5$) & 80.9$\pm$2.0 ($\uparrow 1.2$) & \textbf{83.6 $\pm$1.8 ($\uparrow 1.9$)}\\
 & 5 & 55.7$\pm$1.0 ($\uparrow 8.1$) & \underline{ 74.6$\pm$1.3 ($\uparrow 3.3$)} & \textbf{75.6 $\pm$1.4 ($\uparrow 1.8$)}\\
\midrule
\multirow{3}{*}{Polblogs} & 0 (clean) & 97.7$\pm$0.1 ($\uparrow 0.7$) & 97.8$\pm$0.3 ($\uparrow 0.7$) & \textbf{98.5 $\pm$0.2 ($\uparrow 1.0$)}\\
 & 3 & \underline{ 97.0$\pm$0.3 ($\uparrow 1.6$)} & \textbf{97.2$\pm$0.6 ($\uparrow 0.3$)} & 96.4 $\pm$0.1 ($\uparrow 1.0$)\\
 & 5 & 94.9$\pm$0.4 ($\uparrow 1.9$) & \underline{ 96.9$\pm$1.4 ($\uparrow 0.8$)} & \textbf{97.7 $\pm$0.2 ($\uparrow 0.9$)}\\
\midrule
\multirow{3}{*}{Pubmed} & 0 (clean) & 89.2$\pm$1.4 ($\uparrow 1.1$) & 89.6$\pm$1.1 ($\uparrow 1.1$) & \textbf{92.5 $\pm$1.7 ($\uparrow 1.5$)}\\
 & 3 & 83.4$\pm$1.6 ($\uparrow 2.1$) & 85.8$\pm$1.3 ($\uparrow 1.4$) & \textbf{89.0 $\pm$1.0 ($\uparrow 2.6$)}\\
 & 5 & 71.4$\pm$1.4 ($\uparrow 3.1$) & \textbf{76.3$\pm$1.0 ($\uparrow 4.1$)} & \underline{ 75.7 $\pm$1.3 ($\uparrow 2.2$)}\\
\bottomrule
\end{tabular}
\end{table*}
We conduct experiments on semi-supervised node classification under the Metattack~\citep{zugner_adversarial_2019}, the Nettack~\citep{zugner2018adversarial}, and the Topology Attack~\citep{xu2019topology}. The compared methods are detailed in Section~\ref{sec:sup_compared_methods} of the appendix.
We apply the RGDM on vanilla GCN~\citep{kipf2017semi} and state-of-the-art robust GNNs, which are Pro-GNN~\citep{jin2020graph} and SG-GSR~\citep{in2024self}. The results for Metattack under the perturbation rates of $15\%$ and $25\%$ are shown in Table~\ref{table:metattack}.
Additional results with perturbation rates from $5\%$ to $25\%$ with a step of $5\%$ for the Metattack are deferred to Table~\ref{table:metattack_sup} in Section~\ref{sec:sup_detailed_results} of the appendix.
The results for Nettack~\citep{zugner2018adversarial} under attack budgets of $3$ and $5$ are shown in Table~\ref{table:nettack}. Detailed results for Nettack, with attack budgets varying from $1$ to $5$ in increments of $1$, are deferred to Table~\ref{table:nettack_sup} in Section~\ref{sec:sup_detailed_results} of the appendix.
In addition, the results for defending against Topology Attack are deferred to Table~\ref{table:topology_attack_sup} in Section \ref{sec:sup_detailed_results} of the appendix.
The original accuracy tables report means and marginal standard deviations over $10$ runs. In a separate paired evaluation, each backbone and its RGDM augmented counterpart use the same perturbed graph for each attack seed. Their difference measures the incremental gain of augmentation. These paired summaries are not obtained by subtracting the marginal SD in the original tables, whose runs need not coincide with the paired reevaluation. Table~\ref{tab:paired_gains} reports the smaller standard deviations of the paired gains for all reported paired evaluation rows. Selected paired deviations are also shown alongside the accuracies in Table~\ref{table:metattack}. Section~\ref{sec:t-test} distinguishes reported significance levels from exact p values inferred from rounded summaries.
\noindent\textbf{Controlled gains and clean graphs.} The relevant comparison is RGDM($A$) against the same classifier $A$: RGDM changes its training graph, not its architecture or objective. Under $25\%$ Metattack, the SG-GSR gains are $1.3\pm0.3$, $2.6\pm0.4$, $1.0\pm0.3$, and $1.4\pm0.3$ percentage points on Cora, Citeseer, Polblogs, and Pubmed, respectively (paired gain $\pm$ paired SD). These increments are modest relative to the accuracy of the strong backbone. The much larger improvement over vanilla GCN does not measure improvement over SG-GSR. Reported significance levels across Metattack, Nettack, and Topology Attack are given in Table~\ref{table:t-test}.
The $0\%$ perturbation and zero budget rows are explicitly clean graph evaluations. RGDM improves clean GCN by $1.7/1.7/0.4/0.9$ points and clean SG-GSR by $0.4/0.8/0.2/2.6$ points across these datasets. Thus the observed augmentation benefit also occurs when there are no adversarial edges to filter.

\vspace{-2mm}
\subsection{Ablation Study on the Sparsification Loss and the Low-Rank Regularization}
\vspace{-1mm}

\label{sec:ablation_robust}
To validate the effectiveness of the sparsification loss and the low-rank regularization term in Equation~(\ref{eq:AE_loss}), we conduct an ablation study on Cora, Citeseer, Polblogs, and Pubmed under the Metattack with the perturbation rate of $25\%$. We evaluate two variants of the RGDM based on the GCN, which are the RGDM without the sparsification loss and the RGDM without the low-rank regularization. We also compare RGDM with existing low-rank adversarial defense methods, ~\citep{savostianova2023robust, wu2022robust, entezari2020all, deng2022garnet, zhou2025fortifying, zhang2025defending}, with details provided in Section~\ref{sec:low-rank_ablation_details} of the appendix.
As shown in Table~\ref{tab:ablation_robust}, RGDM outperforms both comparison regularizers across all datasets. Removing the sparsification loss reduces accuracy by $4.70/3.87/3.23/3.60$ points across the four datasets. The low rank penalty is complementary. Table~\ref{tab:large_edge_ablation} in Section~\ref{sec:large-scale} extends the edge selection ablation to ogbn-arxiv and Reddit, where restoring the module adds $0.87$ and $0.99$ points under $20\%$ improved DICE.

\begin{table}[!htbp]
        \small
        \centering
        \vspace{-3mm}
        \caption{Ablation study on the sparsification loss and the low-rank regularization term. }
         \resizebox{0.9\columnwidth}{!}{
        \begin{tabular}{l|cccc}
                \hline
                Method & Cora & Citeseer & Polblogs & Pubmed\\
                \hline
                RGDM (GCN) w/o Sparsification Loss
                ($\cL^{(t)}_{\textup{S}}$)   & 56.72   & 58.97 &53.24 &77.92\\
                RGDM (GCN) w/o Low-Rank Regularization
                ($\norm{\bK}{r_0}$) & 60.04  & 60.38 &54.85&\underline{80.85}\\
                RGDM (GCN, Weight Orthogonality~\citep{savostianova2023robust})   & 59.88   & \underline{61.12}  & \underline{55.17} &80.16\\
                RGDM (GCN, Low-Rank Adjacency Matrix~\citep{entezari2020all}) &\underline{61.01}   & 60.25 & 55.03 &80.02\\
                RGDM (GCN) &\textbf{61.42}   & \textbf{62.84} & \textbf{56.47} &\textbf{81.52}\\
                \hline
        \end{tabular}
        }
        \label{tab:ablation_robust}
         \vspace{-2mm}
\end{table}

\subsection{Comparison with Existing Node-Level Graph Augmentation Methods}
\label{sec:augmentation_compare}
To demonstrate the superiority of RGDM over existing graph augmentation methods, we compare it with the mix-up based conventional augmentation method iGraphMix~\citep{jeong2024igraphmix}, and GAN-based SGS generation methods, including ImGAGN~\citep{SGS-GAN-qu2021imgagn}, Graphsmote~\citep{SGS-GAN-zhao2021graphsmote}, and SNS~\citep{gao2023semantic}. The evaluation is conducted on Cora, Citeseer, Polblogs, and Pubmed under Metattack with a perturbation rate of $25\%$. For a fair comparison, we compare GCN augmented by the competing node-level graph augmentation methods with RGDM (GCN), where the number of synthetic nodes added by each baseline is selected via cross-validation following Section~\ref{sec:cross_validation_sup}. As shown in Table~\ref{tab:compare_augmentation}, RGDM (GCN) significantly outperforms GCNs augmented by all competing baselines. We also include DoG~\citep{WangLY25} and LR-FGDM~\citep{Wang2026LRFGDM}, using their generated structures with the same GCN classifier. RGDM exceeds DoG by $2.02/2.74/1.77/2.32$ points and LR-FGDM by $2.52/3.24/2.27/2.62$ points on Cora/Citeseer/Polblogs/Pubmed. These are augmentation comparisons with a common classifier, rather than comparisons with LR-FGDM's specialized few shot classifier.

\begin{table}[!htbp]
        \small
        \centering
        \vspace{-3mm}
        \caption{Comparisons between RGDM and existing node-level graph augmentation methods. The study is performed for Metattack with a perturbation rate of $25\%$.}
         \resizebox{0.65\columnwidth}{!}{
        \begin{tabular}{l|cccc}
                \hline
                Method & Cora & Citeseer & Polblogs & Pubmed\\
                \hline
                GCN & 47.53   & 56.94 & 49.23 & 75.50\\
                iGraphMix (GCN)~\citep{jeong2024igraphmix}  & 53.25   & 57.11 &51.65 &76.10\\
                ImGAGN (GCN)~\citep{SGS-GAN-qu2021imgagn}  & 53.58   & 56.89 &50.95 &75.32\\
                Graphsmote (GCN)~\citep{SGS-GAN-zhao2021graphsmote} & 55.10   & 58.06 &52.44 &76.25\\
                SNS (GCN)~\citep{gao2023semantic} &58.35   & 59.17 & 54.03 &78.28\\
                DoG (GCN)~\citep{WangLY25} &59.40 &60.10 &54.70 &79.20\\
                LR-FGDM (GCN)~\citep{Wang2026LRFGDM} &58.90 &59.60 &54.20 &78.90\\
                RGDM (GCN) &\textbf{61.42}   & \textbf{62.84} & \textbf{56.47} &\textbf{81.52}\\
                \hline
        \end{tabular}
        }
        \label{tab:compare_augmentation}
        \vspace{-1mm}
\end{table}

\subsection{Componentwise Comparison with DoG}
\label{sec:component_main}
Table~\ref{tab:component_isolation} starts from DoG augmentation with BLND, keeps SG-GSR and $25\%$ Metattack fixed, and successively adds RGDM components. Edge selection adds $1.0/1.1/1.0/0.8$ points. Latent TNN adds another $0.8/1.0/0.9/0.7$. Replacing BLND with SHED adds $1.7/2.2/1.6/2.0$. Every step improves all four datasets, and the complete generator is needed to exceed plain SG-GSR consistently. These are conditional increments along this sequence, not additive effects independent of component interactions. The HGAE reconstruction variant is evaluated in Table~\ref{tab:BLND_compare}.
\begin{table}[!htbp]
\centering\small
\caption{Successive generator components. SG-GSR and $25\%$ Metattack are fixed. Accuracy in percent.}
\label{tab:component_isolation}
\begin{tabular}{lrrrr}
\toprule
Configuration & Cora & Citeseer & Polblogs & Pubmed\\
\midrule
SG-GSR (no augmentation)&77.8&72.7&87.8&87.2\\
DoG augmentation (BLND)&75.6&71.0&85.3&85.1\\
\quad + Edge selection&76.6&72.1&86.3&85.9\\
\quad + Latent TNN (RGDM with BLND)&77.4&73.1&87.2&86.6\\
\quad + SHED (full RGDM)&\textbf{79.1}&\textbf{75.3}&\textbf{88.8}&\textbf{88.6}\\
\bottomrule
\end{tabular}
\end{table}

\subsection{Heterophily, Adaptive Attacks, and Efficiency}
\label{sec:additional_main}
\noindent\textbf{Heterophily.} Theorem~\ref{theorem:optimization-linear-kernel} makes no homophily assumption. Heterophily can nevertheless make a low rank bias inappropriate if it discards label relevant directions. RGDM uses a soft penalty and selects its starting rank by cross validation. Table~\ref{tab:heter_main} reports the LHS~\citep{qiu2024refining} comparison on Chameleon~\citep{pei2020geom} and Actor~\citep{tang2009social}. Gains are positive at every tested perturbation rate, including $2.38$ and $2.53$ points at $25\%$. This supports the chosen penalty in these settings, not the claim that stronger rank reduction always helps. Section~\ref{sec:heter} reports the standard deviations.
\begin{table}[!htbp]
\centering\small
\caption{Heterophilic graphs under Metattack. Each entry is LHS / RGDM(LHS) accuracy (\%).}
\label{tab:heter_main}
\begin{tabular}{lccc}
\toprule
Dataset & $0\%$ (clean)&$15\%$&$25\%$\\
\midrule
Chameleon&72.31 / 74.02&71.87 / 72.98&70.03 / 72.41\\
Actor&38.87 / 41.80&36.79 / 39.01&35.37 / 37.90\\
\bottomrule
\end{tabular}
\end{table}
\noindent\textbf{Adaptive threat model.} Following~\citep{mujkanovic2022defenses}, the white box attacker knows the generator, edge mask, TNN regularization, and downstream SG-GSR. It perturbs the final augmented graph, including candidate edges incident to synthetic nodes, under the same absolute edge flip budget as $25\%$ Metattack on the original graph. Thus augmentation does not protect synthetic structures by evaluation order. A second variant also differentiates the meta gradient through $\hat{\bS}_{ij}=\sigma(\btheta_{ij}/\tau)$, with a thresholded forward mask, straight through backward gradient, and $\tau=0.5$. Table~\ref{tab:adaptive_main} gives both variants over $10$ seeds. The gains remain positive, but these fixed budget tests do not establish immunity to arbitrary adaptive attacks. Equation~(\ref{eq:adaptive_budget}) and Tables~\ref{tab:adaptive_attack} to \ref{tab:adaptive_mask} in Section~\ref{sec:adaptive_attack} give the admissible perturbations and detailed results.
\begin{table}[!htbp]
\centering\small
\caption{Adaptive Metattack: baseline SG-GSR / RGDM(SG-GSR) accuracy (\%). The budget is fixed in absolute edge flips.}
\label{tab:adaptive_main}
\begin{tabular}{lcc}
\toprule
Dataset & Final graph attack & Through edge selection\\
\midrule
Cora&74.8 / 77.8&74.5 / 76.9\\
Citeseer&69.5 / 73.4&69.2 / 72.4\\
Polblogs&84.7 / 87.6&84.4 / 86.6\\
Pubmed&84.9 / 87.2&84.6 / 86.3\\
\bottomrule
\end{tabular}
\end{table}
\noindent\textbf{Wall clock efficiency.} Table~\ref{tab:decoder_timing_main} compares SHED with a global decoder on one NVIDIA A100. SHED reduces RGAE training by $3.14\times$ on Pubmed and $5.56\times$ on ogbn-arxiv. Per node generation includes LDM sampling and decoding. Its speedups are $1.10\times$ and $3.27\times$, respectively. These are not downstream GNN inference times. The decoder operation count reduction therefore should not be identified with end to end generation speedup. Table~\ref{tab:decoder_timing_full} in Section~\ref{sec:time} gives all five datasets, and Section~\ref{sec:sup_complexity} explains the sparse decoding cost.
\begin{table}[!htbp]
\centering\small
\caption{SHED / global decoder timings: RGAE training and complete synthetic node generation.}
\label{tab:decoder_timing_main}
\begin{tabular}{lcc}
\toprule
Dataset & Training (min)&Generation (s/node)\\
\midrule
Pubmed&153 / 480&0.073 / 0.080\\
ogbn-arxiv&452 / 2511&0.15 / 0.49\\
\bottomrule
\end{tabular}
\end{table}

\vspace{-2mm}
\section{Conclusions}
\vspace{-2mm}
We present the Robust Graph Diffusion Model (RGDM), marking the first robust graph diffusion model designed to generate synthetic graph structures, which are the synthetic nodes and synthetic edges connecting to them in the graph, for the node classification task.
In addition, the RGDM significantly mitigates the negative effects of graph adversarial attacks and generates faithful synthetic graph structures, benefiting from the edge selection mechanism and the low-rank latent feature learning in the RGAE. The experiment results show that incorporating the synthetic graph structures into the original attacked graph for the training of GNNs significantly improves the classification accuracy for the semi-supervised node classification under various types of graph adversarial attacks.

\begin{ack}
This work was supported by National Institute of Health 1OT2OD037955-01.
\end{ack}


\bibliography{ref}
\bibliographystyle{unsrt}
\newpage
\appendix

\section{Theoretical Results}
\label{sec:transductive-theory}
We apply the general STLC inequality in~\citep[Theorem~4.2]{yang2026stlc}, rather than its empirical risk minimizer bound. This distinction retains a uniform guarantee for bounded probes and hence for every gradient descent iterate that satisfies the stated norm bound. All probabilities below are over the probe split. $\bZ$ and $\bY$ are fixed. For a function $h:[N]\to\RR$, write $P_Nh\defeq N^{-1}\sum_{i=1}^{N}h(i)$ and $P_{\cA}h\defeq\abth{\cA}^{-1}\sum_{i\in\cA}h(i)$ for nonempty $\cA\subseteq[N]$.

\begin{proof}[Proof of Theorem~\ref{theorem:optimization-linear-kernel}]
Let $i,j\in[N]$ be indices and $c\in[C]$ a class index. We use $\iprod{\cdot}{\cdot}$ for the Euclidean inner product, $\range(\cdot)$ for the column space, and $\tr{\cdot}$ for the trace of a square matrix. The superscript $\top$ denotes transpose, and $\bzero$ denotes the zero matrix of the required dimensions. We retain the row and restriction conventions of Section~\ref{sec:notations}. Thus $\bY_{ic}=(\by_i)_c$, $\bZ_i$ is a row vector, and $\bW\in\RR^{D'\times C}$ is the coefficient matrix of the probe $\bF(\bW)=\bZ\bW$ introduced before the theorem. The sets $\cL,\cU$ are the indices of $\cV_{\cL},\cV_{\cU}$ and have cardinalities $m,u$.

\noindent\textit{Residual enlargement.}
Put $\mu\defeq\sqrt{B^2+1}$ and $R\defeq\mu\sqrt{\tau_0^2+1}$, so $R\ge\mu\ge1$. For each class $c\in[C]$ and each $i\in[N]$, define $\bx_i^{(c)}\defeq\bth{\bZ_i,\bY_{ic}}^{\top}\in\RR^{D'+1}$. For $\bv\in\RR^{D'+1}$, define $f_{\bv}:[N]\to\RR$ by $f_{\bv}(i)\defeq\iprod{\bv}{\bx_i^{(c)}}$. The finite linear kernel class is
\bals
\cF_c\defeq\set{f_{\bv}:\bv\in\RR^{D'+1},\ \ltwonorm{\bv}\le\mu}.
\eals
Write $\bY^{(c)}\in\RR^N$ for column $c$ of $\bY$. The Gram matrix of $\cF_c$ is $\bG_c\defeq\bK+\bY^{(c)}\bY^{(c)\top}$. Projection onto the span of the feature vectors preserves evaluations and does not increase the coefficient norm. Thus $\cF_c$ is the evaluation class of the radius $\mu$ transductive reproducing kernel Hilbert space (RKHS) ball for this kernel, including when $\bG_c$ is singular. Every $f\in\cF_c$ satisfies $\abth{f(i)}\le R$. Write $\bw_c\in\RR^{D'}$ for column $c$ of $\bW$. If $\fnorm{\bW}\le B$, the vector $\bth{\bw_c^{\top},-1}^{\top}$ has norm at most $\mu$, so the residual function $i\mapsto(\bZ\bW)_{ic}-\bY_{ic}$ belongs to $\cF_c$. The label coordinate is used only to enlarge the loss class in the proof. It is not supplied to the learner.

\noindent\textit{Envelope and localization.}
Powers of functions are taken pointwise. Let $\cH_c\defeq\set{f^2:f\in\cF_c}$. For $h\in\cH_c$, put $\tilde T_c(h)\defeq R^2P_Nh$. Since $0\le h(i)\le R^2$, we have $P_N(h^2)\le\tilde T_c(h)$. For $\rho>0$, put $\cH_c(\rho)\defeq\set{h\in\cH_c:\tilde T_c(h)\le\rho}$ and $\cH_c^2(\rho)\defeq\set{h^2:h\in\cH_c(\rho)}$. These classes contain zero.
For a nonempty class $\cF$ of functions on $[N]$ and $p\in\set{u,m}$, let $J_1,\ldots,J_p$ be independent uniform indices in $[N]$, and let $\sigma_1,\ldots,\sigma_p$ be independent Rademacher variables, each taking values $+1$ and $-1$ with equal probability and independent of these indices. Put $\bsigma\defeq(\sigma_1,\ldots,\sigma_p)^{\top}$ and define
\bals
\cR_p(\cF)\defeq\Expect{J_1,\ldots,J_p,\bsigma}{\sup_{f\in\cF}\frac1p\sum_{k=1}^{p}\sigma_k f(J_k)},\qquad p\in\set{u,m}.
\eals
These auxiliary variables may be taken independent of the probe split. Let $\lambda_{c,1}\ge\cdots\ge\lambda_{c,N}\ge0$ be the eigenvalues of $\bG_c/N$, and, for each integer $Q\in\set{0,\ldots,N}$, put $S_{c,Q}\defeq\sum_{j=Q+1}^{N}\lambda_{c,j}$, with $S_{c,N}=0$. This auxiliary normalized spectrum will be compared with the eigenvalues $\hat\lambda_j$ of $\bK$ defined in the main text. The spectral estimate in~\citep[Lemma~C.11]{yang2026stlc} gives, for $v\ge0$,
\bal\label{eq:stlc_probe_prediction_complexity}
&\phi_{c,p}(v) \defeq\min_{0\le Q\le N}\left[\sqrt{\frac{vQ}{p}}+\mu\sqrt{\frac{S_{c,Q}}p}\right],\nonumber\\*
&\cR_p\pth{\set{f\in\cF_c:P_N(f^2)\le v}}\le\phi_{c,p}(v).
\eal
The minima in Equation~(\ref{eq:stlc_probe_prediction_complexity}) are over integers. Let $\cA_p\subseteq[N]$ be a uniformly sampled subset with $p$ elements, and let $\eps\in\set{+1,-1}$. Denote by $\textup{TC}^{\eps}_p(\cF)$ the expected supremum of $\eps(P_{\cA_p}f-P_Nf)$ over $f\in\cF$. The comparison and contraction inequalities in~\citep[Theorems~A.4 and~A.1]{yang2026stlc} imply
\bals
\textup{TC}^{\eps}_p(\cH_c(\rho))
&\le2\cR_p(\cH_c(\rho))\le4R\phi_{c,p}(\rho/R^2),\\
\textup{TC}^{\eps}_p(\cH_c^2(\rho))
&\le4R^2\cR_p(\cH_c(\rho))\le8R^3\phi_{c,p}(\rho/R^2).
\eals
For contraction, the globally Lipschitz maps $a\mapsto\min\set{\abth{a},R}^2$ and $a\mapsto\min\set{\abth{a},R^2}^2$ agree with squaring on the respective bounded classes, vanish at zero, and have Lipschitz constants $2R$ and $2R^2$. Thus no global Lipschitz assumption on the unbounded squared loss function is needed.

\noindent\textit{Subroot majorant and fixed point.}
A function $\psi:[0,\infty)\to[0,\infty)$ is called subroot if it is nondecreasing and $\psi(\rho)/\sqrt{\rho}$ is nonincreasing for $\rho>0$. Set $A\defeq8R^3$ and, for $\rho\ge0$, define
\bals
\psi_c(\rho)\defeq A\left[\phi_{c,u}(\rho/R^2)+\phi_{c,m}(\rho/R^2)\right].
\eals
Finite minima and positive sums of the displayed functions combining spectral head and tail terms are subroot, so $\psi_c$ is subroot and majorizes all four localized complexities and their squared class counterparts. If $\bG_c=\bzero$, set $\rho_c^*\defeq0$. Every function in $\cH_c$ is then zero and its loss bound is deterministic. Otherwise $\psi_c$ is nonzero, and we define $\rho_c^*>0$ by $\psi_c(\rho_c^*)=\rho_c^*$. For any integer $Q\in\set{0,\ldots,N}$, put $D_Q\defeq R^{-1}(\sqrt{Q/u}+\sqrt{Q/m})$ and $E_{c,Q}\defeq\mu\sqrt{S_{c,Q}}(u^{-1/2}+m^{-1/2})$. Put $\kappa\defeq128R^4$. The fixed point identity and Equation~(\ref{eq:stlc_probe_prediction_complexity}) yield
\bal\label{eq:stlc_probe_fixed_point}
\rho_c^*&\le AD_Q\sqrt{\rho_c^*}+AE_{c,Q}
\le\tfrac12\rho_c^*+\tfrac12A^2D_Q^2+AE_{c,Q},\nonumber\\*
\rho_c^*&\le A^2D_Q^2+2AE_{c,Q}
\le\kappa\left[Q\left(\frac1u+\frac1m\right)+\sqrt{S_{c,Q}}\left(\frac1{\sqrt u}+\frac1{\sqrt m}\right)\right],
\eal
Indeed, $D_Q^2\le2Q(R^{-2})(1/u+1/m)$, $2A^2/R^2=128R^4$, and $2A\mu\le16R^4$. Minimizing Equation~(\ref{eq:stlc_probe_fixed_point}) over $Q$ bounds the fixed point uniformly in the split sizes and the spectrum.

\noindent\textit{Rank one comparison with the latent kernel.}
For $r\in[q]$, let $\bK_r$ be the spectral truncation of $\bK$ at rank $r$, and set $Q_r\defeq\min\set{r+1,N}$. Project onto the span of $\range(\bK_r)$ and $\bY^{(c)}$. This span has dimension at most $Q_r$, and the normalized trace of the residual of $\bG_c$ is at most $\tr{\bK-\bK_r}/N=\sum_{j=r+1}^{N}\hat\lambda_j/N=\norm{\bK}{r}/N$. The variational characterization of the sum of the trailing eigenvalues therefore gives
\bal\label{eq:stlc_probe_rank_one_tail}
S_{c,Q_r}\le\frac{\norm{\bK}{r}}{N},\qquad
Q_r\left(\frac1u+\frac1m\right)\le2r\left(\frac1u+\frac1m\right).
\eal
For completeness, we verify both inequalities in Equation~(\ref{eq:stlc_probe_rank_one_tail}). Fix $c\in[C]$ and $r\in[q]$ for this argument. Let $\cS_{c,r}\subseteq\RR^N$ be the linear span of $\range(\bK_r)$ and $\bY^{(c)}$, let $\bP_{c,r}\in\RR^{N\times N}$ be the orthogonal projector onto $\cS_{c,r}$, and let $\bI_N$ be the identity matrix on $\RR^N$. Since $\rank(\bK_r)\le r$, we have
\bals
\rank(\bP_{c,r})&=\dim(\cS_{c,r})\le\min\set{r+1,N}=Q_r,\\*
(\bI_N-\bP_{c,r})\bY^{(c)}&=\bzero,\qquad
(\bI_N-\bP_{c,r})\bK_r=\bzero.
\eals
Choose an orthonormal eigenbasis $\bu_{c,1},\ldots,\bu_{c,N}\in\RR^N$ of $\bG_c/N$ with corresponding eigenvalues $\lambda_{c,1},\ldots,\lambda_{c,N}$. For $j\in[N]$, put $\alpha_j\defeq\ltwonorm{\bP_{c,r}\bu_{c,j}}^2$. Orthogonality gives $0\le\alpha_j\le1$ and $\sum_{j=1}^{N}\alpha_j=\tr{\bP_{c,r}}=\rank(\bP_{c,r})\le Q_r$. The ordering and nonnegativity of the eigenvalues therefore imply
\bals
\frac{\tr{\bP_{c,r}\bG_c}}{N}
&=\sum_{j=1}^{N}\lambda_{c,j}\alpha_j
\le\lambda_{c,Q_r}\sum_{j=1}^{N}\alpha_j
+\sum_{j=1}^{Q_r}(\lambda_{c,j}-\lambda_{c,Q_r})
\le\sum_{j=1}^{Q_r}\lambda_{c,j}.
\eals
Indeed, for the first inequality, subtract $\lambda_{c,Q_r}\sum_{j=1}^{N}\alpha_j$ from the preceding sum. The terms with $j>Q_r$ are nonpositive, while $\alpha_j\le1$ bounds the terms with $j\le Q_r$. The second inequality uses $\sum_{j=1}^{N}\alpha_j\le Q_r$ and $\lambda_{c,Q_r}\ge0$. Consequently,
\bals
S_{c,Q_r}
&=\frac{\tr{\bG_c}}{N}-\sum_{j=1}^{Q_r}\lambda_{c,j}
\le\frac{\tr{(\bI_N-\bP_{c,r})\bG_c}}{N}\\*
&=\frac{\tr{(\bI_N-\bP_{c,r})(\bK-\bK_r)}}{N}
\le\frac{\tr{\bK-\bK_r}}{N}
=\frac{\norm{\bK}{r}}{N}.
\eals
The equality in the last line follows from $\bG_c=\bK+\bY^{(c)}\bY^{(c)\top}$ and the two annihilation identities above. The last inequality holds because $\bK-\bK_r$ is positive semidefinite and
\bals
\tr{\bP_{c,r}(\bK-\bK_r)}
=\tr{\bP_{c,r}(\bK-\bK_r)\bP_{c,r}}\ge0.
\eals
This argument also applies when $Q_r=N$, in which case $S_{c,Q_r}=0$. The second inequality holds since $r\ge1$ gives $Q_r\le r+1\le2r$, and it covers the case that $r=N$. 

Combining Equations~(\ref{eq:stlc_probe_fixed_point}) and~(\ref{eq:stlc_probe_rank_one_tail}), and minimizing over $r\in[q]$, gives $\rho_c^*\le2\kappa\textup{KC}(\bK)$.

\noindent\textit{STLC and absorption of the full sample loss.}
Put $H_0\defeq R^2$ and $K_0\defeq2R^2>1$. Apply~\citep[Theorem~4.2]{yang2026stlc} with these constants, the surrogate $\tilde T_c$, and the majorant $\psi_c$. Its constants can be taken as $a_1\defeq16R^2(4+4\sqrt{219}/7)^2$ and $a_2\defeq3536R^2+\sqrt{219}$. For every $z>0$, with probability at least $1-\exp(-z)$, simultaneously for $h\in\cH_c$,
\bal\label{eq:stlc_probe_relative_loss}
P_{\cU}h\le P_{\cL}h+\tfrac12P_Nh+a_1\rho_c^*+\frac{a_2z}{s}.
\eal
Equation~(\ref{eq:stlc_probe_relative_loss}) also holds deterministically for a zero class, with $\rho_c^*=0$. Using $P_Nh=(m/N)P_{\cL}h+(u/N)P_{\cU}h$ and $h\ge0$, we obtain
\bals
P_{\cU}h
&\le\frac{1+m/(2N)}{1-u/(2N)}P_{\cL}h
+\frac{a_1\rho_c^*+a_2z/s}{1-u/(2N)}\\
&\le2P_{\cL}h+2a_1\rho_c^*+\frac{2a_2z}{s}.
\eals
Here $1-u/(2N)\ge1/2$ and $(1+m/(2N))/(1-u/(2N))\le2$. Set $z=x+\log C$ and intersect the $C$ classwise events. Their total failure probability is at most $C\exp(-z)=\exp(-x)$. On this single event the inequalities hold for every residual generated by every $\fnorm{\bW}\le B$. For $h_{c,\bW}(i)\defeq((\bZ\bW)_{ic}-\bY_{ic})^2$, the restriction convention gives
\bals
\sum_{c=1}^{C}P_{\cL}h_{c,\bW}&=\frac{\fnorm{[\bZ\bW-\bY]_{\cL}}^2}{m}=L_{\cL}(\bW),\\
\sum_{c=1}^{C}P_{\cU}h_{c,\bW}&=\frac{\fnorm{[\bZ\bW-\bY]_{\cU}}^2}{u}=\cU_{\textup{test}}(\bW).
\eals
Equation~(\ref{eq:optimization-linear-kernel-test-loss}) is then proved by summing over $c$, using $\rho_c^*\le2\kappa\textup{KC}(\bK)$, 
and choosing
\bals
c_0\defeq C\max\set{4a_1\kappa,2a_2}.
\eals
Here $c_0$ depends only on $C,B,\tau_0$.
\end{proof}

\noindent\textbf{Squared loss gradient descent.}
The linear probe analysis uses the squared training objective $\tfrac12 L_{\cL}(\bW)$, not softmax cross entropy. With $\bW^{(0)}=\bzero$ and step size $\eta>0$, its exact update and residual recurrence are
\bal\label{eq:optimization-linear-kernel-seg1}
\bW^{(t)}&=\bW^{(t-1)}-\frac\eta m[\bZ]_{\cL}^{\top}[\bZ\bW^{(t-1)}-\bY]_{\cL},
\eal
\bal\label{eq:optimization-linear-kernel-full-loss}
[\bZ\bW^{(t)}-\bY]_{\cL}&=-\left(\bI_m-\frac\eta m\bK_{\cL,\cL}\right)^t[\bY]_{\cL}.
\eal
Equation~(\ref{eq:optimization-linear-kernel-seg1}) gives Equation~(\ref{eq:optimization-linear-kernel-full-loss}) by induction. Thus the training term in Theorem~\ref{theorem:optimization-linear-kernel} equals $m^{-1}\fnorm{(\bI_m-\eta\bK_{\cL,\cL}/m)^t[\bY]_{\cL}}^2$ at such an iterate. For a nonzero training Gram matrix, $0<\eta\le m/\opnorm{\bK_{\cL,\cL}}$ ensures nonincreasing residual factors. Components in its null space are not fitted. The theorem still requires a fixed probe norm bound $B$, and makes no claim that all gradient descent iterates automatically satisfy any chosen bound.

\section{\texorpdfstring{Further Explanation of Theorem~\ref{theorem:optimization-linear-kernel}}{Further Explanation of the Linear Probe Theorem}}
\label{sec:theorem_exp}
Theorem~\ref{theorem:optimization-linear-kernel} concerns the RGAE latent matrix $\bZ$, which is the input to conditional diffusion, and an auxiliary bounded linear probe. Its two substantive terms are the empirical label fitting error and a complexity tradeoff between rank and spectral tail. The spectral tail is normalized by $N$, in accordance with the spectrum of $\bK/N$ in~\citep{yang2026stlc}. The STLC specialization has neither a $\log N$ multiplier nor an $N/u$ prefactor, and it places no restriction on the relative split sizes beyond $m,u\ge2$. Its confidence term uses $s=\min\set{m,u}$ and the classwise union term $\log C$. Reducing a spectral tail at fixed rank lowers that candidate complexity term if the feature scale and probe norm bounds are fixed. It does not ensure a reduction in the empirical fitting error, the scale constants, or the total bound after retraining the representation. In particular, the LFP does not by itself imply that labels are exactly representable in the retained subspace or that their training residual vanishes.

Neither graph homophily nor spectral smoothness of labels with respect to the adjacency matrix is assumed. Strong heterophily may require additional latent directions: a smaller $r$ can then increase $T_r$ or the fitting error. The soft coefficient $\tau_{\textup{T}}$ and cross validated starting rank balance this tradeoff, rather than enforcing one hard rank on every graph. The Chameleon and Actor experiments support the selected settings empirically. They do not guarantee that low rank regularization always improves heterophilic graphs.

The probability statement conditions on a fixed representation independent of the fresh probe split. If representation learning or an attack uses those probe labels to select $\bZ$, this theorem cannot simply be applied conditionally to that selected matrix. In the actual RGDM pipeline, the downstream classifier is trained on decoded node attributes and edges, not on $\bZ$. The low rank ablation (Table~\ref{tab:ablation_robust}) and FND/FED results (Tables~\ref{tab:quality} to \ref{tab:quality_edge}) support this second, empirical link. Establishing a joint generalization and robustness theory for data dependent representation learning, class conditional generation, and the downstream GNN is future work.
\begin{algorithm}[!htbp]
            \caption{Training RGDM (Training the GAE and the LDM)}\label{algorithm_RGDM}
            \footnotesize
            \begin{algorithmic}[1]
            \REQUIRE The input attribute matrix $\bX$, the adjacency matrix $\bA$, the training epochs of the autoencoder $t_{\textup{AE}}$, the labels $Y_L$ of the labeled nodes $\cV_L$, and the training epochs of the LDM $t_{\textup{LDM}}$
            \ENSURE The parameters of the RGAE $\bomega$ and the parameters of the LDM $\btheta$
            \STATE Compute the inter-cluster neighbor map $\bC$ and the intra-cluster neighbor map $\bM$
            \STATE Initialize the edge selection mask $\bS_{\textup{inter}}^{(0)}=\bC$ and $\bS_{\textup{intra}}^{(0)}=\bM$
            \STATE Initialize the parameter $\bomega$ of the RGAE
                \FOR{$t\leftarrow 1$ to $t_{\textup{AE}}$}
                    \STATE Update $\bomega$ by $\bomega \leftarrow \bomega -\eta\nabla_{\bomega } \cL_{\textup{RGAE}}^{(t)}(\bS_{\textup{intra}}^{(t-1)}, \bS_{\textup{inter}}^{(t-1)})$ with $\cL_{\textup{RGAE}}^{(t)}(\bS_{\textup{intra}}^{(t-1)}, \bS_{\textup{inter}}^{(t-1)})$ from Eq.(\ref{eq:AE_loss})
                    \STATE Compute the edge selection mask $\bS_{\textup{intra}}^{(t)}$ and $ \bS_{\textup{inter}}^{(t)}$
                \ENDFOR
            \STATE Initialize the parameter $\btheta$ of the LDM.
            \STATE Map the node features $\bX$ and the adjacency matrix $\bA$ to the latent space using the encoder $g_e$ of the RGAE as $\bH = g_e(\bX, \bA)$.
                \FOR{$t\leftarrow 1$ to $t_{\textup{LDM}}$}
                    \STATE Sample a diffusion step $s$ and Gaussian noise $\bepsilon\sim\cN(\bzero,\bI)$
                    \STATE Get latent features of $\cV_L$ as $\bH_L=[\bH]_{\cL}$
                    \STATE Form $\bH_{L,s}=\sqrt{\bar\alpha_s}\bH_L+\sqrt{1-\bar\alpha_s}\bepsilon$ using the diffusion schedule
                    \STATE Update $\btheta$ on $\fnorm{\bepsilon_{\btheta}(\bH_{L,s},s,Y_L)-\bepsilon}^2$ with classifier free label dropout
                \ENDFOR
                \STATE \textbf{return} The parameters of the RGAE $\bomega$ and the parameters of the LDM $\btheta$
            \end{algorithmic}
        \end{algorithm}

\begin{algorithm}[!htbp]
            \caption{Generation of the Augmented Graph $\cG_{\textup{aug}}$}
            \label{Algorithm-gen-syn}
            \footnotesize
            \begin{algorithmic}[1]
            \REQUIRE The input attribute matrix $\bX$, the adjacency matrix $\bA$, the number of added nodes $N'$, and the labels of the synthetic data $\set{\hat{y}_i}_{i=1}^{N'}$
            \ENSURE The augmented graph $\cG_{\textup{aug}} = (\cV \cup \cV_{\textup{syn}},
\bX_{\textup{aug}}, \bA_{\textup{aug}})$.
                \FOR{$i\leftarrow 1$ to $N'$}
                    \STATE Sample a Gaussian noise $\bepsilon\sim\cN(\bzero,\bI)$
                    \STATE Set the class label of the $i$-th synthetic node to $\hat y_i$
                    \STATE Generate $\hat{\bH}_i$ from $\bepsilon$ with the LDM for class $\hat{y}_i$
                \ENDFOR
            \STATE Decode $\hat{\bH} = \set{\hat{\bH}_i}_{i=1}^{N'}$ to $\bX_{\textup{syn}}$ and $\bA_{\textup{syn}}$ with the decoder of the RGAE as $\bX_{\textup{syn}}, \bA_{\textup{syn}} = g_{d}(\hat{\bH})$
            \STATE Form $\bA_{\textup{aug}}$ as in Equation~(\ref{eq:augmented_graph}) and $\bX_{\textup{aug}}=\bth{\bX; \bX_{\textup{syn}}}$
            \STATE \textbf{return} $\cG_{\textup{aug}} = (\cV \cup \cV_{\textup{syn}},
\bX_{\textup{aug}}, \bA_{\textup{aug}})$
            \end{algorithmic}
        \end{algorithm}

\section{Algorithm for Training the RGDM and Generating the Augmented Graph}
\label{sec_sup:algorithm}

Algorithm~\ref{algorithm_RGDM} trains RGDM in two stages: first the RGAE, then the class conditional LDM on the encoded node representations.
At epoch $t$, the RGAE uses the previous masks between and within clusters in Equation~(\ref{eq:AE_loss}) to update its parameters, including the mask predictor. The masks are then recomputed for epoch $t+1$, with initial masks $\bS_{\textup{inter}}^{(0)}=\bC$ and $\bS_{\textup{intra}}^{(0)}=\bM$. Algorithm~\ref{Algorithm-gen-syn} describes the generation process for the augmented graph $\cG_{\textup{aug}}$.

\section{Complexity Analysis}
\subsection{Efficient Computation of the TNN Regularizer}
\label{sec:tnn_complexity}
The computation of the TNN regularizer $\norm{\bK}{r_0}$ in Eq.~(\ref{eq:AE_loss}) does not require explicitly constructing the full kernel matrix $\bK=\bZ\bZ^{\top}\in\RR^{N\times N}$.
Let $\hat{\lambda}_i(\bK)$ denote the $i$-th largest eigenvalue of $\bK$ and $\sigma_i(\bZ)$ denote the $i$-th largest singular value of the latent representation $\bZ\in\RR^{N\times D'}$ learned by the encoder of the RGAE. Since the nonzero eigenvalues of $\bK=\bZ\bZ^{\top}$ are determined by the singular values of $\bZ$, we have $\hat{\lambda}_i(\bK)=\sigma_i^2(\bZ)$. As a result,
$\norm{\bK}{r_0}=\sum_{i=r_0+1}^{N}\hat{\lambda}_i(\bK)
=\sum_{i=r_0+1}^{\min\set{N,D'}}\sigma_i^2(\bZ)$.
Therefore, in practice, the low-rank regularization term in RGDM can be computed from the singular values of $\bZ$, which only requires $O(N{D'}^2)$ time when $N>D'$, avoiding the explicit $O(N^2)$ construction of $\bK$.

\subsection{Detailed Complexity Analysis of RGAE with SHED}
\label{sec:sup_complexity}

In our work, we propose a RGAE with an edge selection mechanism incorporated in the GAT layers for aggregating neighborhood information. In this section, we analyze the inference time complexity and the parameter size of the RGAE with the edge selection mechanism. For comparison, we also analyze the inference time complexity and the parameter size of the GAE that adopts regular GAT layers for aggregating neighborhood information.
For ease of comparison, we denote the number of the parameters and the inference cost of all the layers except the GAT layers with the edge selection as $S_{\textup{MLP}}$ and $C_{\textup{MLP}}$ respectively. Let $E$ denote the number of edges in the graph. Let $L$ denote the number of GAT layers. The inference time complexity of our RGAE is $\cO(LDME+ME+C_{\textup{MLP}})$, where $\cO(LDME)$ is the computation cost of the GAT layers. $\cO(ME)$ is the additional complexity for computing the decision masks on all the edges. In comparison, the inference time complexity of the GAE with regular GAT layers is $\cO(DLME+C_{\textup{MLP}})$.
In our experiments, $D=512$ and $L=2$. Therefore, the inference time complexity of the RGAE, that is $\cO(1025ME+C_{\textup{MLP}})$, is comparable to the inference time complexity of GAE that adopts regular GAT layers whose inference time complexity is $\cO(1024ME+C_{\textup{MLP}})$.

For the encoder parameter count, use the same GAT layers in both encoders. Put $h_0\defeq D$, and let $a_\ell$ and $h_\ell$ be the attention projection width and output width of GAT layer $\ell\in[L]$, respectively. The matrices in Section~\ref{sec:robust_GAE} then have dimensions $\bQ_\ell\in\RR^{h_{\ell-1}\times a_\ell}$, $\bV_\ell\in\RR^{2a_\ell\times1}$, and $\bW_\ell\in\RR^{h_{\ell-1}\times h_\ell}$. For these separately parameterized matrices, let $b_{\textup{GAT}}$ be the number of GAT bias parameters. Let $b_{\textup{mask}}\in\set{0,1}$ be the bias count of the scalar mask predictor, and let $S_{\textup{MLP}}$ count all remaining parameters outside the GAT layers and the mask predictor. The parameter counts are
\bals
S_{\textup{GAT}}&\defeq\sum_{\ell=1}^{L}\left(h_{\ell-1}a_\ell+2a_\ell+h_{\ell-1}h_\ell\right)+b_{\textup{GAT}},\\*
S_{\textup{RGAE}}&\defeq S_{\textup{GAT}}+2D+b_{\textup{mask}}+S_{\textup{MLP}},\\*
S_{\textup{GAE}}&\defeq S_{\textup{GAT}}+S_{\textup{MLP}}.
\eals
The mask predictor takes $\bX'_i\Vert\bX'_j\in\RR^{2D}$ and has a scalar output, so edge selection adds exactly $2D+b_{\textup{mask}}$ parameters when all other layers are fixed.

To thoroughly study the complexity of the RGAE, we also analyze its training time complexity. Let $t_{\textup{train}}$ be the number of training epochs. The sparse encoder processes all $N$ nodes and $E$ edges in a full graph pass. Its cost must not be multiplied by $N$ a second time. Under the preceding encoder cost convention, $t_{\textup{train}}$ passes cost $\cO(t_{\textup{train}}(DLME+ME+C_{\textup{MLP}}))$, before loss evaluation. The latent TNN adds $\cO(t_{\textup{train}}N{D\prime}^2)$ as described in Section~\ref{sec:tnn_complexity}. SHED's cost depends on the active cluster count described below. Explicitly materializing every $N\times K\times M$ neighbor map or evaluating a dense pairwise loss still incurs quadratic work or storage. Sparse decoder inference complexity alone does not establish a subquadratic bound for every implementation of RGAE training.

\textbf{Complexity Analysis of SHED.}
We propose SHED to reconstruct the edges connected to a node in the graph. To show its efficiency, we analyze the inference time complexity and the parameter size of the GAE with SHED. For comparison, we also analyze the inference time complexity and the parameter size of GAE, where SHED is replaced by a regular edge decoder that directly reconstructs the adjacency matrix $\bA$. For ease of comparison, we denote the number of parameters and inference cost of all the MLP and GAT layers except SHED as $S_{\text{MLP}}$ and $C_{\text{MLP}}$, respectively. For a node $v_i$ in the graph, let $d_i\defeq\abth{\cC(i)}=\sum_{k=1}^{K}\indict{\hat{\bC}_{ik}>1/2}$ be the number of active clusters in the set $\cC(i)$ defined in Section~\ref{sec:robust_GAE}. Let $D'$ be the dimension of the input feature for SHED. The inference time complexity of GAE with SHED is $\cO(KD'+d_iD'M+C_{\text{MLP}})$, where $\cO(KD')$ is the additional complexity for computing the inter-cluster neighbor map and encoding the cluster indices. $\cO(d_iD'M)$ is the computation cost for computing the intra-cluster neighbor map. In contrast, the inference time complexity of GAE with a regular edge decoder is $\cO(D'KM+C_{\text{MLP}})$. For an observed neighborhood, its number of occupied clusters is at most its degree. This bound need not hold for a predicted cluster map. Let $\bar d_{\mathrm{cl}}\defeq N^{-1}\sum_i d_i$ be the average number of active predicted clusters. The operation count below is conditional on this sparsity. Using the reported Pubmed illustration $\bar d_{\mathrm{cl}}=2.25$ gives an average decoder cost. It is not a pointwise bound on $d_i$ or a consequence of the undirected degree statistics.
As a result, $D'(K+d_iM) \ll D'KM$. For example, when we set $K=100$ and $M=198$ on Pubmed, the illustrative average cost is $\cO(545.5D'+C_{\text{MLP}})$, which is much more efficient than the regular edge decoder whose inference time complexity is $\cO(19800D'+C_{\text{MLP}})$. In general, the inference time complexity of GAE with SHED is much lower than that of GAE with a regular edge decoder.

In addition, the parameter size of GAE with SHED is $D'^{2}+D'K+D'M+S_{\text{MLP}}$, where $D'^{2}$ is the number of parameters in the layer for encoding cluster indices. $D'K$ is the number of parameters in the layer for predicting the inter-cluster neighbor map. $D'M$ is the number of parameters in the layer for reconstructing the intra-cluster neighbor map. In contrast, the parameter size of GAE with a regular edge decoder is $D'KM+S_{\text{MLP}}$. When $D'+K+M \ll KM$,
the parameter size of GAE with SHED is much smaller than that of GAE with a regular edge decoder. For example, $N=19717$ on Pubmed, and with $M=198$ and $K=100$, we have $KM=19800\approx N$.

\section{Additional Experimental Settings}
\label{sec:settings}
\subsection{Datasets}
\label{sec:datasets}
Following previous works on adversarial attacks and defense of GNNs \citep{jin2020graph, zugner_adversarial_2019, entezari2020all},
we evaluate the RGDM on four public benchmark datasets for node classification, including three citation graphs, which are Cora, Citeseer, and Pubmed, and one blog graph, that is, Polblogs. Following previous works on graph adversarial attacks, we evaluate our method and baselines on the largest connected component (LCC) of the graphs. We show the statistics of the datasets in Table~\ref{dataset}.
\begin{table}[!htpb]
\centering
\small
\caption{Statistics of the datasets.}

\resizebox{0.6\columnwidth}{!}{
\begin{tabular}{c|cccc}
\toprule
Datasets    & \# Node & \# Edge  & Classes & Features \\ \midrule
Cora        & 2,485   & 5,069    & 7       & 1,433    \\
Citeseer    & 2,110   & 3,668    & 6       & 3,703    \\
Polblogs    & 1,222   & 16,714   & 2       & 1,222    \\
Pubmed      & 19,717  & 44,338   & 3       & 500      \\
ogbn-arxiv  & 169,343 & 1,166,243 & 40    & 128      \\
Reddit      & 232,965 & 11,606,919 & 41    & 602      \\
 \bottomrule
\end{tabular}
}

\label{dataset}
\end{table}
\subsection{Training Settings}
\label{sec:train_settings}
The training of the RGAE is divided into two phases. In the first phase, we pre-train the RGAE by only optimizing the consistency loss $ \cL^{(t)}_{\textup{C}}$ in Equation~(\ref{eq:AE_loss}) for 2000 epochs. In the second phase, we optimize $\cL^{(t)}$ for another 1000 epochs. We use the Adam optimizer with a learning rate of $0.001$ for the training. The weight decay is set to $1\times10^{-5}$.
We train the LDM in the RGDM after finishing the training of the RGAE. We use the Adam optimizer with a learning rate of $0.0002$ to train the LDM for $3000$ epochs.

To mitigate the effects of potentially unstable masks $\bS_{\textup{inter}}$ and $\bS_{\textup{intra}}$ during early training, we adopt a warm-up strategy widely used in graph structure learning. During the first $100$ epochs, the RGAE is trained only with the reconstruction loss and the low-rank regularization term while keeping $\bS_{\textup{inter}}$ and $\bS_{\textup{intra}}$ fixed to the original hierarchical edge maps. 

\subsection{Attack Settings}
\label{sec:attack_settings}
\textbf{Non-targeted Adversarial Attacks (Metattack)~\citep{zugner_adversarial_2019}.} We first evaluate the robustness of our method against the non-targeted adversarial attack method Metattack. Metattack treats the graph as a hyperparameter to optimize and uses the meta-gradients to solve the bi-level optimization problem, which minimizes the node classification accuracy. We follow the implementation in~\citep{zugner_adversarial_2019}. As Metattack has several variants, we follow \citep{jin2020graph} and adopt the most destructive attack version, Meta-Self, on Cora, Citeseer, and Polblogs datasets. On Pubmed, we adopt the approximate version of Meta-Self, A-Meta-Self, following the settings in \citep{jin2020graph}. We measure the strength of the attack by the perturbation rate, which is the ratio of perturbed edges among all the edges in the graph. We evaluate the RGDM and all the baselines with the perturbation rates ranging from $0$ to $25\%$ with a step of $5\%$.

\noindent\textbf{Targeted Adversarial Attack (Nettack)~\citep{zugner2018adversarial}.} We adopt Nettack as the targeted attack method in evaluating the robustness of our method. Nettack manipulates the edges and node features to degrade the classification accuracy on the target nodes while minimizing the change in the graph’s degree distribution and the feature co-occurrences. We use the default attack settings in the original implementation in~\citep{zugner2018adversarial}. The nodes in the test set whose degree is larger than $10$ are set as the target nodes. The number of perturbations made on every targeted node is defined as the attack budget. Following \citep{jin2020graph}, we evaluate the RGDM and all the baselines with attack budgets ranging from $1$ to $5$ with a step size of $1$. Following the settings in \citep{jin2020graph}, we only sample $10\%$ of the target nodes for the evaluation on the Nettack.


\noindent\textbf{Gradient-based Attack Method (Topology Attack)~\citep{xu2019topology}.} We evaluate the robustness of the RGDM against a gradient-based attack method named Topology Attack~\citep{xu2019topology}. The topology Attack proposes a projected gradient descent (PGD) adversarial attack method that perturbs the edges in a graph to degrade the overall node classification accuracy on the perturbed graph. We adopt the same settings in~\citep{xu2019topology} to attack the graphs in our experiments. Similar to the settings for Metattack, we measure the strength of the attack by the perturbation rate, which is the ratio of the perturbed edges among all the edges in the graph. We evaluate the RGDM and all the baselines with perturbation rates ranging from $0$ to $25\%$ with a step of $5\%$.

\subsection{Tuning the Starting Rank \texorpdfstring{$r_0$}{r0} by Cross-Validation}
\label{sec:rank_CV}
We tune the rank $r_0$ for the TNN, $\norm{\bK}{r_0}$, in Equation (\ref{eq:AE_loss}) by standard cross-validation on each dataset. Let $r_0=\lceil \gamma \min\set{N,D\prime}\rceil$ where $\gamma$ is the rank ratio. We select the values of $\gamma$ by performing $5$-fold cross-validation on $20\%$ of the training data in each dataset. The value of $\gamma$ is selected from $\set{0.1, 0.2,0.3,0.4,0.5,0.6,0.7,0.8,0.9}$.

\subsection{Details on Learning the Low-Rank Weight Matrix and Low-Rank Adjacency Matrix}
\label{sec:low-rank_ablation_details}

To evaluate the robustness of the proposed low-rank regularization in RGDM, we construct two comparison baselines using weight orthogonality regularization and low rank adjacency reconstruction in the RGAE. Next, we detail the implementation of each ablation model.
Following the setting in~\citep{savostianova2023robust}, we regularize the weight matrices of the encoder in the RGAE to approximate orthonormality by adding a penalty term to the training loss of robust RAE, that is
$\cL_{\textup{orth}} = \sum_{l} \fnorm{\bW_l^\top\bW_l-\bI}^2$,
where $\bW_l$ denotes the weight matrix of the $l$-th layer and $\bI$ is the identity matrix. This penalty encourages approximately orthonormal weight columns. By itself, it controls conditioning rather than imposing a low rank. We therefore identify this comparator as weight orthogonality regularization and distinguish it from the spectral tail penalty on the diffusion latent. The strength of this regularization is controlled by the scale factor for $\cL_{\textup{orth}}$, which is tuned via cross-validation following the setting in~\citep{savostianova2023robust}.
For the ablation model that learns the low-rank adjacency matrix, we replace the original attacked adjacency matrix $\bA$ with its low-rank approximation. Following~\citep{entezari2020all}, we compute the top-$r_A$ eigenvectors $\bU_{r_A}$ of the symmetrically normalized adjacency matrix $\tilde{\bA} = \bD^{-1/2} \bA \bD^{-1/2}$, and reconstruct the low-rank adjacency matrix as $\bA_{\text{LR}} = \bD^{1/2} \bU_{r_A} \bLambda_{r_A} \bU_{r_A}^\top \bD^{1/2}$,
where $\bLambda_{r_A}$ is the diagonal matrix containing the top-$r_A$ eigenvalues. The rank $r_A$ is selected via cross-validation following~\citep{entezari2020all}. For all datasets, we select $r_A$ from the same candidate sets as described in Section~\ref{sec:rank_CV}.

\subsection{Compared Methods}
\label{sec:sup_compared_methods}
In our empirical evaluation, we comprehensively benchmark the proposed RGDM against various GNNs, including models that exemplify advances in expressive power, structural modeling, and resilience to adversarial or noisy perturbations. The standard GNN baselines include Graph Convolutional Networks (GCN)~\citep{kipf2017semi}, Graph Attention Networks (GAT)~\citep{GAT}, and Relational GCNs (RGCN)~\citep{rgcn}, which serve as foundational architectures widely adopted for semi-supervised node classification. Additionally, we incorporate GCN-SVD~\citep{entezari2020all}, which leverages truncated singular value decomposition to mitigate the impact of noisy graph spectra.
To rigorously assess the robustness of RGDM under adversarial settings, we further compare it with the state-of-the-art robust GNNs explicitly designed to enhance stability and generalization in the presence of graph corruption. These include GraphBel~\citep{song2022robustness}, which integrates belief propagation with GNNs for noise-tolerant inference; UAG~\citep{feng2021uag}, which unifies adversarial training with graph structure learning; GCORNs~\citep{GCORNs}, a framework that co-regularizes neighborhood consistency and robustness objectives; GADC~\citep{GADC}, which employs adaptive denoising through graph diffusion mechanisms; and HANG~\citep{zhao2023adversarial}, a hierarchical adversarial defense method. We also include Pro-GNN~\citep{jin2020graph}, which jointly optimizes graph structure and model parameters via low-rank and sparsity constraints, and STABLE~\citep{STABLE}, which stabilizes message passing through spectral smoothing.

\section{Additional Experimental Results}
\label{sec:add_results}
\subsection{Eigen-projection and Concentration Ratio of the Ground Truth Label and the Adversarial Noise}
\label{sec:proj_sup}

\noindent\textbf{Projection diagnostics.} We study how the information of the ground-truth class label and the adversarial noise is distributed on different eigenvectors of the feature gram matrix $\bK=\bZ\bZ^{\top}$ by performing eigen-projection, where $\bZ$ is the latent feature learned on the graph with the adversarial noise.
We first compute the eigenvectors $\bU$ of the feature gram matrix $\bK$. Let $\bU^{(1:r)} \in \RR^{N \times r}$ be the top $r$-eigenvectors of $\bK$ and $\bU^{(r)}$ be the $r$-th eigenvector of $\bK$. Then, the eigen-projection value of the ground-truth label $\bY$ on $\bU^{(r)}$ is computed by $p_r = \frac{1}{C}\sum_{c=1}^{C} \ltwonorm{{\bU^{(r)}}^{\top} \bY^{(c)}}^2/ \ltwonorm{ \bY^{(c)}}^2$ for $r \in [N]$,  where $\bY^{(c)}$ is the $c$-th column of $\bY$. We let $\bp = \bth{p_1,\ldots,p_N}  \in \RR^N$. The eigen-projection $p_r$ reflects the amount of the signal projected onto the $r$-th eigenvector of $\bK$, and the signal concentration ratio of a rank $r$ reflects the proportion of signal projected onto the top $r$ eigenvectors of $\bK$. The signal concentration ratio for rank $r$ is computed by $\lonenorm{\bp(1:r)}$, where $\bp(1:r)$ contains the first $r$ elements of $\bp$.
\begin{figure}[!htbp]
\centering
\includegraphics[width=0.7\textwidth]{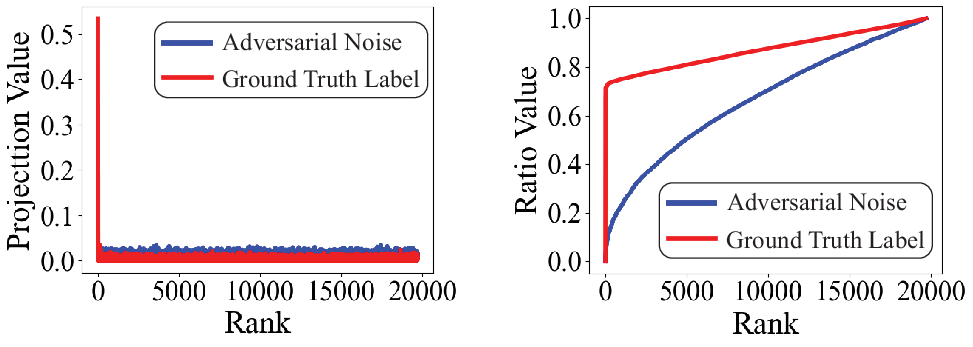}
\vspace{-2mm}
\caption{Eigen-projection (left) and concentration ratio (right) of the ground truth label and the adversarial noise on Pubmed. The study is performed for Metattack with a perturbation ratio of $25\%$. By the rank $r = r_0 = 0.2\min\set{N,D'}$, the signal concentration ratio of the ground truth label is $0.784$. Results on Cora and Citeseer are illustrated in Figure~\ref{fig:eigen-projection} in Section~\ref{sec:proj_sup}.}
\vspace{-3mm}
\label{fig:eigen-projection_pubmed}
\end{figure}
It is observed from the red curves in the left of Figure~\ref{fig:eigen-projection_pubmed} and Figure~\ref{fig:eigen-projection} in Section~\ref{sec:proj_sup} of the appendix that the projection of the ground truth labels mostly concentrates on the top eigenvectors of $\bK$, known as the Low Frequency Property (LFP) widely studied in deep learning~\citep{rahaman19a-spectral-bias,AroraDHLW19-fine-grained-two-layer,CaoFWZG21-spectral-bias,ChorariaD0MC22-spectral-bias-pnns, WangGNSWY24, WangLY25}.
On the other hand, we study how the information of the adversarial noise is distributed on the eigenvectors of the feature gram matrix $\bK$.
Let $\bZ^{(\textup{clean})}$ be the latent features obtained on the clean input graph without adversarial noise. Then $\Delta \bZ = \bZ - \bZ^{(\textup{clean})}$ is the adversarial noise in the feature due to the attacks.
The low-rank part of $\bZ$ characterized by its top-$r$ eigenvectors is $\bZ^{(\textup{low-rank})} = \bU^{(1:r)} \bU^{(1:r)\top} \bZ = \bU^{(1:r)} \bU^{(1:r)\top} (\bZ^{(\textup{clean})} + \Delta \bZ)$. Then $\bU^{(1:r)} \bU^{(1:r)\top} \bZ^{(\textup{clean})} $ is the low-rank part of the clean latent feature $ \bZ^{(\textup{clean})}$, and $\bU^{(1:r)} \bU^{(1:r)\top} \Delta \bZ$ is the adversarial noise induced by the adversarial attack in $\bZ^{(\textup{low-rank})}$. We compute eigen-projection value of the adversarial noise by $p^{(\Delta \bZ)}_r = \fnorm{{\bU^{(r)}}^{\top} \Delta \bZ}/ \fnorm{ \Delta \bZ}$ for $r \in [N]$, which reflects the amount of the adversarial noise projected onto the $r$-th eigenvector of $\bK$. To obtain a ratio in $[0,1]$, we normalize the cumulative projection amplitudes as $\sum_{j=1}^{r}p_j^{(\Delta\bZ)}/\sum_{j=1}^{N}p_j^{(\Delta\bZ)}$ when $\Delta\bZ\ne\bzero$. This is a normalized projection amplitude diagnostic, distinct from the squared energy fraction $\fnorm{\bU^{(1:r)\top}\Delta\bZ}^2/\fnorm{\Delta\bZ}^2$. If $\Delta\bZ=\bzero$, there is no noise concentration ratio to estimate.

The full spectral diagnostics on Cora, Citeseer, and Pubmed are shown in Figure~\ref{fig:eigen-projection}. The concentration of label information in leading eigendirections motivates a soft low rank bias in these experiments. It is neither a universal property of attacked graphs nor an assumption of Theorem~\ref{theorem:optimization-linear-kernel}.
\begin{figure*}[!htb]
\centering
\includegraphics[width=1\textwidth]{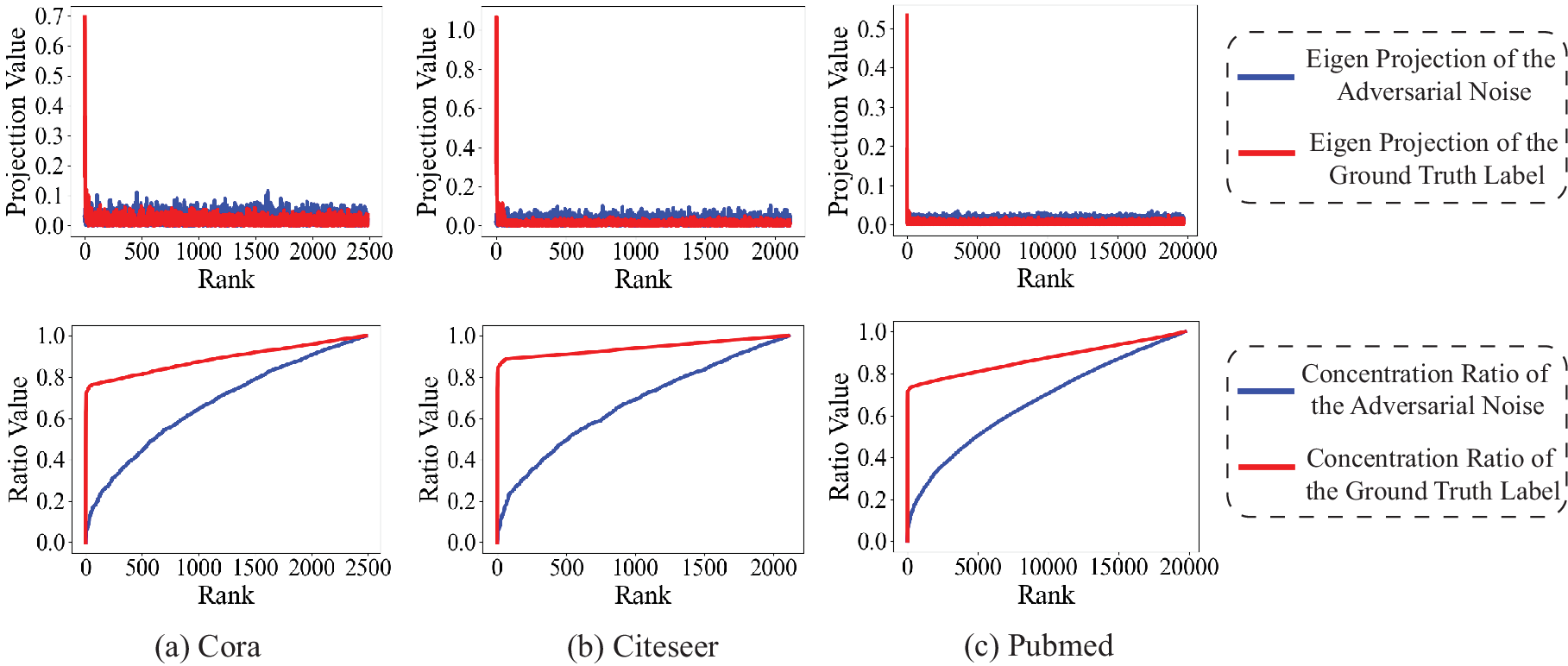}
\caption{Eigen-projection (first row) and concentration ratio (second row) of the ground truth label and the adversarial noise on Cora, Citeseer, and Pubmed. The study in this figure is performed for Metattack with a perturbation ratio of $25\%$. By the rank $r= r_0 =0.2\min\set{N,D'}$, the signal concentration ratios of the ground truth label for Cora, Citeseer, and Pubmed are $0.844$, $0.809$, and $0.784$ respectively.}
\label{fig:eigen-projection}
\end{figure*}
\subsection{Detailed Semi-Supervised Node Classification Results}
\label{sec:sup_detailed_results}

The detailed experiment results for the RGDM under the Metattack, Nettack, and Topology Attack are shown in Table~\ref{table:metattack_sup}, Table~\ref{table:nettack_sup} and Table~\ref{table:topology_attack_sup}.
\begin{table*}[ht]
\small
\caption{Node classification performance (Accuracy$\pm$Std) under the non-targeted poisoning attack (Metattack)~\citep{zugner_adversarial_2019}. The improvements of the RGDM over the corresponding baselines are attached in parentheses after the results of the RGDM.}

\label{table:metattack_sup}
\centering\resizebox{0.95\textwidth}{!}{
\begin{tabular}{c|c|cccccccccc|ccc}
\toprule
Dataset        & Ptb Rate (\%)  & GCN                     & GAT                        & RGCN            & GCN-SVD         &GraphBel              & UAG             & GCORNs        &GADC          &HANG  & Pro-GNN    & RGDM (GCN)                                              & RGDM (HANG) &RGDM (Pro-GNN)                          \\ \midrule
\multirow{6}{*}{Cora}     & 0             & 83.50$\pm$0.44 & 83.97$\pm$0.65 & 83.09$\pm$0.44 & 80.63$\pm$0.45 & 83.42$\pm$0.52 & 82.05$\pm$0.51 & 82.56$\pm$0.48 & 83.22$\pm$0.61 & 80.07$\pm$0.32 & 82.98$\pm$0.23 & 85.23$\pm$0.44 ($\uparrow 1.73$) & 82.31$\pm$0.51 ($\uparrow 2.24$) & 84.29$\pm$0.52 ($\uparrow 1.31$) \\
                          & 5             & 76.55$\pm$0.79 & 80.44$\pm$0.74 & 77.42$\pm$0.39 & 78.39$\pm$0.54 & 82.78$\pm$0.39 & 79.13$\pm$0.59 & 75.91$\pm$1.88 & 75.68$\pm$1.10 & 78.12$\pm$0.72 & 82.27$\pm$0.45 & 80.19$\pm$0.39 ($\uparrow 3.64$) & 81.04$\pm$0.59 ($\uparrow 2.92$) & 83.19$\pm$0.39 ($\uparrow 0.92$) \\
                          & 10            & 70.39$\pm$1.28 & 75.61$\pm$0.59 & 72.22$\pm$0.38 & 71.47$\pm$0.83 & 77.91$\pm$0.86 & 75.16$\pm$0.76 & 74.63$\pm$1.02 & 73.70$\pm$1.17 & 75.96$\pm$0.62 & 79.03$\pm$0.59 & 76.73$\pm$0.38 ($\uparrow 6.34$) & 78.03$\pm$0.76 ($\uparrow 2.07$) & 81.16$\pm$0.86 ($\uparrow 2.13$) \\
                          & 15            & 65.10$\pm$0.71 & 69.78$\pm$1.28 & 66.82$\pm$0.39 & 66.69$\pm$1.18 & 76.01$\pm$1.12 & 71.03$\pm$0.64 & 72.57$\pm$1.24 & 71.39$\pm$1.22 & 72.89$\pm$0.96 & 76.40$\pm$1.27 & 69.37$\pm$0.39 ($\uparrow 4.27$) & 76.39$\pm$0.64 ($\uparrow 3.50$) & 78.29$\pm$1.12 ($\uparrow 1.89$) \\
                          & 20            & 59.56$\pm$1.72 & 59.94$\pm$0.92 & 59.27$\pm$0.37 & 58.94$\pm$1.13 & 68.78$\pm$1.84 & 65.71$\pm$0.89 & 71.94$\pm$1.51 & 71.20$\pm$1.49 & 70.36$\pm$1.09 & 73.32$\pm$1.56 & 65.97$\pm$0.37 ($\uparrow 6.41$) & 73.25$\pm$0.89 ($\uparrow 2.89$) & 74.85$\pm$1.84 ($\uparrow 1.53$) \\
                          & 25            & 47.53$\pm$1.91 & 54.78$\pm$0.74 & 50.51$\pm$0.78 & 52.06$\pm$1.19 & 56.54$\pm$1.58 & 60.82$\pm$1.08 & 69.48$\pm$1.94 & 68.29$\pm$1.36 & 68.72$\pm$1.87 & 69.72$\pm$1.69 & 61.42$\pm$0.78 ($\uparrow 13.89$) & 70.44$\pm$1.08 ($\uparrow 1.72$) & 72.31$\pm$1.58 ($\uparrow 2.59$) \\ \hline
\multirow{6}{*}{Citeseer} & 0             & 71.96$\pm$0.55 & 73.26$\pm$0.83 & 71.20$\pm$0.83 & 70.65$\pm$0.32 & 72.26$\pm$0.83 & 72.10$\pm$0.63 & 72.63$\pm$0.40 & 73.05$\pm$0.30 & 73.26$\pm$0.38 & 73.28$\pm$0.69 & 73.66$\pm$0.83 ($\uparrow 1.70$) & 74.21$\pm$0.63 ($\uparrow 0.95$) & 74.47$\pm$0.83 ($\uparrow 1.19$) \\
                          & 5             & 70.88$\pm$0.62 & 72.89$\pm$0.83 & 70.50$\pm$0.43 & 68.84$\pm$0.72 & 71.31$\pm$0.43 & 70.51$\pm$0.97 & 68.18$\pm$1.68 & 68.67$\pm$0.89 & 73.09$\pm$0.34 & 72.93$\pm$0.57 & 73.14$\pm$0.43 ($\uparrow 2.26$) & 74.04$\pm$0.97 ($\uparrow 0.95$) & 74.29$\pm$0.43 ($\uparrow 1.36$) \\
                          & 10            & 67.55$\pm$0.89 & 70.63$\pm$0.48 & 67.71$\pm$0.30 & 68.87$\pm$0.62 & 71.63$\pm$0.30 & 69.54$\pm$0.56 & 66.24$\pm$1.52 & 68.66$\pm$1.48 & 72.43$\pm$0.52 & 72.51$\pm$0.75 & 71.35$\pm$0.30 ($\uparrow 3.80$) & 73.85$\pm$0.56 ($\uparrow 1.42$) & 74.03$\pm$0.30 ($\uparrow 1.52$) \\
                          & 15            & 64.52$\pm$1.11 & 69.02$\pm$1.09 & 65.69$\pm$0.37 & 63.26$\pm$0.96 & 69.82$\pm$0.37 & 65.95$\pm$0.94 & 65.48$\pm$1.03 & 68.29$\pm$1.88 & 70.82$\pm$0.87 & 72.03$\pm$1.11 & 68.72$\pm$0.37 ($\uparrow 4.20$) & 73.21$\pm$0.94 ($\uparrow 2.39$) & 73.65$\pm$0.37 ($\uparrow 1.62$) \\
                          & 20            & 62.03$\pm$1.49 & 61.04$\pm$1.52 & 62.49$\pm$1.22 & 58.55$\pm$1.09 & 64.22$\pm$1.22 & 59.30$\pm$1.40 & 65.55$\pm$1.55 & 67.87$\pm$1.57 & 66.19$\pm$1.38 & 70.02$\pm$1.28 & 66.17$\pm$1.22 ($\uparrow 4.14$) & 69.57$\pm$1.40 ($\uparrow 3.38$) & 71.87$\pm$1.22 ($\uparrow 1.85$) \\
                          & 25            & 56.94$\pm$1.09 & 61.85$\pm$1.12 & 55.35$\pm$0.66 & 57.18$\pm$1.87 & 66.23$\pm$0.66 & 59.89$\pm$1.47 & 65.23$\pm$1.26 & 64.82$\pm$1.31 & 66.40$\pm$1.57 & 68.95$\pm$1.78 & 62.84$\pm$0.66 ($\uparrow 5.90$) & 69.05$\pm$1.47 ($\uparrow 2.65$) & 70.24$\pm$0.66 ($\uparrow 1.29$) \\ \hline
\multirow{6}{*}{Polblogs} & 0             & 95.69$\pm$0.38 & 95.35$\pm$0.20 & 95.22$\pm$0.14 & 95.31$\pm$0.18 & 85.13$\pm$1.22 & 90.13$\pm$1.22 & 95.31$\pm$0.88 & 95.54$\pm$0.17 & 94.77$\pm$1.07 & 93.20$\pm$0.64 & 96.13$\pm$0.14 ($\uparrow 0.44$) & 94.95$\pm$1.22 ($\uparrow 0.18$) & 95.70$\pm$1.22 ($\uparrow 2.50$) \\
                          & 5             & 73.07$\pm$0.80 & 83.69$\pm$1.45 & 74.34$\pm$0.19 & 89.09$\pm$0.22 & 51.84$\pm$1.38 & 84.84$\pm$1.38 & 89.32$\pm$0.33 & 90.52$\pm$0.27 & 80.19$\pm$1.52 & 93.29$\pm$0.18 & 78.57$\pm$0.19 ($\uparrow 5.50$) & 88.25$\pm$1.38 ($\uparrow 8.06$) & 94.33$\pm$1.38 ($\uparrow 1.04$) \\
                          & 10            & 70.72$\pm$1.13 & 76.32$\pm$0.85 & 71.04$\pm$0.34 & 81.24$\pm$0.49 & 56.54$\pm$1.30 & 75.54$\pm$1.30 & 87.34$\pm$1.77 & 86.30$\pm$1.45 & 74.92$\pm$1.32 & 89.42$\pm$1.09 & 74.33$\pm$0.34 ($\uparrow 3.61$) & 80.47$\pm$1.30 ($\uparrow 5.55$) & 91.35$\pm$1.30 ($\uparrow 1.93$) \\
                          & 15            & 64.96$\pm$1.91 & 68.80$\pm$1.14 & 67.28$\pm$0.38 & 68.10$\pm$1.73 & 53.41$\pm$1.08 & 66.41$\pm$1.08 & 82.07$\pm$1.41 & 83.45$\pm$1.75 & 71.65$\pm$1.34 & 86.04$\pm$1.21 & 69.85$\pm$0.38 ($\uparrow 4.89$) & 78.49$\pm$1.08 ($\uparrow 6.84$) & 88.01$\pm$1.08 ($\uparrow 1.97$) \\
                          & 20            & 51.27$\pm$1.23 & 51.50$\pm$1.63 & 59.89$\pm$0.34 & 57.33$\pm$1.15 & 52.18$\pm$0.54 & 58.18$\pm$0.54 & 69.92$\pm$1.26 & 72.84$\pm$1.24 & 66.27$\pm$1.39 & 79.56$\pm$1.68 & 59.38$\pm$0.34 ($\uparrow 8.11$) & 72.51$\pm$0.54 ($\uparrow 6.24$) & 81.34$\pm$0.54 ($\uparrow 1.78$) \\
                          & 25            & 49.23$\pm$1.36 & 51.19$\pm$1.49 & 56.02$\pm$0.56 & 48.66$\pm$1.93 & 51.39$\pm$1.36 & 55.39$\pm$1.36 & 66.59$\pm$1.74 & 65.52$\pm$1.93 & 65.80$\pm$1.33 & 63.18$\pm$1.40 & 56.47$\pm$0.56 ($\uparrow 7.24$) & 70.79$\pm$1.36 ($\uparrow 4.99$) & 66.92$\pm$1.36 ($\uparrow 3.74$) \\ \hline
\multirow{6}{*}{Pubmed}   & 0             & 87.19$\pm$0.09 & 83.73$\pm$0.40 & 86.16$\pm$0.18 & 83.44$\pm$0.21 & 84.02$\pm$0.26 & 87.06$\pm$0.06 & 86.42$\pm$0.71 & 87.26$\pm$0.25 & 85.08$\pm$0.20 & 87.33$\pm$0.18 & 88.10$\pm$0.18 ($\uparrow 0.91$) & 87.15$\pm$0.11 ($\uparrow 2.07$) & 88.41$\pm$0.26 ($\uparrow 1.08$) \\
                          & 5             & 83.09$\pm$0.13 & 78.00$\pm$0.44 & 81.08$\pm$0.20 & 83.41$\pm$0.15 & 83.91$\pm$0.26 & 86.39$\pm$0.06 & 82.96$\pm$0.89 & 81.64$\pm$0.64 & 85.08$\pm$0.18 & 87.25$\pm$0.09 & 86.99$\pm$0.20 ($\uparrow 3.90$) & 87.02$\pm$0.11 ($\uparrow 1.94$) & 88.26$\pm$0.26 ($\uparrow 1.01$) \\
                          & 10            & 81.21$\pm$0.09 & 74.93$\pm$0.38 & 77.51$\pm$0.27 & 83.27$\pm$0.21 & 84.62$\pm$0.26 & 85.70$\pm$0.07 & 80.76$\pm$1.65 & 80.43$\pm$1.64 & 85.17$\pm$0.23 & 87.25$\pm$0.09 & 85.83$\pm$0.27 ($\uparrow 4.62$) & 87.13$\pm$0.14 ($\uparrow 1.96$) & 88.15$\pm$0.26 ($\uparrow 0.90$) \\
                          & 15            & 78.66$\pm$0.12 & 71.13$\pm$0.51 & 73.91$\pm$0.25 & 83.10$\pm$0.18 & 84.83$\pm$0.20 & 84.76$\pm$0.08 & 76.82$\pm$1.83 & 75.72$\pm$1.24 & 85.00$\pm$0.22 & 87.20$\pm$0.09 & 84.97$\pm$0.25 ($\uparrow 6.31$) & 86.75$\pm$0.13 ($\uparrow 1.75$) & 88.09$\pm$0.20 ($\uparrow 0.89$) \\
                          & 20            & 77.35$\pm$0.19 & 68.21$\pm$0.96 & 71.18$\pm$0.31 & 83.01$\pm$0.22 & 84.89$\pm$0.45 & 83.88$\pm$0.05 & 71.28$\pm$1.52 & 72.80$\pm$1.89 & 85.20$\pm$0.19 & 87.15$\pm$0.15 & 82.45$\pm$0.31 ($\uparrow 5.10$) & 86.13$\pm$0.11 ($\uparrow 0.93$) & 88.00$\pm$0.45 ($\uparrow 0.85$) \\
                          & 25            & 75.50$\pm$0.17 & 65.41$\pm$0.77 & 67.95$\pm$0.15 & 82.72$\pm$0.18 & 85.07$\pm$0.15 & 83.66$\pm$0.06 & 70.75$\pm$1.86 & 71.60$\pm$1.50 & 85.06$\pm$0.17 & 86.76$\pm$0.19 & 81.52$\pm$0.15 ($\uparrow 6.02$) & 86.26$\pm$0.12 ($\uparrow 1.20$) & 87.58$\pm$0.15 ($\uparrow 0.82$) \\ \hline
\end{tabular}

}

\end{table*}

\begin{table*}[!ht]
\small
\caption{Node classification performance (Accuracy$\pm$Std) under targeted attack (Nettack)~\citep{zugner2018adversarial}. The improvements of RGDM over the corresponding baselines are attached in parentheses after the results of the RGDM.}

\label{table:nettack_sup}
\resizebox{\textwidth}{!}{
\begin{tabular}{c|c|cccccccccc|ccc}
\toprule
Dataset                   & Attack Budget  & GCN              & GAT                        & RGCN           & GCN-SVD         &GraphBel        &UAG                 &GCORNs        &GADC        & HANG                     & Pro-GNN                     & RGDM (GCN)  & RGDM (HANG) & RGDM (Pro-GNN)       \\ \midrule
\multirow{6}{*}{Cora}     & 0              & 81.43$\pm$1.04   & 82.14$\pm$1.13             & 81.34$\pm$1.55 & 81.10$\pm$1.34  & 71.55$\pm$1.01 & 82.34$\pm$1.08     & 82.61$\pm$0.41 & 81.36$\pm$1.83        &80.75$\pm$1.27            & 84.78$\pm$0.59            &83.43$\pm$1.08 ($\uparrow 2.00$)  & 82.10$\pm$1.34 ($\uparrow 1.35$) & 86.00$\pm$1.13 ($\uparrow 1.22$)    \\
                          & 1              & 75.06$\pm$1.02   & 76.04$\pm$1.08             & 76.75$\pm$1.71 & 77.23$\pm$1.82  & 63.73$\pm$1.25 & 81.75$\pm$1.16     & 81.05$\pm$1.14 & 80.05$\pm$1.53        &76.99$\pm$1.16            & 83.85$\pm$0.51            &78.41$\pm$1.16 ($\uparrow 3.35$)  & 80.39$\pm$1.82 ($\uparrow 3.40$) & 84.79$\pm$1.08 ($\uparrow 0.94$)    \\
                          & 2              & 70.60$\pm$1.10   & 70.24$\pm$1.43             & 70.96$\pm$1.14 & 72.53$\pm$1.60  & 62.41$\pm$1.94 & 77.96$\pm$1.84     & 78.47$\pm$1.32 & 75.90$\pm$1.29        &76.51$\pm$1.60            & 78.65$\pm$0.66            &73.79$\pm$1.84 ($\uparrow 3.19$)  & 79.14$\pm$1.60 ($\uparrow 2.63$) & 80.02$\pm$1.43 ($\uparrow 1.37$)    \\
                          & 3              & 67.95$\pm$1.72   & 65.54$\pm$1.34             & 66.51$\pm$1.60 & 66.75$\pm$1.54  & 61.20$\pm$1.08 & 72.51$\pm$1.14     & 73.17$\pm$0.74 & 71.94$\pm$1.45        &73.13$\pm$1.85            & 72.40$\pm$0.53              &70.46$\pm$1.14 ($\uparrow 2.51$)  & 76.31$\pm$1.54 ($\uparrow 3.18$) & 74.31$\pm$1.34 ($\uparrow 1.91$)    \\
                          & 4              & 61.57$\pm$1.47   & 61.69$\pm$0.90             & 59.28$\pm$1.68 & 60.72$\pm$1.63  & 56.51$\pm$1.72 & 70.28$\pm$1.52     & 71.26$\pm$0.62 & 70.93$\pm$1.19        &72.53$\pm$1.14            & 70.12$\pm$0.66              &64.37$\pm$1.52 ($\uparrow 2.80$)  & 75.12$\pm$1.63 ($\uparrow 2.59$) & 72.79$\pm$0.90 ($\uparrow 2.67$)    \\
                          & 5              & 55.54$\pm$1.66   & 58.31$\pm$1.03             & 55.30$\pm$1.66 & 57.71$\pm$1.82  & 51.93$\pm$1.77 & 66.30$\pm$1.21     & 66.89$\pm$1.08 & 64.49$\pm$1.42        &68.80$\pm$1.55            & 66.89$\pm$0.73              &59.44$\pm$1.21 ($\uparrow 3.90$)  & 69.81$\pm$1.82 ($\uparrow 1.01$) & 69.81$\pm$1.03 ($\uparrow 2.92$)   \\ \midrule
\multirow{6}{*}{Citeseer} & 0              & 81.02$\pm$1.35   & 82.26$\pm$1.04             & 80.25$\pm$1.12 & 81.08$\pm$1.34  & 76.55$\pm$1.23 & 81.25$\pm$1.12     & 81.94$\pm$1.17 & 80.63$\pm$1.04        &81.09$\pm$1.12            & 82.12$\pm$0.81              &82.87$\pm$1.12 ($\uparrow 1.85$)  & 82.84$\pm$1.34 ($\uparrow 1.75$) & 83.05$\pm$1.04 ($\uparrow 0.93$)   \\
                          & 1              & 78.41$\pm$1.62   & 81.27$\pm$1.38             & 78.25$\pm$0.73 & 80.16$\pm$1.04  & 68.89$\pm$1.67 & 80.25$\pm$0.73     & 80.79$\pm$0.89 & 79.02$\pm$1.84        &79.05$\pm$1.38            & 81.75$\pm$0.79            &80.39$\pm$0.73 ($\uparrow 1.98$)  & 81.40$\pm$1.04 ($\uparrow 2.35$) & 82.47$\pm$1.38 ($\uparrow 0.72$)  \\
                          & 2              & 74.92$\pm$1.54   & 77.43$\pm$1.89             & 75.40$\pm$1.04 & 79.84$\pm$0.73  & 67.62$\pm$1.11 & 79.40$\pm$1.04     & 79.65$\pm$1.84 & 78.22$\pm$1.29        &77.94$\pm$1.29            & 81.27$\pm$0.95            &77.15$\pm$1.04 ($\uparrow 2.23$)  & 80.11$\pm$0.73 ($\uparrow 2.17$) & 82.30$\pm$1.89 ($\uparrow 1.03$)  \\
                          & 3              & 63.97$\pm$1.69   & 60.85$\pm$1.99             & 60.31$\pm$1.19 & 77.14$\pm$1.86  & 60.63$\pm$1.87 & 78.31$\pm$1.19     & 78.10$\pm$1.06 & 78.11$\pm$1.39        &77.14$\pm$1.48            & 79.68$\pm$1.98            &68.53$\pm$1.19 ($\uparrow 4.56$)  & 78.95$\pm$1.86 ($\uparrow 1.81$) & 80.85$\pm$1.99 ($\uparrow 1.17$)  \\
                          & 4              & 55.40$\pm$1.60   & 61.59$\pm$1.64             & 55.49$\pm$1.75 & 69.52$\pm$1.31  & 53.17$\pm$1.48 & 77.49$\pm$1.75     & 77.75$\pm$1.66 & 77.49$\pm$1.54        &78.41$\pm$1.62            & 77.78$\pm$1.84              &59.57$\pm$1.75 ($\uparrow 4.17$)  & 78.99$\pm$1.31 ($\uparrow 0.58$) & 79.11$\pm$1.64 ($\uparrow 1.33$)  \\
                          & 5              & 47.62$\pm$1.17   & 55.56$\pm$1.28             & 47.44$\pm$1.01 & 69.21$\pm$1.48  & 48.73$\pm$1.60 & 71.44$\pm$1.01     & 71.66$\pm$1.79 & 70.54$\pm$1.57        &73.49$\pm$1.48            & 71.27$\pm$1.99              &55.71$\pm$1.01 ($\uparrow 8.09$)  & 75.92$\pm$1.48 ($\uparrow 2.43$) & 74.60$\pm$1.28 ($\uparrow 3.33$)  \\ \midrule
\multirow{6}{*}{Polblogs} & 0              & 97.03$\pm$0.24   & 97.25$\pm$0.32             & 97.04$\pm$0.11 & 97.50$\pm$0.22  & 71.35$\pm$1.01 & 97.11$\pm$0.11     & 97.52$\pm$0.26 & 97.38$\pm$0.34        & 97.41$\pm$0.50         &  97.12$\pm$0.15             &97.69$\pm$0.11 ($\uparrow 0.66$)  & 97.65$\pm$0.22 ($\uparrow 0.24$) & 97.80$\pm$0.32 ($\uparrow 0.68$)   \\
                          & 1              & 96.83$\pm$0.17   & 97.22$\pm$0.25             & 97.00$\pm$0.07 & 97.56$\pm$0.20  & 68.17$\pm$1.25 & 96.00$\pm$0.07     & 97.61$\pm$0.59 & 97.18$\pm$0.35        & 97.37$\pm$0.37         &  96.83$\pm$0.06             &97.67$\pm$0.07 ($\uparrow 0.84$)  & 97.63$\pm$0.20 ($\uparrow 0.26$) & 97.69$\pm$0.25 ($\uparrow 0.86$)  \\
                          & 2              & 95.61$\pm$0.20   & 96.11$\pm$0.65             & 95.87$\pm$0.23 & 97.12$\pm$0.09  & 65.48$\pm$1.85 & 97.02$\pm$0.23     & 96.08$\pm$0.45 & 95.37$\pm$0.59        & 96.89$\pm$0.16           &  97.17$\pm$0.12           &97.10$\pm$0.23 ($\uparrow 1.49$)  & 97.24$\pm$0.09 ($\uparrow 0.35$) & 97.58$\pm$0.65 ($\uparrow 0.41$)  \\
                          & 3              & 95.41$\pm$0.18   & 95.81$\pm$0.56             & 95.59$\pm$0.27 & 96.61$\pm$0.14  & 62.59$\pm$1.99 & 96.59$\pm$0.27     & 95.65$\pm$1.01 & 95.47$\pm$0.13        & 96.65$\pm$0.15         &  96.93$\pm$0.12             &96.99$\pm$0.27 ($\uparrow 1.58$)  & 97.01$\pm$0.14 ($\uparrow 0.36$) & 97.22$\pm$0.56 ($\uparrow 0.29$)  \\
                          & 4              & 94.24$\pm$0.24   & 94.80$\pm$0.66             & 94.37$\pm$0.26 & 96.17$\pm$0.19  & 58.68$\pm$0.40 & 96.07$\pm$0.26     & 94.44$\pm$0.53 & 94.97$\pm$0.61        & 96.26$\pm$0.53         &  96.89$\pm$0.16           &95.44$\pm$0.26 ($\uparrow 1.20$)  & 97.03$\pm$0.19 ($\uparrow 0.77$) & 97.12$\pm$0.66 ($\uparrow 0.23$)  \\
                          & 5              & 93.00$\pm$0.48   & 93.28$\pm$1.43             & 93.20$\pm$0.43 & 95.13$\pm$0.25  & 59.02$\pm$1.19 & 95.60$\pm$0.43     & 93.42$\pm$0.85 & 93.27$\pm$0.46        & 95.91$\pm$0.33         &  96.13$\pm$0.25           &94.90$\pm$0.43 ($\uparrow 1.90$)  & 96.70$\pm$0.25 ($\uparrow 0.79$) & 96.88$\pm$1.43 ($\uparrow 0.75$)  \\ \midrule
\multirow{6}{*}{Pubmed}   & 0              & 88.13$\pm$1.35   & 87.03$\pm$1.13             & 84.87$\pm$1.35 & 87.27$\pm$1.74  & 85.47$\pm$1.45 & 87.31$\pm$1.35     & 84.56$\pm$1.39 & 85.49$\pm$1.16        & 85.35$\pm$1.20           & 88.46$\pm$1.20            &89.21$\pm$1.35 ($\uparrow 1.08$)  & 87.28$\pm$1.74 ($\uparrow 1.93$) & 89.58$\pm$1.13 ($\uparrow 1.12$)  \\
                          & 1              & 87.04$\pm$1.62   & 84.95$\pm$1.08             & 83.75$\pm$1.71 & 84.15$\pm$1.25  & 84.34$\pm$1.33 & 86.57$\pm$1.71     & 83.88$\pm$1.94 & 84.58$\pm$1.28        & 84.42$\pm$1.01           & 87.58$\pm$1.01            &88.03$\pm$1.71 ($\uparrow 0.99$)  & 86.88$\pm$1.25 ($\uparrow 2.46$) & 88.80$\pm$1.08 ($\uparrow 1.22$)  \\
                          & 2              & 84.12$\pm$1.54   & 83.90$\pm$1.43             & 84.21$\pm$1.14 & 83.90$\pm$1.94  & 84.27$\pm$1.25 & 85.21$\pm$1.14     & 84.40$\pm$1.73 & 84.17$\pm$1.53        & 84.38$\pm$1.05           & 85.87$\pm$1.05            &85.33$\pm$1.14 ($\uparrow 1.21$)  & 86.35$\pm$1.94 ($\uparrow 1.97$) & 87.14$\pm$1.43 ($\uparrow 1.27$)  \\
                          & 3              & 81.25$\pm$1.69   & 81.27$\pm$1.34             & 82.48$\pm$1.60 & 81.22$\pm$1.08  & 84.11$\pm$1.67 & 83.88$\pm$1.60     & 82.26$\pm$1.49 & 84.35$\pm$1.33        & 84.01$\pm$1.12           & 84.35$\pm$1.12            &83.37$\pm$1.60 ($\uparrow 2.12$)  & 85.21$\pm$1.08 ($\uparrow 1.20$) & 85.79$\pm$1.34 ($\uparrow 1.44$)  \\
                          & 4              & 76.35$\pm$1.60   & 78.54$\pm$0.90             & 80.02$\pm$1.68 & 77.02$\pm$1.72  & 79.03$\pm$1.22 & 80.00$\pm$1.32     & 81.06$\pm$1.72 & 78.01$\pm$1.20        & 79.73$\pm$1.31           & 80.04$\pm$1.31            &78.45$\pm$1.32 ($\uparrow 2.10$)  & 81.65$\pm$1.23 ($\uparrow 1.92$) & 81.95$\pm$0.90 ($\uparrow 1.91$)  \\
                          & 5              & 68.32$\pm$1.17   & 74.25$\pm$1.03             & 73.14$\pm$1.66 & 71.84$\pm$1.77  & 70.11$\pm$1.36 & 72.20$\pm$1.39     & 72.28$\pm$1.13 & 71.26$\pm$1.61        & 70.59$\pm$1.31           & 72.18$\pm$1.31              &71.42$\pm$1.39 ($\uparrow 3.10$)  & 74.43$\pm$1.34 ($\uparrow 3.84$) & 76.27$\pm$1.03 ($\uparrow 4.09$)  \\
\bottomrule
\end{tabular}
}

\end{table*}

\begin{table*}[!htbp]
\small
\caption{Node classification performance (Accuracy$\pm$Std) under Topology Attack~\citep{xu2019topology}. The improvements of the RGDM over the corresponding baselines are attached in parentheses after the results of the RGDM.}

\label{table:topology_attack_sup}

 \resizebox{\textwidth}{!}{
\begin{tabular}{c|c|cccccccccc|ccc}
\toprule
Dataset                   & Ptb Rate (\%)        & GCN                           & GAT                        & RGCN           & GCN-SVD        & GraphBel        &UAG                 &GCORNs          &GADC                 & HANG                                & Pro-GNN  & RGDM(GCN) & RGDM(HANG) & RGDM(Pro-GNN)        \\ \midrule
\multirow{6}{*}{Cora}       & 0                    & 83.50$\pm$0.44               & 83.97$\pm$0.65           & 83.09$\pm$0.44  & 80.63$\pm$0.45  & 83.42$\pm$0.52  & 82.05$\pm$0.51    & 82.63$\pm$0.47 & 83.21$\pm$0.72         & 80.07$\pm$0.32 & 82.98$\pm$0.23   & 85.23$\pm$0.44 ($\uparrow 1.73$)  & 82.31$\pm$0.51 ($\uparrow 2.24$) & 84.29$\pm$0.52 ($\uparrow 1.31$)\\
                            & 5                    & 75.50$\pm$0.44               & 76.97$\pm$0.65             & 75.04$\pm$1.32 & 75.82$\pm$1.24  & 77.01$\pm$1.33   & 76.22$\pm$1.24    & 75.96$\pm$1.81 & 76.83$\pm$1.25          & 76.99$\pm$1.52            & 77.63$\pm$1.85           & 77.92$\pm$1.08 ($\uparrow 2.42$)  & 79.39$\pm$1.33 ($\uparrow 2.40$) & 81.10$\pm$1.16 ($\uparrow 3.47$) \\
                            & 10                   & 72.02$\pm$1.32               & 74.10$\pm$1.84             & 73.05$\pm$1.25 & 74.00$\pm$1.76  & 74.00$\pm$1.57   & 76.00$\pm$1.76    & 74.67$\pm$1.03 & 75.90$\pm$1.84          & 76.99$\pm$1.52            & 77.63$\pm$1.85           & 74.22$\pm$1.43 ($\uparrow 2.20$)  & 79.44$\pm$1.57 ($\uparrow 2.45$) & 80.36$\pm$1.84 ($\uparrow 2.73$) \\
                            & 15                   & 69.42$\pm$1.71               & 70.55$\pm$1.69             & 71.61$\pm$1.34 & 72.90$\pm$1.46  & 72.79$\pm$1.26   & 74.90$\pm$1.33    & 72.58$\pm$1.27 & 74.34$\pm$1.52          & 75.63$\pm$1.59          & 75.40$\pm$1.32             & 72.42$\pm$1.34 ($\uparrow 3.00$)  & 76.92$\pm$1.26 ($\uparrow 1.29$) & 76.57$\pm$1.14 ($\uparrow 1.17$) \\
                            & 20                   & 67.61$\pm$1.02               & 68.84$\pm$1.33             & 67.87$\pm$1.22 & 70.00$\pm$1.83  & 70.55$\pm$1.17   & 71.85$\pm$1.22    & 71.99$\pm$1.54 & 72.06$\pm$1.80          & 72.15$\pm$1.27          & 71.32$\pm$1.79             & 71.61$\pm$0.90 ($\uparrow 4.00$)  & 74.35$\pm$1.17 ($\uparrow 2.20$) & 74.44$\pm$1.52 ($\uparrow 3.12$) \\
                            & 25                   & 64.81$\pm$1.14               & 65.51$\pm$1.45             & 66.02$\pm$1.39 & 67.89$\pm$1.21  & 66.32$\pm$1.07   & 68.03$\pm$1.39    & 69.47$\pm$1.98 & 68.92$\pm$1.27          & 69.10$\pm$1.33          & 68.72$\pm$1.50             & 69.53$\pm$1.03 ($\uparrow 4.72$)  & 71.67$\pm$1.07 ($\uparrow 2.57$) & 72.07$\pm$1.21 ($\uparrow 3.35$) \\ \midrule
\multirow{6}{*}{Citeseer}   & 0                    & 71.96$\pm$0.55               & 73.26$\pm$0.83             & 71.20$\pm$0.83  & 70.65$\pm$0.32  & 72.26$\pm$0.83  & 72.10$\pm$0.63    & 72.71$\pm$0.53 & 72.54$\pm$0.21           & 73.26$\pm$0.38 & 73.28$\pm$0.69   & 73.66$\pm$0.83 ($\uparrow 1.70$)  & 74.21$\pm$0.63 ($\uparrow 0.95$) & 74.47$\pm$0.83 ($\uparrow 1.19$)\\
                            & 5                    & 67.96$\pm$0.59               & 68.15$\pm$0.89             & 68.20$\pm$0.83 & 68.65$\pm$1.32  & 69.76$\pm$1.48   & 71.32$\pm$0.83    & 68.16$\pm$1.68 & 69.57$\pm$1.38          & 69.36$\pm$1.20            & 71.24$\pm$1.13           & 69.76$\pm$1.38 ($\uparrow 1.80$)  & 72.10$\pm$1.48 ($\uparrow 2.74$) & 72.77$\pm$0.73 ($\uparrow 1.53$) \\
                            & 10                   & 64.10$\pm$1.34               & 66.67$\pm$1.35             & 65.92$\pm$1.86 & 66.57$\pm$1.35  & 66.59$\pm$1.98   & 67.86$\pm$1.86    & 66.26$\pm$1.54 & 68.06$\pm$1.46          & 67.52$\pm$1.20            & 68.19$\pm$1.13           & 67.30$\pm$1.89 ($\uparrow 3.20$)  & 68.53$\pm$1.98 ($\uparrow 1.01$) & 69.32$\pm$1.04 ($\uparrow 1.13$) \\
                            & 15                   & 61.12$\pm$1.37               & 63.62$\pm$1.47             & 64.64$\pm$1.72 & 65.88$\pm$1.16  & 65.37$\pm$1.34   & 66.43$\pm$1.72    & 65.49$\pm$1.08 & 66.99$\pm$1.25          & 66.89$\pm$1.03            & 66.95$\pm$1.85           & 65.29$\pm$1.99 ($\uparrow 4.17$)  & 68.25$\pm$1.34 ($\uparrow 1.36$) & 68.89$\pm$1.19 ($\uparrow 1.94$) \\
                            & 20                   & 60.26$\pm$1.44               & 61.85$\pm$1.35             & 62.93$\pm$1.01 & 64.66$\pm$1.35  & 63.99$\pm$1.44   & 64.00$\pm$1.01    & 65.58$\pm$1.55 & 65.35$\pm$1.44          & 66.37$\pm$1.04          & 65.89$\pm$1.25             & 63.55$\pm$1.64 ($\uparrow 3.29$)  & 68.17$\pm$1.44 ($\uparrow 1.80$) & 68.44$\pm$1.75 ($\uparrow 2.55$) \\
                            & 25                   & 59.02$\pm$1.39               & 59.10$\pm$1.24             & 60.87$\pm$1.37 & 63.22$\pm$1.24  & 63.01$\pm$1.26   & 62.85$\pm$1.37    & 64.28$\pm$1.23 & 64.25$\pm$1.23          & 65.11$\pm$1.79          & 64.93$\pm$1.70             & 63.02$\pm$1.28 ($\uparrow 4.00$)  & 65.24$\pm$1.26 ($\uparrow 0.13$) & 66.59$\pm$1.01 ($\uparrow 1.66$) \\\midrule
\multirow{6}{*}{Polblogs}   & 0                    & 95.69$\pm$0.38               & 95.35$\pm$0.20             & 95.22$\pm$0.14  & 95.31$\pm$0.18  & 85.13$\pm$1.22  & 90.13$\pm$1.22    & 95.36$\pm$0.82 & 95.13$\pm$0.85           & 94.77$\pm$1.07     & 93.20$\pm$0.64 & 96.13$\pm$0.14 ($\uparrow 0.44$)  &  94.95$\pm$1.22 ($\uparrow 0.18$) & 95.70$\pm$1.22 ($\uparrow 2.50$) \\
                            & 5                    & 87.69$\pm$0.38               & 88.35$\pm$0.20             & 89.22$\pm$0.14 & 90.15$\pm$0.18  & 85.01$\pm$1.22   & 90.31$\pm$0.14    & 89.35$\pm$0.32 & 90.05$\pm$0.20          & 90.31$\pm$1.31            & 90.85$\pm$1.39           & 89.72$\pm$0.25 ($\uparrow 2.03$)  & 92.27$\pm$1.22 ($\uparrow 1.96$) & 92.36$\pm$0.07 ($\uparrow 1.51$) \\
                            & 10                   & 84.64$\pm$1.69               & 85.88$\pm$1.40             & 85.85$\pm$1.91 & 86.02$\pm$1.33  & 84.56$\pm$1.66   & 87.00$\pm$1.91    & 87.33$\pm$1.74 & 86.68$\pm$1.36          & 86.71$\pm$1.27            & 87.09$\pm$1.41           & 87.71$\pm$0.65 ($\uparrow 3.07$)  & 88.75$\pm$1.66 ($\uparrow 2.04$) & 89.33$\pm$0.23 ($\uparrow 2.24$) \\
                            & 15                   & 71.55$\pm$1.70               & 72.01$\pm$0.20             & 72.10$\pm$1.80 & 78.18$\pm$1.89  & 79.89$\pm$1.75   & 81.95$\pm$1.80    & 82.04$\pm$1.47 & 82.12$\pm$1.99          & 82.26$\pm$1.55            & 83.31$\pm$1.87           & 76.58$\pm$0.56 ($\uparrow 5.03$)  & 84.11$\pm$1.75 ($\uparrow 1.85$) & 84.27$\pm$0.27 ($\uparrow 0.96$) \\
                            & 20                   & 65.00$\pm$1.03               & 67.12$\pm$1.21             & 67.32$\pm$1.15 & 69.35$\pm$1.32  & 71.25$\pm$1.82   & 71.51$\pm$1.15    & 69.98$\pm$1.22 & 69.42$\pm$1.39          & 72.59$\pm$1.12          & 72.45$\pm$1.69             & 69.44$\pm$0.66 ($\uparrow 4.44$)  & 73.26$\pm$1.82 ($\uparrow 0.67$) & 73.81$\pm$0.26 ($\uparrow 1.36$) \\
                            & 25                   & 64.02$\pm$1.35               & 64.16$\pm$1.13             & 66.10$\pm$1.11 & 67.20$\pm$1.44  & 68.39$\pm$1.01   & 69.02$\pm$1.11    & 66.55$\pm$1.79 & 68.22$\pm$1.47          & 70.80$\pm$1.33          & 69.37$\pm$1.01             & 67.92$\pm$1.43 ($\uparrow 3.90$)  & 71.92$\pm$1.01 ($\uparrow 1.12$) & 72.21$\pm$0.43 ($\uparrow 2.84$) \\ \midrule
\multirow{6}{*}{Pubmed}     & 0                    & 87.19$\pm$0.09               & 83.73$\pm$0.40             & 86.16$\pm$0.18  & 83.44$\pm$0.21  & 84.02$\pm$0.26  & 87.06$\pm$0.06    & 86.43$\pm$0.72 & 86.74$\pm$0.15         & 85.08$\pm$0.20 & 87.33$\pm$0.18 & 88.10$\pm$0.18 ($\uparrow 0.91$)  & 87.15$\pm$0.11 ($\uparrow 2.07$) & 88.41$\pm$0.26 ($\uparrow 1.08$)\\
                            & 5                    & 79.19$\pm$0.09               & 80.73$\pm$0.40             & 80.16$\pm$0.18 & 81.44$\pm$0.21  & 80.02$\pm$0.26   & 82.44$\pm$0.18    & 82.96$\pm$0.86 & 82.42$\pm$0.21          & 82.17$\pm$1.02            & 83.13$\pm$1.67           & 81.32$\pm$1.08 ($\uparrow 2.13$)  & 83.15$\pm$0.26 ($\uparrow 0.98$) & 84.33$\pm$1.71 ($\uparrow 1.20$) \\
                            & 10                   & 75.42$\pm$1.40               & 76.85$\pm$1.59             & 77.93$\pm$1.40 & 80.25$\pm$1.34  & 78.40$\pm$1.70   & 80.45$\pm$1.40    & 80.74$\pm$1.66 & 80.81$\pm$1.30          & 80.59$\pm$1.84            & 81.26$\pm$1.59           & 78.22$\pm$1.43 ($\uparrow 2.80$)  & 82.10$\pm$1.70 ($\uparrow 1.51$) & 82.74$\pm$1.14 ($\uparrow 1.48$) \\
                            & 15                   & 71.96$\pm$1.55               & 72.34$\pm$1.55             & 75.75$\pm$1.60 & 75.21$\pm$1.22  & 73.97$\pm$1.76   & 76.89$\pm$1.60    & 76.87$\pm$1.82 & 76.90$\pm$1.25          & 77.31$\pm$1.35          & 77.09$\pm$1.32             & 75.47$\pm$1.34 ($\uparrow 3.51$)  & 78.49$\pm$1.76 ($\uparrow 1.18$) & 78.88$\pm$1.60 ($\uparrow 1.79$) \\
                            & 20                   & 68.46$\pm$0.65               & 68.99$\pm$1.72             & 70.03$\pm$1.49 & 72.03$\pm$1.52  & 71.32$\pm$1.60   & 71.62$\pm$1.49    & 71.28$\pm$1.58 & 73.02$\pm$1.52          & 73.20$\pm$1.75          & 72.85$\pm$1.14             & 73.91$\pm$0.90 ($\uparrow 5.45$)  & 74.81$\pm$1.60 ($\uparrow 1.61$) & 74.90$\pm$1.32 ($\uparrow 2.05$) \\
                            & 25                   & 69.01$\pm$1.03               & 68.16$\pm$1.01             & 69.02$\pm$1.18 & 71.02$\pm$1.32  & 70.01$\pm$1.51   & 70.86$\pm$1.18    & 70.76$\pm$1.85 & 72.85$\pm$1.31          & 72.69$\pm$1.12          & 71.66$\pm$1.67             & 72.43$\pm$1.03 ($\uparrow 3.42$)  & 73.96$\pm$1.51 ($\uparrow 1.27$) & 74.24$\pm$1.39 ($\uparrow 2.58$) \\
\bottomrule
\end{tabular}

}
\end{table*}

\subsection{Quality Evaluation of the Augmented Graph}
\label{sec:quality_sup}
This paper introduces a novel graph diffusion model named RGDM, which synthesizes the graph structures. The synthetic graph structures generated by the RGDM are subsequently combined with the original attacked graph to form an augmented graph. By incorporating an edge selection mechanism in the RGDM, we aim to improve the robustness of the GNNs trained on the augmented graphs against the graph adversarial attacks. In Section~\ref{sec:results}, we have shown that both the vanilla GCN and the robust GNNs trained on the augmented graph achieve significantly better performance for semi-supervised node classification under graph adversarial attacks. In this section, we directly evaluate the data quality of the synthetic graph structures generated by the RGDM. In the visual domain, the Frechet Inception Distance (FID) is a widely used metric to evaluate the quality of the synthetic images generated by the generative models~\citep{BrockDS19, DDPM}. The FID score measures the similarity between the distribution of the generated images and the distribution of the real images. To compute the FID score, the pre-trained Inception v3~\citep{szegedy2016rethinking} is used to extract the features from both the real images and the generated images, which are then modeled as the multivariate Gaussian distributions. The FID score is then calculated using the Frechet Distance (FD)~\citep{BrockDS19} between the two multivariate Gaussian distributions modeling the real and the generated images~\citep{dowson1982frechet}. A lower FID score indicates that the generated images are more similar to the real images, suggesting better quality.

\noindent\textbf{Quality Evaluation of the Synthetic Nodes.} Although the Inception model cannot be applied to the graph data, we can replace the Inception model in the computation of the FID score with the pre-trained GCN~\citep{kipf2017semi} for extracting node features to adapt the metric to evaluate the quality of synthetic nodes generated by the RGDM. To this end, we define the Frechet Node Distance (FND), which is the FD between the multivariate Gaussians modeling the node features extracted by pre-trained GCN.
We randomly split the nodes in the original clean graph into two partitions of equal size, which are the base partition and the test partition.
To mitigate the influence of the randomness, we compute the FND scores with $10$ different random splits and report the mean and the standard deviation of the FND scores across different runs.
The FND computed between the nodes in the test partition and the base partition of the original clean graph establishes the baseline of the expected FND score for high-quality nodes.
By computing the FND score between the features of nodes in the synthetic graph structures generated by the RGDM and the features of nodes in the base partition of the original clean graph, we evaluate the quality of the synthetic nodes in the synthetic graph structures.
For simplicity, we refer to the FND score for the synthetic nodes as the FND between their features and the features of the nodes in the base partition of the original clean graph.

To show the effectiveness of the robustness loss in the training of the RGAE for the RGDM, we also compute the FND for the nodes in the synthetic graph structures generated by the RGDM without the node robustness loss and the edge robustness loss.
To demonstrate the advantages of the RGDM over the vanilla diffusion model, the DDPM~\citep{DDPM}, we train a baseline DDPM model on the input node attributes of the original attacked graph and synthesize the same number of synthetic nodes as the RGDM. The synthetic edges are then generated by connecting each synthetic node to its K-nearest neighbors in the original attacked graph using the K-nearest neighbors (KNN) algorithm with $K=\ceil{d_{\textup{ave}}}$, where $d_{\textup{ave}}$ is the average degree of the original clean graph. The synthetic graph structures, including the synthetic nodes and edges generated by the baseline DDPM model, are combined with the original attacked graph to form the augmented graph. Next, we compute the FND score for the nodes in the synthetic graph structures generated by the baseline DDPM model.
In addition, we compute FND for nodes generated by the four baseline node augmentation methods, which are the iGraphMix~\citep{jeong2024igraphmix}, ImGAGN~\citep{SGS-GAN-qu2021imgagn}, Graphsmote~\citep{SGS-GAN-zhao2021graphsmote}, and SNS~\citep{gao2023semantic}.
We also calculate the FND score for nodes in the original attacked graph. We use the same GCN~\citep{kipf2017semi} pre-trained on the original clean graph to extract the node features for computing the FND score. The ablation study is performed on the original attacked graph perturbed by the Metattack with a perturbation rate of $25\%$.

\begin{table}[!htbp]
\vspace{-2mm}
        \centering
        \caption{Frechet Node Distance (FND) to the nodes in the base partition of the original clean graph. The mean and standard deviation of the FND scores computed with $10$ different random splits of the base partition and the test partition in the original clean graph are reported.
        The evaluation is performed for Metattack with a perturbation rate of $25\%$. The FND score for the original clean graph is computed between the nodes in the test partition and the nodes in the base partition of the original clean graph.}

        \resizebox{0.775\columnwidth}{!}{
        \begin{tabular}{l|cccc}
                \toprule
                Data & Cora & Citeseer & Polblogs& Pubmed\\
                \midrule
                Original Attacked Graph                  & 12.10$\pm$0.45  & 7.53$\pm$0.22 & 8.26$\pm$0.37  & 6.11$\pm$0.38 \\
                iGraphMix~\citep{jeong2024igraphmix}   & 13.44$\pm$0.33  & 9.02$\pm$0.42 & 10.75$\pm$0.44 & 7.65$\pm$0.38 \\
                ImGAGN~\citep{SGS-GAN-qu2021imgagn}   & 13.10$\pm$0.38  & 7.96$\pm$0.42 & 9.86$\pm$0.43 & 6.95$\pm$0.27 \\
                Graphsmote~\citep{SGS-GAN-zhao2021graphsmote}   & 13.52$\pm$0.42  & 8.95$\pm$0.22 & 10.60$\pm$0.27 & 7.45$\pm$0.51 \\
                SNS~\citep{gao2023semantic}   & 13.35$\pm$0.42  & 7.88$\pm$0.27 & 9.30$\pm$0.85 & 6.74$\pm$0.41 \\
                Baseline DDPM                   & 13.28$\pm$0.40  & 8.82$\pm$0.26 & 10.25$\pm$0.39 & 7.25$\pm$0.41 \\
                RGDM (w/o Robustness Loss)      & 11.17$\pm$0.46  & 6.15$\pm$0.31 & 8.06$\pm$0.44  & 5.89$\pm$0.35 \\
                RGDM (w/o Node Robustness Loss) & 10.02$\pm$0.38  & 5.89$\pm$0.28 & 6.95$\pm$0.35  & \underline{5.01$\pm$0.36} \\
                RGDM (w/o Edge Robustness Loss) & \underline{9.67$\pm$0.27}  & \underline{5.18$\pm$0.49}  & \underline{6.65$\pm$0.25} & 5.11$\pm$0.38 \\
                RGDM                            & \textbf{8.16$\pm$0.29}     & \textbf{4.44$\pm$0.32} &\textbf{4.35$\pm$0.18} & \textbf{3.95$\pm$0.43}\\
                \midrule
                Original Clean Graph                     & 7.90$\pm$0.31              &  4.28$\pm$0.25         & 4.16$\pm$0.27         & 3.79$\pm$0.35\\
                \bottomrule
        \end{tabular}

        }
        \label{tab:quality}
        \vspace{-2mm}
\end{table}

The evaluation results are shown in Table~\ref{tab:quality}, where the lower FND scores indicate the node features are more similar to the features of nodes in the base partition of the original clean graph.
The FND score of the nodes in the synthetic graph structures generated by the RGDM is closest to the FND score of the original clean graph, which demonstrates that the RGDM generates faithful synthetic nodes. For example, the FND score for the RGDM on Pubmed is $3.95$, which is only $4.2\%$ higher than $3.79$, which is the FND score of the original clean graph.
In addition, we have several key observations from Table~\ref{tab:quality}. First, the synthetic nodes generated by the baseline DDPM are even worse than the original attacked graph. This is because the DDPM is trained on the original attacked graph without any robustness adaptations, and the noises in the original attacked graph are propagated to the synthetic nodes in the augmented graph. Second, the robustness loss is critical in generating clean augmented graphs. Without either the node robustness loss or the edge robustness loss, the synthetic nodes generated by the ablation models are worse than the synthetic nodes generated by the RGDM. This observation proves that both the node robustness loss and the edge robustness loss are beneficial for improving the quality of the synthetic nodes in the synthetic graph structures generated by the RGDM, as mentioned in Section~\ref{sec:robust_GAE}.
Third, the synthetic nodes generated by RGDM are significantly more faithful than those generated by the baseline node-level graph augmentation methods.

\noindent\textbf{Quality Evaluation of the Synthetic Edges.}
Similar to the design of the FND score for evaluating the quality of synthetic nodes, we replace the Inception model in the computation of FID with the pre-trained GNN~\citep{NBFNet} for edge feature extraction to adapt the metric to evaluate the quality of the synthetic edges generated by the RGDM. To this end, we define Frechet Edge Distance (FED), which is the FD between the multivariate Gaussians modeling the edge features extracted by pre-trained GNN.
Similar to the evaluation of the FND, we randomly split the edges in the original clean graph into two partitions of equal size, which are the base partition and the test partition.
To mitigate the influence of the randomness, we compute the FED scores with $10$ different random splits and report the mean and the standard deviation of the FED across different runs.
The FED computed between the edges in the test partition and the edges in the base partition of the original clean graph establishes the baseline of the expected FED score for the high-quality edges.
By computing the FED between the features of edges in the synthetic graph structures generated by the RGDM and the features of edges in the base partition of the original clean graph, we evaluate the quality of the edges in the synthetic graph structures. For simplicity, we refer to the FED score for the synthetic edges as the FED between their features and the features of edges in the base partition of the original clean graph.
Similar to the evaluation of the synthetic nodes, we also compute the FED score for the edges in the synthetic graph structures generated by the RGDM without the node robustness loss or the edge robustness loss. We compute the FED score for the edges in the synthetic graph structures generated by the baseline DDPM model. Since the edges in the synthetic graph structures are generated by the KNN algorithm, we evaluate the baseline DDPM models using different values of $K$ from $\set{\ceil{d_{\textup{ave}}/4}, \ceil{d_{\textup{ave}}/2}, \ceil{d_{\textup{ave}}}, \ceil{2\times d_{\textup{ave}}}, \ceil{4\times d_{\textup{ave}}}}$.
In addition, we compute FED for edges generated by the four baseline node augmentation methods, which are the iGraphMix~\citep{jeong2024igraphmix}, ImGAGN~\citep{SGS-GAN-qu2021imgagn}, Graphsmote~\citep{SGS-GAN-zhao2021graphsmote}, and SNS~\citep{gao2023semantic}.
We also calculate the FED score for nodes in the original attacked graph.
The FED score for edges in the original attacked graph is also computed. We use the same NBFNet~\citep{NBFNet} pre-trained on the original clean graph to extract the edge features for computing the FED score. The ablation study is performed on the original attacked graph perturbed by the Metattack with a perturbation rate of $25\%$.

\begin{table}[!htbp]
        \centering
        \vspace{-2mm}
        \caption{Frechet Edge Distance (FED) to the edges in the base partition of the original clean graph. The mean and standard deviation of the FED scores computed with $10$ different random splits of the base partition and the test partition in the original clean graph are reported. The evaluation is performed for Metattack with a perturbation rate of $25\%$. The FED score for the original clean graph is computed between the edges in the test partition and the edges in the base partition of the original clean graph.}

        \resizebox{0.775\columnwidth}{!}{
        \begin{tabular}{l|cccc}
                \toprule
                Data & Cora & Citeseer & Polblogs & Pubmed\\
                \midrule
                Original Attacked Graph                & 10.32$\pm$0.42 & 7.83$\pm$0.28 & 8.92$\pm$0.39 & 6.53$\pm$0.37 \\
                iGraphMix~\citep{jeong2024igraphmix} & 11.25$\pm$0.28 & 8.42$\pm$0.43 & 9.63$\pm$0.44  & 7.11$\pm$0.27 \\
                ImGAGN~\citep{SGS-GAN-qu2021imgagn} & 11.05$\pm$0.38 & 8.04$\pm$0.26 &8.21$\pm$0.51  & 7.16$\pm$0.79 \\
                Graphsmote~\citep{SGS-GAN-zhao2021graphsmote} & 11.13$\pm$0.29 & 8.16$\pm$0.51 & 9.32$\pm$0.41 & 7.18$\pm$0.44 \\
                SNS~\citep{gao2023semantic} & 11.05$\pm$0.38 & 8.04$\pm$0.26 &8.21$\pm$0.51  & 7.16$\pm$0.79 \\
                Baseline DDPM ($K=\ceil{d_{\textup{ave}}/4}$)   & 12.03$\pm$0.45 & 8.72$\pm$0.31 & 10.77$\pm$0.40 & 7.93$\pm$0.42 \\
                Baseline DDPM ($K=\ceil{d_{\textup{ave}}/2}$)   & 11.27$\pm$0.38 & 8.49$\pm$0.27 & 9.32$\pm$0.35  & 7.28$\pm$0.34 \\
                Baseline DDPM ($K=\ceil{d_{\textup{ave}}}$)     & 10.39$\pm$0.40 & 7.96$\pm$0.30 & 8.97$\pm$0.37  & 6.83$\pm$0.33 \\
                Baseline DDPM ($K=\ceil{2\times d_{\textup{ave}}}$) & 10.47$\pm$0.36 & 7.93$\pm$0.29 & 9.21$\pm$0.36  & 6.91$\pm$0.35 \\
                Baseline DDPM ($K=\ceil{4\times d_{\textup{ave}}}$) & 10.88$\pm$0.41 & 8.01$\pm$0.32 & 9.38$\pm$0.38  & 7.08$\pm$0.38 \\
                RGDM (w/o Robustness Loss)    & 10.07$\pm$0.39 & 7.64$\pm$0.26 & 8.74$\pm$0.33  & 6.27$\pm$0.31 \\
                RGDM (w/o Node Robustness Loss) & \underline{8.87$\pm$0.35} & \underline{6.73$\pm$0.25} & 7.93$\pm$0.31 & \underline{5.78$\pm$0.32} \\
                RGDM (w/o Edge Robustness Loss) & 8.98$\pm$0.34  & 6.82$\pm$0.27 & \underline{7.72$\pm$0.29} & 5.91$\pm$0.30 \\
                RGDM                          & \textbf{8.34$\pm$0.29} & \textbf{5.28$\pm$0.22} & \textbf{6.61$\pm$0.25} & \textbf{5.34$\pm$0.28} \\
                \midrule
                Original Clean Graph                   & 8.03$\pm$0.28  & 5.10$\pm$0.23 & 6.31$\pm$0.24  & 5.15$\pm$0.25 \\
                \bottomrule
        \end{tabular}
        }
        \label{tab:quality_edge}
\vspace{-2mm}
\end{table}

The evaluation results are shown in Table~\ref{tab:quality_edge}, where the lower FED scores indicate the edge features are more similar to the features of edges in the base partition of the original clean graph.
The FED score of the edges in the synthetic graph structures generated by the RGDM is closest to the FED score of the original clean graph, which demonstrates that the RGDM generates faithful synthetic edges. For example, the FED score for the RGDM on Pubmed is $5.34$, which is only $3.7\%$ higher than $5.15$, which is the FED score of the original clean graph.
In addition, we have similar observations from Table~\ref{tab:quality_edge} as those from Table~\ref{tab:quality}. First, the synthetic edges generated by the baseline DDPM are even worse than the original attacked graph. This is because the DDPM is trained on the original attacked graph without any robustness adaptations, and the simple edge generation method using the KNN on top of the synthetic nodes generated by the DDPM is vulnerable to noises in the original attacked graph. Edge generation with smaller $K$ in the KNN algorithm is more vulnerable to the noises in the original attacked graph. Although increasing the value of $K$ improves the quality of the synthetic edges, the edges generated by the KNN on top of the nodes generated by the baseline DDPM are still worse than the original attacked graphs. Second, the robustness loss is critical in generating clean augmented graphs. Without either the node robustness loss or the edge robustness loss, the synthetic edges generated by the ablation models are worse than the synthetic edges generated by the RGDM. This observation proves that both the node robustness loss and the edge robustness loss are beneficial for improving the quality of synthetic edges in the synthetic graph structures generated by the RGDM, as mentioned in Section~\ref{sec:robust_GAE}.

The synthetic node and edge distributions have FND/FED values close to the clean references in Tables~\ref{tab:quality} and~\ref{tab:quality_edge}. This is empirical evidence of fidelity, rather than a guarantee that generated structures are free of errors. The role of the normalized perturbation density is specified below.

\subsection{Normalized Perturbation Density and the Scope of Augmentation}
\label{sec:perturbation_density}
The main augmentation procedure retains the original attacked graph. Let $b$ denote the number of original edge perturbations, $E_{\textup{att}}$ its edge count, and $E_{\textup{syn}}$ the number of added synthetic edges, counting each undirected edge once. The descriptive normalized densities are
\bal\label{eq:perturbation_density}
\rho_{\textup{att}}\defeq\frac{b}{E_{\textup{att}}},\qquad
\rho_{\textup{aug}}\defeq\frac{b}{E_{\textup{att}}+E_{\textup{syn}}}.
\eal
Equation~(\ref{eq:perturbation_density}) keeps the original perturbation count fixed and changes only the denominator. The reported $25\%$ to $20.6\%$ Citeseer example adds $785$ synthetic edges. The Cora example adds $912$ and yields approximately $21.1\%$. These are descriptive normalizations, not estimates of the proportion of all harmful edges after generation. An edge flip budget can include deletions, and synthetic edges can themselves contain errors. In particular, a lower normalized density neither proves that original adversarial edges have been removed nor establishes improved robustness by itself.
The proposed mechanism is additional labeled supervision and relatively clean message passing paths, supported empirically by the fidelity and controlled classifier comparisons. Edge selection acts during generator learning and reconstruction, while the original attacked block $\bA$ remains unchanged in Equation~(\ref{eq:augmented_graph}). RGDM (Purification) in Section~\ref{sec:sup_rgdm_purification} is a separate intervention that reconstructs the original graph.

\subsection{Evaluation on Training Time and Synthetic Data Generation Time}
\label{sec:time}
In this section, we detail the training and synthetic data generation times for the Robust Graph Diffusion Model (RGDM) across various datasets in Table~\ref{tab:training_time}. The training time for the RGAE and the LDM in the RGDM using a single NVIDIA A100 GPU is measured in minutes. In addition, we present the time for generating synthetic graph structures by the RGDM in seconds per synthetic node generated in Table~\ref{tab:training_time} as well.

\noindent\textbf{Controlled decoder comparison.}
Table~\ref{tab:decoder_timing_full} compares SHED with a traditional global decoder that reconstructs a graph sized neighbor vector, using the reported single A100 setup. RGAE training is measured in minutes. Synthetic node generation is measured in seconds per node and includes both LDM sampling and decoding. Neither quantity is downstream GNN inference time. The training speedups are $480/153=3.14$ on Pubmed and $2511/452=5.56$ on ogbn-arxiv. Their end to end generation speedups are $0.080/0.073=1.10$ and $0.49/0.15=3.27$. LDM sampling limits the generation time improvement on smaller graphs, so a decoder operation count reduction does not imply the same wall clock speedup for the entire pipeline.
\begin{table}[!htbp]
\centering\small
\caption{SHED versus a global decoder on one NVIDIA A100. The generation measurement includes LDM sampling and decoding.}
\label{tab:decoder_timing_full}
\begin{tabular}{lrrrr}
\toprule
&\multicolumn{2}{c}{RGAE training (min)}&\multicolumn{2}{c}{Generation (s/node)}\\
Dataset&SHED&Global&SHED&Global\\
\midrule
Cora&20&29&0.066&0.089\\
Citeseer&23&30&0.067&0.086\\
Polblogs&65&88&0.069&0.090\\
Pubmed&153&480&0.073&0.080\\
ogbn-arxiv&452&2511&0.15&0.49\\
\bottomrule
\end{tabular}
\end{table}

\begin{table}[!htbp]
\centering
\caption{Time for the training of the RGDM and data generation with the RGDM on different datasets.}

\resizebox{0.7\columnwidth}{!}{
\begin{tabular}{cccc}
\toprule
\multirow{2}{*}{Datasets} & \multicolumn{2}{c}{RGDM Training Time (minutes)} & \multirow{2}{*}{Generation Time (s/sample)} \\ \cmidrule{2-3}
                          & RGAE               & LDM              &                                             \\ \hline
Cora                      & 20                       & 39               & 0.066                                       \\
Citeseer                  & 23                       & 41               & 0.067                                       \\
Polblogs                  & 65                      & 31               & 0.069                                       \\
Pubmed                    & 153                      & 154              & 0.073                                       \\ \bottomrule
\end{tabular}
}
\label{tab:training_time}
\end{table}
\vspace{-3mm}

\begin{table}[!htbp]
\centering
\caption{Time for selecting the optimal number of synthetic nodes $N'$ with cross-validation.}

\resizebox{0.675\columnwidth}{!}{
\begin{tabular}{cccc}
\toprule
\multirow{2}{*}{Datasets} & \multicolumn{3}{c}{Cross-validation Time (minutes)} \\ \cmidrule{2-4}
                          & RGDM (GCN)        & RGDM (SG-GSR)       & RGDM (Pro-GNN)     \\ \midrule
Cora                      & 1.1       & 3.5       &  43.3     \\
Citeseer                  & 1.0       & 3.4       &  38.5     \\
Polblogs                  & 1.6       & 3.9       &  26.9     \\
Pubmed                    & 9.4       & 20.5      &  176.5     \\ \bottomrule
\end{tabular}
}
\label{tab:cv}
\end{table}

\begin{table}[!htbp]
\small
\centering
\caption{The number of synthetic nodes $N'$ ($\times\abth{\cV_L}$) selected for Metattack, where $\cV_L$ is the labeled nodes of the original graph.}

\label{tab:metattack_synnum}
\resizebox{0.7\columnwidth}{!}{
\begin{tabular}{c|c|ccc}
\toprule
Dataset        & Ptb Rate (\%)  & RGDM (GCN)             & RGDM (SG-GSR)           &RGDM (Pro-GNN)                      \\ \midrule
\multirow{6}{*}{Cora}     & 0   & $2 \times\abth{\cV_L}$      &  $1 \times\abth{\cV_L}$    & $2  \times\abth{\cV_L}$                                  \\
                          & 5   & $2 \times\abth{\cV_L}$      &  $1 \times\abth{\cV_L}$    & $2  \times\abth{\cV_L}$                                  \\
                          & 10  & $2 \times\abth{\cV_L}$      &  $1 \times\abth{\cV_L}$    & $2  \times\abth{\cV_L}$                                  \\
                          & 15  & $3 \times\abth{\cV_L}$      &  $1 \times\abth{\cV_L}$    & $3  \times\abth{\cV_L}$                                  \\
                          & 20  & $3 \times\abth{\cV_L}$      &  $2 \times\abth{\cV_L}$    & $3  \times\abth{\cV_L}$                                  \\
                          & 25  & $5 \times\abth{\cV_L}$      &  $5 \times\abth{\cV_L}$    & $5  \times\abth{\cV_L}$                                  \\\midrule
\multirow{6}{*}{Citeseer} & 0   & $1 \times\abth{\cV_L}$      &  $1 \times\abth{\cV_L}$    & $1  \times\abth{\cV_L}$                                  \\
                          & 5   & $2 \times\abth{\cV_L}$      &  $2 \times\abth{\cV_L}$    & $2  \times\abth{\cV_L}$                                  \\
                          & 10  & $1 \times\abth{\cV_L}$      &  $1 \times\abth{\cV_L}$    & $2  \times\abth{\cV_L}$                                  \\
                          & 15  & $3 \times\abth{\cV_L}$      &  $2 \times\abth{\cV_L}$    & $3  \times\abth{\cV_L}$                                  \\
                          & 20  & $3 \times\abth{\cV_L}$      &  $3 \times\abth{\cV_L}$    & $3  \times\abth{\cV_L}$                                  \\
                          & 25  & $4 \times\abth{\cV_L}$      &  $5 \times\abth{\cV_L}$    & $4  \times\abth{\cV_L}$                                  \\\midrule
\multirow{6}{*}{Polblogs} & 0   & $1 \times\abth{\cV_L}$      &  $1 \times\abth{\cV_L}$    & $1  \times\abth{\cV_L}$                                   \\
                          & 5   & $1 \times\abth{\cV_L}$      &  $1 \times\abth{\cV_L}$    & $1  \times\abth{\cV_L}$                                   \\
                          & 10  & $1 \times\abth{\cV_L}$      &  $2 \times\abth{\cV_L}$    & $1  \times\abth{\cV_L}$                                   \\
                          & 15  & $2 \times\abth{\cV_L}$      &  $1 \times\abth{\cV_L}$    & $2  \times\abth{\cV_L}$                                   \\
                          & 20  & $2 \times\abth{\cV_L}$      &  $1 \times\abth{\cV_L}$    & $2  \times\abth{\cV_L}$                                   \\
                          & 25  & $2 \times\abth{\cV_L}$      &  $2 \times\abth{\cV_L}$    & $3  \times\abth{\cV_L}$                                   \\\midrule
\multirow{6}{*}{Pubmed}   & 0   & $2 \times\abth{\cV_L}$      &  $1 \times\abth{\cV_L}$    & $2  \times\abth{\cV_L}$                                  \\
                          & 5   & $2 \times\abth{\cV_L}$      &  $1 \times\abth{\cV_L}$    & $2  \times\abth{\cV_L}$                                  \\
                          & 10  & $3 \times\abth{\cV_L}$      &  $1 \times\abth{\cV_L}$    & $2  \times\abth{\cV_L}$                                  \\
                          & 15  & $3 \times\abth{\cV_L}$      &  $1 \times\abth{\cV_L}$    & $2  \times\abth{\cV_L}$                                  \\
                          & 20  & $3 \times\abth{\cV_L}$      &  $1 \times\abth{\cV_L}$    & $3  \times\abth{\cV_L}$                                  \\
                          & 25  & $2 \times\abth{\cV_L}$      &  $1 \times\abth{\cV_L}$    & $3  \times\abth{\cV_L}$                                  \\ 
\bottomrule
\end{tabular}

}
\end{table}

\begin{table}[!ht]
\small
\centering
\caption{The number of synthetic nodes $N'$ ($\times\abth{\cV_L}$) for Nettack, where $\cV_L$ is the labeled nodes of the original graph.}

\label{tab:nettack_synnum}
\resizebox{0.7\columnwidth}{!}{
\begin{tabular}{c|c|ccc}
\toprule
Dataset        & Attack Budget  & RGDM (GCN)                      & RGDM (SG-GSR)             &RGDM (Pro-GNN)                  \\ \midrule
\multirow{6}{*}{Cora}     & 0    & $2 \times\abth{\cV_L}$                                 & $1 \times\abth{\cV_L}$                               & $2 \times\abth{\cV_L}$                                 \\
                          & 1    & $3 \times\abth{\cV_L}$                                 & $2 \times\abth{\cV_L}$                               & $2 \times\abth{\cV_L}$                                 \\
                          & 2    & $3 \times\abth{\cV_L}$                                 & $2 \times\abth{\cV_L}$                               & $2 \times\abth{\cV_L}$                                 \\
                          & 3    & $3 \times\abth{\cV_L}$                                 & $2 \times\abth{\cV_L}$                               & $3 \times\abth{\cV_L}$                                 \\
                          & 4    & $3 \times\abth{\cV_L}$                                 & $3 \times\abth{\cV_L}$                               & $3 \times\abth{\cV_L}$                                 \\
                          & 5    & $4 \times\abth{\cV_L}$                                 & $3 \times\abth{\cV_L}$                               & $3 \times\abth{\cV_L}$                                 \\\midrule
\multirow{6}{*}{Citeseer} & 0    & $2 \times\abth{\cV_L}$                                 & $1 \times\abth{\cV_L}$                               & $1 \times\abth{\cV_L}$                                 \\
                          & 1    & $2 \times\abth{\cV_L}$                                 & $2 \times\abth{\cV_L}$                               & $2 \times\abth{\cV_L}$                                 \\
                          & 2    & $1 \times\abth{\cV_L}$                                 & $2 \times\abth{\cV_L}$                               & $2 \times\abth{\cV_L}$                                 \\
                          & 3    & $3 \times\abth{\cV_L}$                                 & $3 \times\abth{\cV_L}$                               & $3 \times\abth{\cV_L}$                                 \\
                          & 4    & $3 \times\abth{\cV_L}$                                 & $2 \times\abth{\cV_L}$                               & $3 \times\abth{\cV_L}$                                 \\
                          & 5    & $3 \times\abth{\cV_L}$                                 & $3 \times\abth{\cV_L}$                               & $3 \times\abth{\cV_L}$                                 \\\midrule
\multirow{6}{*}{Polblogs} & 0    & $1 \times\abth{\cV_L}$                                 & $1 \times\abth{\cV_L}$                               & $1 \times\abth{\cV_L}$                                  \\
                          & 1    & $1 \times\abth{\cV_L}$                                 & $1 \times\abth{\cV_L}$                               & $1 \times\abth{\cV_L}$                                  \\
                          & 2    & $1 \times\abth{\cV_L}$                                 & $2 \times\abth{\cV_L}$                               & $1 \times\abth{\cV_L}$                                  \\
                          & 3    & $1 \times\abth{\cV_L}$                                 & $2 \times\abth{\cV_L}$                               & $2 \times\abth{\cV_L}$                                  \\
                          & 4    & $2 \times\abth{\cV_L}$                                 & $2 \times\abth{\cV_L}$                               & $2 \times\abth{\cV_L}$                                  \\
                          & 5    & $2 \times\abth{\cV_L}$                                 & $3 \times\abth{\cV_L}$                               & $3 \times\abth{\cV_L}$                                  \\\midrule
\multirow{6}{*}{Pubmed}   & 0    & $1 \times\abth{\cV_L}$                                 & $1 \times\abth{\cV_L}$                               & $2 \times\abth{\cV_L}$                                 \\
                          & 1    & $1 \times\abth{\cV_L}$                                 & $1 \times\abth{\cV_L}$                               & $2 \times\abth{\cV_L}$                                 \\
                          & 2    & $2 \times\abth{\cV_L}$                                 & $2 \times\abth{\cV_L}$                               & $2 \times\abth{\cV_L}$                                 \\
                          & 3    & $2 \times\abth{\cV_L}$                                 & $2 \times\abth{\cV_L}$                               & $2 \times\abth{\cV_L}$                                 \\
                          & 4    & $2 \times\abth{\cV_L}$                                 & $3 \times\abth{\cV_L}$                               & $3 \times\abth{\cV_L}$                                 \\
                          & 5    & $2 \times\abth{\cV_L}$                                 & $3 \times\abth{\cV_L}$                               & $3 \times\abth{\cV_L}$                                 \\ 
\bottomrule
\end{tabular}
}
\end{table}

\begin{table}[!ht]
\small
\centering
\caption{The number of synthetic nodes $N'$ ($\times\abth{\cV_L}$) for Topology Attack, where $\cV_L$ is the labeled nodes of the original graph.}

\label{tab:topology_synnum}
\resizebox{0.7\columnwidth}{!}{
\begin{tabular}{c|c|ccc}
\toprule
Dataset        & Ptb Rate (\%)                & RGDM (GCN)               & RGDM (SG-GSR)               &RGDM (Pro-GNN)                 \\ \midrule
\multirow{6}{*}{Cora}     & 0   & $1 \times\abth{\cV_L}$ & $1 \times\abth{\cV_L}$ & $1 \times\abth{\cV_L}$                                 \\
                          & 5   & $1 \times\abth{\cV_L}$ & $1 \times\abth{\cV_L}$ & $1 \times\abth{\cV_L}$                                 \\
                          & 10  & $2 \times\abth{\cV_L}$ & $2 \times\abth{\cV_L}$ & $2 \times\abth{\cV_L}$                                 \\
                          & 15  & $2 \times\abth{\cV_L}$ & $1 \times\abth{\cV_L}$ & $2 \times\abth{\cV_L}$                                 \\
                          & 20  & $2 \times\abth{\cV_L}$ & $2 \times\abth{\cV_L}$ & $3 \times\abth{\cV_L}$                                 \\
                          & 25  & $2 \times\abth{\cV_L}$ & $2 \times\abth{\cV_L}$ & $2 \times\abth{\cV_L}$                                 \\\midrule
\multirow{6}{*}{Citeseer} & 0   & $1 \times\abth{\cV_L}$ & $1 \times\abth{\cV_L}$ & $1 \times\abth{\cV_L}$                                 \\
                          & 5   & $2 \times\abth{\cV_L}$ & $1 \times\abth{\cV_L}$ & $1 \times\abth{\cV_L}$                                 \\
                          & 10  & $1 \times\abth{\cV_L}$ & $2 \times\abth{\cV_L}$ & $2 \times\abth{\cV_L}$                                 \\
                          & 15  & $2 \times\abth{\cV_L}$ & $2 \times\abth{\cV_L}$ & $3 \times\abth{\cV_L}$                                 \\
                          & 20  & $2 \times\abth{\cV_L}$ & $2 \times\abth{\cV_L}$ & $2 \times\abth{\cV_L}$                                 \\
                          & 25  & $3 \times\abth{\cV_L}$ & $2 \times\abth{\cV_L}$ & $2 \times\abth{\cV_L}$                                 \\\midrule
\multirow{6}{*}{Polblogs} & 0   & $1 \times\abth{\cV_L}$ & $1 \times\abth{\cV_L}$ & $1 \times\abth{\cV_L}$                                  \\
                          & 5   & $1 \times\abth{\cV_L}$ & $1 \times\abth{\cV_L}$ & $1 \times\abth{\cV_L}$                                  \\
                          & 10  & $1 \times\abth{\cV_L}$ & $2 \times\abth{\cV_L}$ & $2 \times\abth{\cV_L}$                                  \\
                          & 15  & $2 \times\abth{\cV_L}$ & $2 \times\abth{\cV_L}$ & $3 \times\abth{\cV_L}$                                  \\
                          & 20  & $2 \times\abth{\cV_L}$ & $3 \times\abth{\cV_L}$ & $1 \times\abth{\cV_L}$                                  \\
                          & 25  & $2 \times\abth{\cV_L}$ & $2 \times\abth{\cV_L}$ & $2 \times\abth{\cV_L}$                                  \\\midrule
\multirow{6}{*}{Pubmed}   & 0   & $1 \times\abth{\cV_L}$ & $1 \times\abth{\cV_L}$ & $2 \times\abth{\cV_L}$                                 \\
                          & 5   & $2 \times\abth{\cV_L}$ & $2 \times\abth{\cV_L}$ & $3 \times\abth{\cV_L}$                                 \\
                          & 10  & $2 \times\abth{\cV_L}$ & $2 \times\abth{\cV_L}$ & $2 \times\abth{\cV_L}$                                 \\
                          & 15  & $3 \times\abth{\cV_L}$ & $1 \times\abth{\cV_L}$ & $2 \times\abth{\cV_L}$                                 \\
                          & 20  & $2 \times\abth{\cV_L}$ & $3 \times\abth{\cV_L}$ & $4 \times\abth{\cV_L}$                                 \\
                          & 25  & $1 \times\abth{\cV_L}$ & $2 \times\abth{\cV_L}$ & $2 \times\abth{\cV_L}$                                 \\ 
\bottomrule
\end{tabular}

}
\end{table}

\subsection{Cross-validation on the Number of Synthetic Nodes}
\label{sec:cross_validation_sup}
In our experiments, we generate $10\times\abth{\cV_L}$ synthetic nodes, where $\cV_L$ is the training set of the original graph. Then, we search for the optimal size of synthetic nodes $N'$ added in the augmented graph from $1\times\abth{\cV_L}$ to $10\times\abth{\cV_L}$ with a step size of $\abth{\cV_L}$. In the cross-validation, we train different GNN classifiers with different numbers of synthetic data and set $N'$ to the one that achieves the best validation accuracy by the end of $40\%$ of their total epoch number. To study the effectiveness of the cross-validation process for selecting the optimal number of synthetic nodes, we calculate the complete cross-validation time of different models on different datasets. All the experiments are performed on one Nvidia A100 GPU. It is observed from the results in Table~\ref{tab:cv} that the cross-validation process does not largely increase the computation overhead. The numbers of synthetic nodes added to the augmented graphs for all the experiments are shown in Table~\ref{tab:metattack_synnum}, Table~\ref{tab:nettack_synnum}, and Table~\ref{tab:topology_synnum}.

\subsection{Adversarial Defense on Large Scale Graphs}
\label{sec:large-scale}
To evaluate the scalability of the proposed method, we conduct additional experiments on graph adversarial defense for large-scale graph datasets, which are the large-scale social network, Reddit~\citep{HamiltonYL17}, and the large-scale knowledge graph, ogbn-arxiv~\citep{hu2020open}. The statistics of the Reddit and ogbn-arxiv datasets are detailed in Table~\ref{dataset}. Since the Metattack~\citep{zugner_adversarial_2019}, the Nettack~\citep{zugner2018adversarial}, and the Topology Attack~\citep{xu2019topology} adopted in Section~\ref{sec:results} do not scale to large graphs at the size of Reddit and ogbn-arxiv, we use a scalable graph adversarial attack method, the improved DICE~\citep{li2022revisiting}, in our experiments on Reddit and ogbn-arxiv.
The experiments are performed for perturbation rates of $0\%$, $10\%$, and $20\%$ following the settings in~\citep{li2022revisiting} with GCN as the baseline model. The results are shown in Table~\ref{table:attack_large}. It is observed that RGDM consistently improves the performance of the GCN baseline across all perturbation levels on both datasets. For example, RGDM achieves an improvement of $2.25\%$ on ogbn-arxiv at the perturbation rate of $10\%$, indicating its effectiveness in defending against adversarial perturbations even on large and complex knowledge graphs.

\begin{table*}[!htbp]
\vspace{-3mm}
\small
\centering
\caption{Node classification performance (Accuracy$\pm$Std) under the adversarial attack by the improved DICE~\citep{li2022revisiting}. The improvements of the RGDM over the baseline are attached in parentheses. }
\label{table:attack_large}
\resizebox{0.6\textwidth}{!}{
\begin{tabular}{c|c|cccccccccccc|ccc}
\toprule
Dataset        & Ptb Rate (\%)  & GCN & RGDM (GCN) \\
\midrule
\multirow{3}{*}{ogbn-arxiv}	& 0  & 72.82$\pm$1.32   &    \textbf{74.48}  $\pm$1.15 ($\uparrow   1.66  $)  \\
							& 10 & 62.25$\pm$1.23   &    \textbf{64.50}  $\pm$1.35 ($\uparrow   2.25  $)   \\
							& 20 & 54.95$\pm$1.44   &    \textbf{56.17}  $\pm$1.66 ($\uparrow   1.22  $)  \\ \midrule
\multirow{3}{*}{Reddit} 	& 0  & 93.52$\pm$1.17   &    \textbf{95.44}  $\pm$1.42 ($\uparrow   1.92  $)  \\
							& 10 & 82.62$\pm$1.32   &    \textbf{84.73}  $\pm$1.12 ($\uparrow   2.11  $)  \\
							& 20 & 74.53$\pm$1.57   &    \textbf{75.99}  $\pm$1.22 ($\uparrow   1.46  $)  \\
\bottomrule
\end{tabular}
}
\end{table*}

\noindent\textbf{Edge selection at scale.}
Table~\ref{tab:large_edge_ablation} isolates the edge selection module at $20\%$ improved DICE, keeping the GCN backbone fixed. Without the module, the gains over GCN are only $0.35$ and $0.47$ points. Restoring it adds $0.87$ and $0.99$ points, respectively. This extends the four dataset sparsification ablation to both large graphs.
\begin{table}[!htbp]
\centering\small
\caption{Edge selection ablation on large graphs under $20\%$ improved DICE, with GCN fixed. Accuracy (\%).}
\label{tab:large_edge_ablation}
\begin{tabular}{lrr}
\toprule
Method&ogbn-arxiv&Reddit\\
\midrule
GCN&54.95&74.53\\
RGDM without edge selection&55.30&75.00\\
RGDM (full)&56.17&75.99\\
\bottomrule
\end{tabular}
\end{table}

\subsection{Sensitivity Analysis on the Hyper-Parameters}
\label{sec:sensitivity}
In this section, we conduct sensitivity analyses on the key hyper-parameters of RGDM, including the temperature parameter $\tau$ for edge selection and the number of clusters $K$ in SHED. The study is carried out using RGDM (Pro-GNN) on the Cora dataset under a $25\%$ Metattack setting. As shown in Table~\ref{tab:sensitivity_temp}, although the highest accuracy is achieved at $\tau=0.5$, the performance remains stable across different values of $\tau$. In particular, even the lowest performing setting, $\tau=0.1$, results in only a marginal $0.19\%$ decrease in accuracy compared to the best result.

\begin{table}[!ht]
\small
\centering
\caption{Sensitivity analysis on the temperature for edge selection $\tau$. The study is performed by using RGDM (Pro-GNN) on Cora with $25\%$ of Metattack.}
\label{tab:sensitivity_temp}
\resizebox{0.9\columnwidth}{!}{
\begin{tabular}{c|ccccccccc}
\hline
$\tau$ & 0.1   & 0.2   & 0.3   & 0.4   & 0.5   & 0.6   & 0.7   & 0.8   & 0.9  \\
\hline
Accuracy            & 72.12 & 72.16 & 72.30 & 72.29 & \textbf{72.31} & 72.28 & 72.24 & 72.22 & 72.25 \\
\hline

\end{tabular}

}
\end{table}
We further study the sensitivity of the proposed SHED module to the number of clusters $K$.
The study is conducted using RGDM (Pro-GNN) on the Cora dataset under a $25\%$ Metattack setting.
Since SHED partitions the graph into $K$ balanced clusters and sets the maximum cluster size as
$M=\lceil N/K\rceil$, changing $K$ affects both the granularity of hierarchical edge decoding and the
generation efficiency. As shown in Table~\ref{tab:sensitivity_cluster}, RGDM is not sensitive to the
choice of $K$. The best performance is achieved at $K=100$, which is the default setting used in our
experiments. Moreover, the accuracy remains stable when we vary $K$ from $20$ to $200$. Even the
lowest-performing setting only decreases the accuracy by $0.19\%$ compared with the best result.
In addition, the generation time remains close to the default setting, indicating that SHED provides
stable robustness and efficient generation across a wide range of cluster numbers.

\begin{table}[!ht]
\small
\centering
\caption{Sensitivity analysis on the number of clusters $K$ in SHED. The study is performed by using RGDM (Pro-GNN) on Cora with $25\%$ of Metattack.}
\label{tab:sensitivity_cluster}
\resizebox{0.9\columnwidth}{!}{
\begin{tabular}{c|cccccccc}
\hline
$K$ & 20 & 40 & 60 & 80 & 100 & 120 & 150 & 200 \\
\hline
$M=\lceil N/K\rceil$ & 125 & 63 & 42 & 32 & 25 & 21 & 17 & 13 \\
Generation Time (s/sample) & 0.075 & 0.070 & 0.068 & 0.067 & 0.066 & 0.067 & 0.068 & 0.071 \\
Accuracy & 72.12 & 72.20 & 72.26 & 72.29 & \textbf{72.31} & 72.28 & 72.24 & 72.20 \\
\hline
\end{tabular}
}
\end{table}

\subsection{Adversarial Defense on Heterophilic Graph}
\label{sec:heter}
In this section, we study the effectiveness of RGDM for adversarial defense on heterophilic graphs. Following~\citep{qiu2024refining}, we perform the study on Chameleon~\citep{pei2020geom} and Actor~\citep{tang2009social} for the defense against the Metattack~\citep{zugner_adversarial_2019}. The adversarially robust GNN, LHS~\citep{qiu2024refining}, which is specifically designed for heterophilic graphs, is adopted as the baseline model.  The results are shown in Table~\ref{table:metattack_heter}. RGDM improves LHS at each tested perturbation rate on both datasets. At $25\%$ Metattack, the improvements are $2.38$ points on Chameleon and $2.53$ points on Actor. These results support the cross validated soft penalty in the tested settings and do not imply that stronger low rank regularization always helps. The main text discussion and mean accuracies appear in Table~\ref{tab:heter_main}.

\begin{table*}[!htbp]
\vspace{-3mm}
\small
\centering
\caption{Node classification performance (Accuracy$\pm$Std) under the non-targeted poisoning attack (Metattack)~\citep{zugner_adversarial_2019}. The improvements of the RGDM over the baseline are attached in parentheses. }
\label{table:metattack_heter}
\resizebox{0.6\textwidth}{!}{
\begin{tabular}{c|c|cccccccccccc|ccc}
\toprule
Dataset        & Ptb Rate (\%)  & LHS & RGDM (LHS) \\
\midrule
\multirow{3}{*}{Chameleon} 	& 0  & 72.31$\pm$1.32   &   \textbf{74.02} $\pm$1.13 ($\uparrow  1.71 $)  \\
							& 15 & 71.87$\pm$1.61   &   \textbf{72.98} $\pm$1.11 ($\uparrow  1.11 $)   \\
							& 25 & 70.03$\pm$1.85   &   \textbf{72.41} $\pm$1.66 ($\uparrow  2.38 $)  \\ \midrule
\multirow{3}{*}{Actor} 		& 0  & 38.87$\pm$1.43   &   \textbf{41.80} $\pm$1.42 ($\uparrow  2.93 $)  \\
							& 15 & 36.79$\pm$1.11   &   \textbf{39.01} $\pm$1.12 ($\uparrow  2.22 $)  \\
							& 25 & 35.37$\pm$1.12   &   \textbf{37.90} $\pm$1.01 ($\uparrow  2.53 $)  \\
\bottomrule
\end{tabular}
}
\end{table*}

\subsection{t-SNE Visualization of the Augmented Graph}
\label{sec:tsne}
\begin{figure*}[!htbp]
  \centering
    \includegraphics[width=0.8\textwidth]{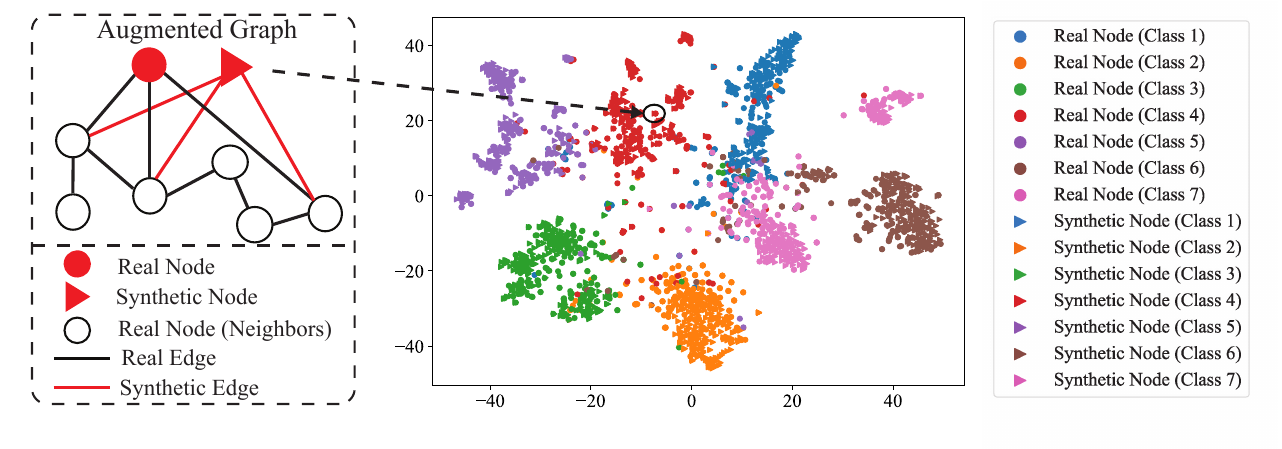}
    \vspace{-2mm}
  \caption{Illustration of the augmented graph after adding the synthetic graph structures into the original graph on Cora. The RGDM is trained on Cora under Metattack with a perturbation rate of $25\%$. The figure on the right illustrates the t-SNE visualization of the node features for both the real nodes in the original clean graph and the synthetic nodes. The figure on the left illustrates the 2-hop neighborhood of a real node and a synthetic node that have similar node features, evidenced by the t-SNE visualization in the augmented graph. It is observed that the synthetic node shares the same neighbors as the real node.}
  \vspace{-5mm}
  \label{fig:tsne}
\end{figure*}

Figure~\ref{fig:tsne} illustrates the t-SNE visualization of both the real and synthetic node features within the augmented graph, revealing that the synthetic nodes closely mimic the real nodes.

\subsection{Statistical Significance Analysis}
\label{sec:t-test}

We evaluate the paired gain $\Delta_s\defeq a^{\textup{RGDM}}_s-a^{\textup{base}}_s$ in percentage points for each shared attack seed $s\in[10]$. The sample mean and sample SD are defined by Equation~(\ref{eq:paired_statistics}):
\bal\label{eq:paired_statistics}
\bar\Delta\defeq\frac1{10}\sum_{s=1}^{10}\Delta_s,\qquad
s_{\Delta}\defeq\sqrt{\frac1{9}\sum_{s=1}^{10}(\Delta_s-\bar\Delta)^2},\qquad
t_{\textup{pair}}\defeq\frac{\sqrt{10}\bar\Delta}{s_{\Delta}}.
\eal
A two sided paired Student test compares $\abth{t_{\textup{pair}}}$ with a Student distribution with $9$ degrees of freedom, under the usual assumptions of independent runs and approximately normal paired differences. This tests within backbone augmentation, not differences between unrelated architectures. Table~\ref{tab:paired_gains} reports the summaries from the separate paired evaluation. Each uncertainty is an SD of differences, not an SD of accuracies or a standard error.
\begin{table}[!htbp]
\centering\small
\caption{Reported Metattack mean gain $\pm$ paired SD over $10$ shared attack seeds (percentage points). Only rows with reported paired SD are included.}
\label{tab:paired_gains}
\begin{tabular}{lrrrr}
\toprule
Dataset&Ptb (\%)&GCN&Pro-GNN&SG-GSR\\
\midrule
Cora&0&$1.7\pm0.3$&$1.3\pm0.3$&$0.4\pm0.1$\\
Cora&15&$4.3\pm0.5$&$1.9\pm0.4$&$1.5\pm0.3$\\
Cora&25&$13.9\pm0.7$&$2.6\pm0.4$&$1.3\pm0.3$\\
Citeseer&15&$4.2\pm0.5$&$1.7\pm0.4$&$2.9\pm0.4$\\
Citeseer&25&$5.9\pm0.5$&$1.2\pm0.3$&$2.6\pm0.4$\\
Polblogs&15&$4.9\pm0.5$&$2.0\pm0.4$&$1.1\pm0.3$\\
Polblogs&25&$7.3\pm0.6$&$3.7\pm0.5$&$1.0\pm0.3$\\
Pubmed&15&$6.3\pm0.5$&$0.9\pm0.2$&$1.4\pm0.3$\\
Pubmed&25&$6.0\pm0.5$&$0.8\pm0.2$&$1.4\pm0.3$\\
\bottomrule
\end{tabular}
\end{table}
\noindent\textbf{Precision of the significance statements.}
The separately reported significance levels in Table~\ref{table:t-test} are retained. For Metattack rows with paired summaries, they are stated as conservative upper thresholds, not exact p values reconstructed from rounded means and SD. This distinction is necessary: for example, the Cora $25\%$ SG-GSR summary $1.3\pm0.3$ gives $t_{\textup{pair}}\approx13.70$ in Equation~(\ref{eq:paired_statistics}), whose nominal two sided p value is smaller than the separately reported $0.008$. Reporting $p\le0.008$ preserves that significance level without identifying it as the exact p value of those rounded moments. No exact p values or confidence intervals are imputed from unreported runs. The reported significance levels at the strongest budgets are $0.006$ to $0.008$ for Metattack, $0.005$ to $0.006$ for Nettack, and $0.006$ to $0.007$ for Topology Attack. The latter two retain their reported values because paired SD were not supplied for them.
\begin{table}[!htbp]
\centering
\small
\caption{Reported significance levels with each backbone fixed: Metattack. For rows with paired summaries in Table~\ref{tab:paired_gains}, $\le$ denotes a conservative upper threshold, not an exact p value reconstructed from rounded moments. Remaining entries are reported p values.}
\label{table:t-test}
\begin{tabular}{lrrrr}
\toprule
Dataset & Ptb (\%) & RGDM (GCN) & RGDM (Pro-GNN) & RGDM (SG-GSR)\\
\midrule
\multirow{6}{*}{Cora} & 0 & $\le0.042$ & $\le0.027$ & $\le0.016$\\
 & 5 & 0.038 & 0.024 & 0.013\\
 & 10 & 0.033 & 0.021 & 0.012\\
 & 15 & $\le0.029$ & $\le0.017$ & $\le0.010$\\
 & 20 & 0.025 & 0.015 & 0.009\\
 & 25 & $\le0.021$ & $\le0.012$ & $\le0.008$\\
\midrule
\multirow{6}{*}{Citeseer} & 0 & 0.039 & 0.022 & 0.018\\
 & 5 & 0.035 & 0.020 & 0.015\\
 & 10 & 0.030 & 0.017 & 0.013\\
 & 15 & $\le0.027$ & $\le0.014$ & $\le0.011$\\
 & 20 & 0.023 & 0.012 & 0.009\\
 & 25 & $\le0.019$ & $\le0.010$ & $\le0.007$\\
\midrule
\multirow{6}{*}{Polblogs} & 0 & 0.041 & 0.023 & 0.019\\
 & 5 & 0.036 & 0.020 & 0.015\\
 & 10 & 0.031 & 0.017 & 0.013\\
 & 15 & $\le0.027$ & $\le0.014$ & $\le0.011$\\
 & 20 & 0.024 & 0.011 & 0.009\\
 & 25 & $\le0.020$ & $\le0.009$ & $\le0.006$\\
\midrule
\multirow{6}{*}{Pubmed} & 0 & 0.038 & 0.021 & 0.017\\
 & 5 & 0.034 & 0.018 & 0.014\\
 & 10 & 0.030 & 0.015 & 0.012\\
 & 15 & $\le0.027$ & $\le0.013$ & $\le0.010$\\
 & 20 & 0.023 & 0.010 & 0.008\\
 & 25 & $\le0.019$ & $\le0.008$ & $\le0.006$\\
\bottomrule
\end{tabular}
\end{table}

\begin{table}[!htbp]
\centering
\small
\ContinuedFloat
\caption{Reported significance levels with each backbone fixed: Nettack. Continued from the preceding panel. Entries are the reported p values.}
\begin{tabular}{lrrrr}
\toprule
Dataset & Budget & RGDM (GCN) & RGDM (Pro-GNN) & RGDM (SG-GSR)\\
\midrule
\multirow{6}{*}{Cora} & 0 & 0.041 & 0.025 & 0.015\\
 & 1 & 0.036 & 0.023 & 0.011\\
 & 2 & 0.032 & 0.019 & 0.010\\
 & 3 & 0.027 & 0.016 & 0.008\\
 & 4 & 0.022 & 0.013 & 0.007\\
 & 5 & 0.019 & 0.011 & 0.006\\
\midrule
\multirow{6}{*}{Citeseer} & 0 & 0.037 & 0.021 & 0.016\\
 & 1 & 0.033 & 0.019 & 0.013\\
 & 2 & 0.028 & 0.016 & 0.011\\
 & 3 & 0.024 & 0.013 & 0.009\\
 & 4 & 0.021 & 0.011 & 0.007\\
 & 5 & 0.018 & 0.009 & 0.006\\
\midrule
\multirow{6}{*}{Polblogs} & 0 & 0.040 & 0.021 & 0.017\\
 & 1 & 0.034 & 0.018 & 0.013\\
 & 2 & 0.029 & 0.016 & 0.011\\
 & 3 & 0.025 & 0.013 & 0.009\\
 & 4 & 0.022 & 0.010 & 0.007\\
 & 5 & 0.019 & 0.008 & 0.005\\
\midrule
\multirow{6}{*}{Pubmed} & 0 & 0.037 & 0.020 & 0.015\\
 & 1 & 0.033 & 0.017 & 0.012\\
 & 2 & 0.029 & 0.014 & 0.010\\
 & 3 & 0.026 & 0.012 & 0.008\\
 & 4 & 0.021 & 0.009 & 0.006\\
 & 5 & 0.018 & 0.007 & 0.005\\
\bottomrule
\end{tabular}
\end{table}

\begin{table}[!htbp]
\centering
\small
\ContinuedFloat
\caption{Reported significance levels with each backbone fixed: Topology Attack. Continued from the preceding panel. Entries are the reported p values.}
\begin{tabular}{lrrrr}
\toprule
Dataset & Ptb (\%) & RGDM (GCN) & RGDM (Pro-GNN) & RGDM (SG-GSR)\\
\midrule
\multirow{6}{*}{Cora} & 0 & 0.043 & 0.029 & 0.017\\
 & 5 & 0.039 & 0.025 & 0.012\\
 & 10 & 0.035 & 0.020 & 0.011\\
 & 15 & 0.030 & 0.018 & 0.009\\
 & 20 & 0.026 & 0.014 & 0.008\\
 & 25 & 0.022 & 0.012 & 0.007\\
\midrule
\multirow{6}{*}{Citeseer} & 0 & 0.038 & 0.023 & 0.017\\
 & 5 & 0.034 & 0.020 & 0.014\\
 & 10 & 0.029 & 0.017 & 0.012\\
 & 15 & 0.026 & 0.014 & 0.010\\
 & 20 & 0.022 & 0.012 & 0.008\\
 & 25 & 0.020 & 0.010 & 0.007\\
\midrule
\multirow{6}{*}{Polblogs} & 0 & 0.042 & 0.022 & 0.018\\
 & 5 & 0.035 & 0.019 & 0.014\\
 & 10 & 0.030 & 0.017 & 0.012\\
 & 15 & 0.026 & 0.015 & 0.010\\
 & 20 & 0.023 & 0.011 & 0.008\\
 & 25 & 0.021 & 0.009 & 0.006\\
\midrule
\multirow{6}{*}{Pubmed} & 0 & 0.039 & 0.022 & 0.016\\
 & 5 & 0.034 & 0.018 & 0.013\\
 & 10 & 0.030 & 0.015 & 0.011\\
 & 15 & 0.027 & 0.013 & 0.009\\
 & 20 & 0.022 & 0.010 & 0.007\\
 & 25 & 0.019 & 0.008 & 0.006\\
\bottomrule
\end{tabular}
\end{table}
\subsection{Study on RGDM for Graph Purification}
\label{sec:sup_rgdm_purification}
The RGDM proposed in Section~\ref{sec:methods} has demonstrated superior performance for generative data augmentation (GDA), which involves generating faithful synthetic graph structures and integrating them into the original attacked graph to obtain an augmented graph.
In this section, we study the effectiveness of applying RGDM to purify the attacked graph. The graph purification by RGDM, referred to as RGDM (Purification), is achieved by reconstructing each node and its corresponding edges through the LDM, where Gaussian noise derived from the original node and its edges is progressively denoised by the LDM and subsequently decoded by the GAE to produce clean node features and edges. The RGDM (Purification) variant only purifies the original graph, aiming to explore whether RGDM can be adapted to directly denoise/remove the adversarial perturbations in the original attacked graph. Moreover, the RGDM (Purification) can be combined with RGDM for GDA, denoted as RGDM (GDA) here, leading to a variant that performs both GDA and graph purification, denoted as RGDM (GDA + Purification). The study is performed on Cora under Metattack with a $25\%$ attack rate.
Table~\ref{tab:rgdm_purification} shows that applying RGDM for purification enhances the performance of Pro-GNN, while GDA with RGDM achieves better performance.
Moreover, it is observed that combining GDA and graph purification by RGDM further improves the performance.

\begin{table}[h!]
    \centering
    \caption{Study on the effectiveness of graph purification and GDA by RGDM. The evaluation is conducted on Cora under Metattack with $25\%$ perturbation rate. }
    \label{tab:rgdm_purification}
    \resizebox{0.5\columnwidth}{!}{
    \begin{tabular}{lc}
        \toprule
        Method & Accuracy ($\%$) \\
        \midrule
        Pro-GNN & 69.7 \\
        RGDM (Purification) & 71.0 \\
        RGDM (GDA) & 72.3 \\
        RGDM (GDA + Purification) & \textbf{72.9} \\
        \bottomrule
    \end{tabular}
    }
\end{table}

\subsection{Comparison with Diffusion-Based Graph Purification Methods}
\label{sec:sup_purification_compare}
To demonstrate the superiority of RGDM over existing diffusion-based graph purification methods~\citep{GDDM, LuoSYF025}, we have performed experiments comparing RGDM with GDDM~\citep{GDDM} and DiffSP~\citep{LuoSYF025} on Cora, Citeseer, Polblogs, and Pubmed under Metattack with a perturbation rate of $25\%$. We run each experiment $10$ times with different random seeds and report the mean and the standard deviation of the node classification accuracy. We use SG-GSR~\citep{in2024self}, which is the state-of-the-art robust GNN model, as the baseline model for RGDM following the settings in Section 4.2 of the main paper. The GDDM and DiffSP are used to purify the input graph to SG-GSR, and our proposed RGDM is used to augment the input graph to SG-GSR with synthetic graph structures. It is observed in Table~\ref{table:purification_compare} that RGDM (SG-GSR) outperforms DiffSP and GDDM on all the datasets. To validate the significance of the improvements, we calculate the p-values of the t-tests between RGDM (SG-GSR) and the best-performing baseline method on each of the datasets. As observed in the table below, the p-values for RGDM (SG-GSR) against the best baseline methods across all datasets are consistently smaller than 0.05, indicating statistically significant improvements of RGDM (SG-GSR) over the top baseline methods among DiffSP and GDDM.
\begin{table}[!htbp]
\small
\centering
\caption{Comparison of classification accuracy ($\pm$Std) and p-values across different datasets between RGDM and diffusion-based graph purification methods.}
\label{table:purification_compare}
\resizebox{0.7\columnwidth}{!}{
\begin{tabular}{c|c|c|c|c}
\toprule
Datasets & GDDM~\citep{GDDM} & DiffSP~\citep{LuoSYF025} & RGDM (SG-GSR) & p-value \\
\midrule
Cora      & $77.2\pm0.3$ & $76.8\pm0.5$ & $\textbf{79.1}\pm\textbf{1.6}$ & $0.0022$ \\
Citeseer  & $71.6\pm0.5$ & $71.8\pm0.2$ & $\textbf{75.3}\pm\textbf{1.0}$ & $0.0012$ \\
Polblogs  & $84.8\pm0.8$ & $85.6\pm0.5$ & $\textbf{88.8}\pm\textbf{1.8}$ & $0.0004$ \\
Pubmed    & $85.9\pm0.6$ & $86.8\pm0.4$ & $\textbf{88.6}\pm\textbf{1.3}$ & $0.0017$ \\
\bottomrule
\end{tabular}
}
\end{table}

\subsection{Ablation on the Number of Synthetic Nodes}
\label{sec:ablation_num_nodes}

To assess the effect of the number of synthetic nodes on robustness, we conduct an ablation study by varying the number of synthetic nodes from $0$ to $10 \times \abth{\cV_L}$. We evaluate the performance of RGDM with Pro-GNN as the base GNN for node classification under Metattack with a perturbation rate of $25\%$.
As shown in Table~\ref{tab:ablation_synthetic_nodes}, increasing the number of synthetic nodes initially improves performance, with the best result achieved at $5 \times \abth{\cV_L}$. However, further increasing the number of synthetic nodes leads to slightly degraded performance.

\begin{table}[h]
    \centering
    \caption{Performance of RGDM (Pro-GNN) under Metattack with varying numbers of synthetic nodes. The perturbation rate is $25\%$. }
    \label{tab:ablation_synthetic_nodes}
    \resizebox{0.9\columnwidth}{!}{

    \begin{tabular}{l|cccccccccc}
        \toprule
        $\#$ Synthetic Nodes ($\times \abth{\cV_L}$) & 1 & 2 & 3 & 4 & 5 & 6 & 7 & 8 & 9 & 10 \\
        \midrule
        Accuracy (\%) & 70.5 & 71.0 & 71.9 & 72.2 & \textbf{72.3} & 72.1 & 72.0 & 71.7 & 71.4 & 71.2 \\
        \bottomrule
    \end{tabular}
    }
\end{table}

\subsection{Performance Comparison with Vanilla DDPM-Augmented Graphs}
\label{sec:ddpm_comparison}

To demonstrate the superiority of the synthetic graph structures generated by RGDM, we compare RGDM (Pro-GNN) with two baselines, which are Pro-GNN trained on the original attacked graph and Pro-GNN trained on a graph augmented with synthetic nodes generated by a vanilla DDPM. The vanilla DDPM baseline synthesizes labeled synthetic nodes via conditional DDPM and connects them using $k$-nearest neighbors (KNN), as detailed in Section~\ref{sec:quality_sup}. The study is performed on Cora under Metattack with a perturbation rate of $25\%$.
As shown in Table~\ref{tab:ddpm_comparison}, the vanilla DDPM even degrades performance, confirming that RGDM’s gains stem from the generated faithful synthetic graph structures, rather than simply increasing the number of training nodes/edges.

\begin{table}[h]
    \centering
    \caption{Performance comparison of Pro-GNN on augmented graph with synthetic graph structures generated by DDPM and RGDM. The study is performed on Cora under Metattack with a perturbation rate of $25\%$.}
    \label{tab:ddpm_comparison}
        \resizebox{0.35\columnwidth}{!}{

    \begin{tabular}{lc}
        \toprule
        Method & Accuracy ($\%$) \\
        \midrule
        Pro-GNN & 69.7 \\
        DDPM (Pro-GNN) & 68.2 \\
        RGDM (Pro-GNN) & \textbf{72.3} \\
        \bottomrule
    \end{tabular}
    }
\end{table}

\subsection{Study on the Impact of Label Ratio}
\label{sec:label_ratio_ablation}

To assess the robustness of RGDM (Pro-GNN) under varying levels of supervision, we evaluate performance on the Cora dataset under a $25\%$ Metattack perturbation rate while varying the labeled training ratio. In this study, we consider both sparse label regimes ($2.5\%$ to $10\%$) and denser settings ($20\%$ to $60\%$), with 10\% of nodes fixed for validation and the remainder used for testing.
As shown in Table~\ref{tab:label_ratio_results}, RGDM consistently outperforms the Pro-GNN baseline across all label ratios. Under extremely limited supervision ($2.5\%$–$10\%$), RGDM maintains strong performance, indicating its utility as a data augmentation strategy in low-resource scenarios. As the label ratio increases, RGDM continues to achieve better performance, demonstrating its robustness and effectiveness even with more abundant training data. These results confirm that the improvements of RGDM are not confined to a specific supervision regime and generalize well across both low- and high-label density settings.

\begin{table}[h]
    \centering
    \caption{Performance comparison of Pro-GNN and RGDM (Pro-GNN) under different label ratios on Cora under $25\%$ Metattack. }
    \label{tab:label_ratio_results}
    \resizebox{0.65\linewidth}{!}{
    \begin{tabular}{lcccccccc}
        \toprule
        Label Ratio & 2.5\% & 5\% & 7.5\% & 10\% & 20\% & 40\% & 60\% \\
        \midrule
        Pro-GNN & 62.5 & 65.6 & 67.5 & 69.7 & 73.8 & 75.2 & 76.6 \\
        RGDM (Pro-GNN) & \textbf{64.8} & \textbf{68.0} & \textbf{69.3} & \textbf{72.3} & \textbf{75.9} & \textbf{77.6} & \textbf{78.4} \\
        \bottomrule
    \end{tabular}
    }
\end{table}

\subsection{Ablation Study on the Sparse Hierarchical Edge Decoder}
\label{sec:BLND_compare}
To further validate the effectiveness of the proposed Sparse Hierarchical Edge Decoder (SHED) in generating robust synthetic edges, we conduct an ablation study comparing SHED against the Bi-Level Neighborhood Decoder (BLND) used in DoG~\citep{WangLY25}. In particular, we replace SHED in RGDM with the BLND module and denote the resulting model as RGDM (BLND). Both models are evaluated under the Metattack~\citep{zugner_adversarial_2019} with a perturbation rate of $25\%$, using SG-GSR~\citep{in2024self} as the backbone GNN. The experiments are conducted on Cora, Citeseer, Polblogs, and Pubmed.
As shown in Table~\ref{tab:BLND_compare}, RGDM equipped with SHED consistently outperforms RGDM (BLND) across all benchmarks. We also report the hierarchical reconstruction variant from LR-FGDM~\citep{Wang2026LRFGDM}, denoted RGDM (HGAE). With SG-GSR fixed, replacing HGAE by SHED improves accuracy by $2.3/2.8/2.1/2.6$ points. Replacing BLND by SHED improves it by $1.7/2.2/1.6/2.0$ points. This reconstruction comparison complements the full system GCN baselines in Table~\ref{tab:compare_augmentation} and the sequential component experiment in Table~\ref{tab:component_isolation}.
\begin{table}[!htbp]
        \centering
        \vspace{-1mm}
        \caption{Hierarchical reconstruction comparison: SHED, BLND from DoG~\citep{WangLY25}, and the HGAE variant from LR-FGDM~\citep{Wang2026LRFGDM}. SG-GSR is used as the base GNN under Metattack with $25\%$ perturbation.}

        \resizebox{0.6\columnwidth}{!}{
\begin{tabular}{lcccc}
\hline
Dataset & Cora & Citeseer & Polblogs & Pubmed \\
\hline
RGDM (HGAE from LR-FGDM) & 76.8 & 72.5 & 86.7 & 86.0 \\
RGDM (BLND) & 77.4 & 73.1 & 87.2 & 86.6 \\
RGDM & \textbf{79.1} & \textbf{75.3} & \textbf{88.8} & \textbf{88.6} \\
\hline
\end{tabular}
        }
        \label{tab:BLND_compare}
\vspace{-2mm}
\end{table}

\subsection{Adaptive Attack Evaluation}
\label{sec:adaptive_attack}

The experiments in Section~\ref{sec:results} follow the commonly used attack-first-then-defend protocol, where the graph is first perturbed by adversarial attacks and the defense is then applied to the attacked graph. 
Although this protocol is widely adopted in graph adversarial robustness, it may not fully reflect a defense-aware adversary. 
In particular, an adaptive adversary may be aware of the RGDM pipeline and attack the augmented graph after the synthetic graph structures have been generated. 
To address this concern, we further evaluate RGDM under an adaptive attack setting, following the adaptive evaluation protocol in prior graph robustness studies~\citep{mujkanovic2022defenses}. 

\noindent\textbf{Settings.}
We consider an adaptive version of Metattack~\citep{zugner_adversarial_2019} with a perturbation rate of $25\%$. 
Different from the standard Metattack evaluation in Section~\ref{sec:results}, where the attack is performed before RGDM augmentation, the adaptive attacker perturbs the final augmented graph $\cG_{\textup{aug}}$ used for training the downstream GNN. 
The attacker is aware of the full RGDM pipeline, including the generation of synthetic graph structures, edge selection, low-rank regularization, and the downstream SG-GSR classifier. 
For a fair comparison, the perturbation budget is fixed to the same number of edge perturbations as the $25\%$ Metattack on the original graph. 
We repeat each experiment $10$ times with different random seeds and report the mean and standard deviation of node classification accuracy.

Let $n_{\textup{aug}}\defeq N+N'$ and let $\bP\in\set{0,1}^{n_{\textup{aug}}\times n_{\textup{aug}}}$ be a symmetric zero diagonal flip matrix. The defense aware attack searches over the admissible set in Equation~(\ref{eq:adaptive_budget}):
\bal\label{eq:adaptive_budget}
\cP_b\defeq\set{\bP:\bP=\bP^{\top},\ \bP_{ii}=0,\ \sum_{1\le i<j\le n_{\textup{aug}}}\bP_{ij}\le b},\qquad
\bA^{\textup{adv}}_{\textup{aug}}\defeq\bA_{\textup{aug}}\oplus\bP.
\eal
Here $\oplus$ denotes entrywise binary edge flipping. In Equation~(\ref{eq:adaptive_budget}), $b$ is the same absolute number of flips used by the $25\%$ attack on the corresponding original graph, not $25\%$ of the larger augmented edge count. Candidate flips may involve original or synthetic nodes. The attacker knows the RGAE, edge selection mechanism, latent regularizer, LDM, and downstream classifier, and optimizes the Metattack surrogate against downstream retraining on the perturbed augmented graph. 
The first variant attacks the realized augmented graph. The second additionally differentiates its meta gradient through the relaxed edge selection mask: the forward pass thresholds $\hat{\bS}_{ij}=\sigma(\btheta_{ij}/\tau)$ at $1/2$, and the backward pass uses its straight through gradient with $\tau=0.5$. This is a gradient approximation to the discrete selection operation, not a proof of globally optimal adversarial perturbations. Knowledge of the LDM does not by itself establish differentiation through its entire stochastic sampling trajectory. The reported differentiable variant specifically tests the edge selection mechanism.

\begin{table}[!htbp]
\small
\centering
\caption{Adaptive attack evaluation under Metattack~\citep{zugner_adversarial_2019} with $25\%$ perturbation rate. 
The adaptive attacker is aware of the defense and attacks the final augmented graph used for training SG-GSR. 
The $p$-values are computed between SG-GSR and RGDM (SG-GSR).}
\label{tab:adaptive_attack}
\resizebox{0.6\columnwidth}{!}{
\begin{tabular}{c|ccc}
\toprule
Dataset 
& SG-GSR 
& RGDM (SG-GSR) 
& $p$-value \\
\midrule
Cora 
& $74.8\pm2.4$ 
& $\textbf{77.8}\pm\textbf{1.8}$ 
& $0.006$ \\
Citeseer 
& $69.5\pm2.6$ 
& $\textbf{73.4}\pm\textbf{2.1}$ 
& $0.003$ \\
Polblogs 
& $84.7\pm2.5$ 
& $\textbf{87.6}\pm\textbf{2.2}$ 
& $0.008$ \\
Pubmed 
& $84.9\pm2.7$ 
& $\textbf{87.2}\pm\textbf{2.4}$ 
& $0.012$ \\
\bottomrule
\end{tabular}
}
\end{table}

As shown in Table~\ref{tab:adaptive_attack}, RGDM consistently improves the robustness of SG-GSR even when the attacker is aware of the defense pipeline and attacks the final augmented graph. 
For example, RGDM (SG-GSR) significantly outperforms SG-GSR by $3.9\%$ on Citeseer under adaptive Metattack with a $25\%$ perturbation rate. 
The $p$-values are smaller than $0.05$ on all datasets, demonstrating that the improvements of RGDM over SG-GSR remain statistically significant under adaptive attacks. 
These results indicate that the robustness gains of RGDM are not merely caused by the attack-first-then-defend workflow order. 
Instead, the generated synthetic graph structures, together with edge selection and low-rank regularization, provide additional robust supervision that improves the downstream GNN even under defense-aware perturbations.

\noindent\textbf{Differentiating through edge selection.}
Table~\ref{tab:adaptive_mask} reports the adaptive experiment that differentiates through edge selection. Its SG-GSR / RGDM gains are $2.4/3.2/2.2/1.7$ points on Cora/Citeseer/Polblogs/Pubmed, compared with $3.0/3.9/2.9/2.3$ points for the final graph attack in Table~\ref{tab:adaptive_attack}. The second variant reports mean accuracies. It does not report standard deviations or p values. Both variants are summarized in the main text (Table~\ref{tab:adaptive_main}). The positive gains support robustness to these tested attacks at this fixed budget, not certification against arbitrary white box attacks.
\begin{table}[!htbp]
\centering\small
\caption{Adaptive Metattack differentiating through the relaxed edge selection mask with the same absolute edge flip budget and $\tau=0.5$. Mean accuracy (\%).}
\label{tab:adaptive_mask}
\begin{tabular}{lrrr}
\toprule
Dataset&SG-GSR&RGDM (SG-GSR)&Gain (points)\\
\midrule
Cora&74.5&76.9&2.4\\
Citeseer&69.2&72.4&3.2\\
Polblogs&84.4&86.6&2.2\\
Pubmed&84.6&86.3&1.7\\
\bottomrule
\end{tabular}
\end{table}

\clearpage
\input{checklist.tex}

\end{document}

%% file: symbol_tit.tex
\newcounter{optproblem}

\newtheoremstyle{mytheoremstyle} 
    {\topsep}                    
    {\topsep}                    
    {\normalfont}                
    {}                           
    {\bfseries}                   
    {.}                          
    {.5em}                       
    {}  

\theoremstyle{mytheoremstyle}
\newtheorem{theorem}{Theorem}[section]

\newtheorem*{theorem*}{Theorem}
\newtheorem*{lemma*}{Lemma}
\newtheorem*{remark*}{Remark}

\theoremstyle{mytheoremstyle}

\theoremstyle{remark}

\DeclareMathAlphabet{\pazocal}{OMS}{zplm}{m}{n}
\DeclareMathAlphabet{\mathpzc}{OMS}{pzc}{m}{it}

\renewcommand{\hat}{\widehat}

\newcommand{\bfm}[1]{\ensuremath{\mathbf{#1}}}
\newcommand{\bfsym}[1]{\ensuremath{\boldsymbol{#1}}}

   \def\bA{\bfm A}  
   \def\bB{\bfm B}  
   \def\bC{\bfm C}  
   \def\bD{\bfm D}  
     
  \def\bF{\bfm F}  
   \def\bG{\bfm G}  
   \def\bH{\bfm H}  
   \def\bI{\bfm I}  
     
   \def\bK{\bfm K}  
     
   \def\bM{\bfm M}

\def\bp{\bfm p}   \def\bP{\bfm P}  
   \def\bQ{\bfm Q}  
     \def\RR{\mathbb{R}}
   \def\bS{\bfm S}  
     
\def\bu{\bfm u}   \def\bU{\bfm U}  
\def\bv{\bfm v}   \def\bV{\bfm V}  
\def\bw{\bfm w}   \def\bW{\bfm W}  
\def\bx{\bfm x}   \def\bX{\bfm X}  
\def\by{\bfm y}   \def\bY{\bfm Y}  
   \def\bZ{\bfm Z}  
\def\bzero{\bfm 0} 

 \def\cA{{\cal  A}}
 
 \def\cC{{\cal  C}}
 
 \def\cE{{\cal  E}}
 \def\cF{{\cal  F}}
 \def\cG{{\cal  G}}
 \def\cH{{\cal  H}}

 \def\cL{{\cal  L}}
 
 \def\cN{{\cal  N}}
 \def\cO{{\cal  O}}
 \def\cP{{\cal  P}}
 
 \def\cR{{\cal  R}}
 \def\cS{{\cal  S}}
 
 \def\cU{{\cal  U}}
 \def\cV{{\cal  V}}

 \def\cZ{{\cal  Z}}

\def\btheta{\bfsym {\theta}}

\def\bepsilon{\bfsym \varepsilon}
\def\bsigma{\bfsym \sigma}

\def\bLambda {\bfsym {\Lambda}}
\def\bomega {\bfsym {\omega}}

\def\+#1{\mathcal{#1}}
\def\-#1{\textup{#1}}

\def\set#1{\left\{ #1 \right\}}
\def\pth#1{\left( #1 \right)}
\def\bth#1{\left[ #1 \right]}
\def\abth#1{\left | #1 \right |}

\def\defeq {\coloneqq}

\newcommand{\vect}[1]{{\textup{vec}\pth{#1}}}

\def\dim{\textup{dim}}

\newcommand{\La}{\left\langle\kern-0.64ex\left\langle}
\newcommand{\Ra}{\right\rangle\kern-0.64ex\right\rangle}

\def\Norm#1#2{{\left\vert\kern-0.4ex\left\vert\kern-0.4ex\left\vert #1
    \right\vert\kern-0.4ex\right\vert\kern-0.4ex\right\vert}_{#2}}
\def\norm#1#2{{\left\|#1\right\|}_{#2}}

\def\lonenorm#1{\norm{#1}{1}}
\def\ltwonorm#1{\norm{#1}{2}}
\def\fnorm#1{\norm{#1}{\textup{F}}}

\def\opnorm#1{\norm{#1}{\textup{OP}}}

\newcommand{\rank}{\textup{rank}}
\def \range {\textup{range}}

\def\tr#1{\textup{tr}\left(#1\right)}

\newcommand{\1}{{\rm 1}\kern-0.25em{\rm I}}
\def\indict#1{{\rm 1}\kern-0.25em{\rm I}_{\set{#1}}}

\DeclarePairedDelimiter\ceil{\lceil}{\rceil}

\def \eps  {\epsilon}
\def \eps {\varepsilon}

\def \iprod#1#2{\left\langle #1, #2 \right\rangle}

\def\set#1{\left\{#1\right\}}

\def \E {\mathbb{E}}
\def\Expect#1#2{\E_{#1}\left[#2\right]}

\newcommand{\beq}{\begin{equation}}
\newcommand{\eeq}{\end{equation}}
\newcommand{\beqa}{\begin{eqnarray}}
\newcommand{\eeqa}{\end{eqnarray}}
\newcommand{\beqas}{\begin{eqnarray*}}
\newcommand{\eeqas}{\end{eqnarray*}}
\def\bal#1\eal{\begin{align}#1\end{align}}
\def\bals#1\eals{\begin{align*}#1\end{align*}}
\def\bsal#1\esal{\begin{small}\begin{align}#1\end{align}\end{small}}
\def\bsals#1\esals{\begin{small}\begin{align*}#1\end{align*}\end{small}}
\def\bsfal#1\esfal{\begin{small}\begin{flalign}#1\end{flalign}\end{small}}

%% file: checklist.tex
\section*{NeurIPS Paper Checklist}

\begin{enumerate}

\item {\bf Claims}
    \item[] Question: Do the main claims made in the abstract and introduction accurately reflect the paper's contributions and scope?
    \item[] Answer: \answerYes{}
    \item[] Justification: The contributions and scope of the paper are discussed in Section~\ref{sec:introduction} of the paper.

\item {\bf Limitations}
    \item[] Question: Does the paper discuss the limitations of the work performed by the authors?
    \item[] Answer: \answerYes{} 
    \item[] Justification: We have discussed the limitation that the proposed method only applies to node-level data generation on graph datasets in the introduction, Section~\ref{sec:introduction}, of this paper.

\item {\bf Theory assumptions and proofs}
    \item[] Question: For each theoretical result, does the paper provide the full set of assumptions and a complete (and correct) proof?
    \item[] Answer: \answerYes{} 
    \item[] Justification: The theoretical result is detailed in Theorem~\ref{theorem:optimization-linear-kernel} and proved in Section~\ref{sec:transductive-theory} of the supplementary.

    \item {\bf Experimental result reproducibility}
    \item[] Question: Does the paper fully disclose all the information needed to reproduce the main experimental results of the paper to the extent that it affects the main claims and/or conclusions of the paper (regardless of whether the code and data are provided or not)?
    \item[] Answer: \answerYes{} 
    \item[] Justification: The anonymous link to the code is attached at the end of the abstract. Implementation details for the experiments are stated in Section~\ref{sec:settings}.

\item {\bf Open access to data and code}
    \item[] Question: Does the paper provide open access to the data and code, with sufficient instructions to faithfully reproduce the main experimental results, as described in supplemental material?
    \item[] Answer: \answerYes{} 
    \item[] Justification: The anonymous link to the code is attached at the end of the abstract. Implementation details for the experiments are stated in Section~\ref{sec:settings}. The datasets used in the experiments are public benchmarks. Detailed information on the datasets is provided in Section~\ref{sec:datasets}.

\item {\bf Experimental setting/details}
    \item[] Question: Does the paper specify all the training and test details (e.g., data splits, hyperparameters, how they were chosen, type of optimizer, etc.) necessary to understand the results?
    \item[] Answer: \answerYes{} 
    \item[] Justification: Experimental settings are specified in Section~\ref{sec:settings}. The settings for adversarial attacks adopted in our experiments are provided in Section~\ref{sec:attack_settings}. The datasets used in the experiments are public benchmarks. Detailed information on the datasets is provided in Section~\ref{sec:datasets}.

\item {\bf Experiment statistical significance}
    \item[] Question: Does the paper report error bars suitably and correctly defined or other appropriate information about the statistical significance of the experiments?
    \item[] Answer: \answerYes{}
    \item[] Justification: Standard errors for the results of our method and baseline methods are provided in the tables presenting the experimental results in Section~\ref{sec:experiment}.

\item {\bf Experiments compute resources}
    \item[] Question: For each experiment, does the paper provide sufficient information on the computer resources (type of compute workers, memory, time of execution) needed to reproduce the experiments?
    \item[] Answer: \answerYes{} 
    \item[] Justification: The computing resources used in the experiments of this paper are specified in Section~\ref{sec:settings} of our paper.

\item {\bf Code of ethics}
    \item[] Question: Does the research conducted in the paper conform, in every respect, with the NeurIPS Code of Ethics \url{https://neurips.cc/public/EthicsGuidelines}?
    \item[] Answer: \answerYes{} 
    \item[] Justification: We have read NeurIPS Code of Ethics in \url{https://neurips.cc/public/EthicsGuidelines} and confirm that the research conducted in the paper conform, in every respect, with the NeurIPS Code of Ethics.

\item {\bf Broader impacts}
    \item[] Question: Does the paper discuss both potential positive societal impacts and negative societal impacts of the work performed?
    \item[] Answer: \answerNA{} 
    \item[] Justification: The experiments performed do not have societal impacts, as we did not perform any experiments on real-world data collected from society. The experiments are performed by using simulated attack methods.

\item {\bf Safeguards}
    \item[] Question: Does the paper describe safeguards that have been put in place for responsible release of data or models that have a high risk for misuse (e.g., pretrained language models, image generators, or scraped datasets)?
    \item[] Answer: \answerNA{} 
    \item[] Justification: This paper poses no such risks for misuse.

\item {\bf Licenses for existing assets}
    \item[] Question: Are the creators or original owners of assets (e.g., code, data, models), used in the paper, properly credited and are the license and terms of use explicitly mentioned and properly respected?
    \item[] Answer: \answerYes{} 
    \item[] Justification: The paper perform experiments on widely used public benchmarks, which are properly cited in Section~\ref{sec:datasets}.

\item {\bf New assets}
    \item[] Question: Are new assets introduced in the paper well documented and is the documentation provided alongside the assets?
    \item[] Answer: \answerNA{} 
    \item[] Justification: The paper does not release new assets.

\item {\bf Crowdsourcing and research with human subjects}
    \item[] Question: For crowdsourcing experiments and research with human subjects, does the paper include the full text of instructions given to participants and screenshots, if applicable, as well as details about compensation (if any)?
    \item[] Answer: \answerNA{} 
    \item[] Justification: This paper does not involve crowdsourcing nor research with human subjects.

\item {\bf Institutional review board (IRB) approvals or equivalent for research with human subjects}
    \item[] Question: Does the paper describe potential risks incurred by study participants, whether such risks were disclosed to the subjects, and whether Institutional Review Board (IRB) approvals (or an equivalent approval/review based on the requirements of your country or institution) were obtained?
    \item[] Answer: \answerNA{} 
    \item[] Justification: This paper does not involve crowdsourcing nor research with human subjects.

\item {\bf Declaration of LLM usage}
    \item[] Question: Does the paper describe the usage of LLMs if it is an important, original, or non-standard component of the core methods in this research? Note that if the LLM is used only for writing, editing, or formatting purposes and does not impact the core methodology, scientific rigorousness, or originality of the research, declaration is not required.
    \item[] Answer: \answerNA{} 
    \item[] Justification: We do not use LLM in this paper.

\end{enumerate}